\def\paperversion{1}

\if\paperversion2
\documentclass[twoside,11pt]{article}
\usepackage{multirow}
\usepackage{jmlr2e}
\fi

\if\paperversion1
\documentclass[opre,nonblindrev]{informs3}
\SingleSpacedXI

\renewcommand{\theARTICLETOP}{}
\renewcommand{\theARTICLERULE}{\vspace{12pt}}
\renewcommand{\theRRHFirstLine}{}
\renewcommand{\theRRHSecondLine}{}
\renewcommand{\theLRHFirstLine}{}
\renewcommand{\theLRHSecondLine}{}
\renewcommand{\theECLRHFirstLine}{}%
\renewcommand{\theECLRHSecondLine}{}%
\renewcommand{\theECRRHFirstLine}{}%
\renewcommand{\theECRRHSecondLine}{}%

\makeatletter
\renewcommand\section{\@startsection {section}{1}{\z@}{-13pt plus -6pt minus -3pt}{4pt}%
  {\fs.13.15.\bfseries\RAGG}}%
\renewcommand\subsection{\@startsection{subsection}{2}{\z@}{-13pt plus -6pt minus -3pt}{4pt}%
  {\TEN\bfseries\RAGG}}%
\makeatother
\let\FigureCaptionFontStyle\undefined
\def\FigureCaptionFontStyle{\mdseries}

\usepackage[scaled=.98,sups,osf]{XCharter}%

\usepackage{booktabs}
\usepackage{multicol}
\usepackage{multirow}
\usepackage[numbers]{natbib}
\usepackage{bm}
\usepackage{floatflt}
 \def\bibfont{\small}%
\TheoremsNumberedThrough   
\ECRepeatTheorems
\theoremstyle{TH}%
\fi 

\usepackage{wrapfig}
\usepackage{endnotes}
\usepackage{anyfontsize}

\let\footnote=\endnote
\newcommand{\RNum}[1]{\uppercase\expandafter{\romannumeral #1\relax}}
\newcommand{\cO}{\mathcal{O}}

\newcommand*{\QED}{%
\leavevmode\unskip\penalty9999 \hbox{}\nobreak\hfill
    \quad\hbox{$\square$}%
}
\newcommand*{\QDEF}{%
\leavevmode\unskip\penalty9999 \hbox{}\nobreak\hfill
    \quad\hbox{$\diamondsuit$}%
}
\newcommand*{\QEG}{%
\leavevmode\unskip\penalty9999 \hbox{}\nobreak\hfill
    \quad\hbox{$\clubsuit$}%
}
\usepackage{booktabs}
\usepackage[font=small]{caption}
\usepackage[size=small]{subcaption}
\DeclareMathAlphabet{\mathsfit}{T1}{\sfdefault}{\mddefault}{\sldefault}
\SetMathAlphabet{\mathsfit}{bold}{T1}{\sfdefault}{\bfdefault}{\sldefault}
\DeclareMathAlphabet{\mathcal}{OMS}{cmsy}{m}{n}
\usepackage{enumitem}
\setlist[enumerate,1]{label=\normalfont{(\Roman*)},leftmargin=*}
\def\argmax{\mathop{\rm arg\,max}}%
\def\argmin{\mathop{\rm arg\,min}}%
\usepackage[margin=1in]{geometry}
\usepackage{caption}
\usepackage{mathtools}
\usepackage{algorithm}
\usepackage{algorithmic}
\RequirePackage{xcolor}
\definecolor{dkgreen}{rgb}{0,0.6,0}
\definecolor{gray}{rgb}{0.5,0.5,0.5}
\definecolor{mauve}{rgb}{0.58,0,0.82}

\DeclareMathOperator{\diag}{diag}
\newcommand{\cv}[1]{\mathbf{#1}}

\newcommand{\bS}{\mathbb{S}}
\newcommand{\bR}{\mathbb{R}}
\newcommand{\bF}{\mathbb{F}}
\newcommand{\cH}{\mathcal{H}}
\newcommand{\dtrue}{d_{\mathrm{true}}}

\newcommand{\cZ}{\mathcal{Z}}

\newcommand{\trans}{^{\mathrm T}}

\newcommand{\MMD}{\mathrm{MMD}}

\def\Tr{\mathop{{\rm Tr}}}
\def\Te{\mathop{{\rm Te}}}
\newcommand{\cQ}{\mathcal{Q}}
\newcommand{\diff}{\,\mathrm{d}}

\DeclarePairedDelimiterX{\inp}[2]{\langle}{\rangle}{#1, #2}

\newcommand{\bE}{\mathbb{E}}
\newcommand{\bP}{\mathbb{P}}

\renewcommand{\algorithmiccomment}[1]{//#1}
\if\paperversion2
\newtheorem{assumption}{Assumption}[section]
\fi
\usepackage{hyperref}
\hypersetup{%
  colorlinks=true,%
  linkcolor={blue!66!black},%
  linkbordercolor=red,%
  citecolor={blue!66!black},
}
\if\paperversion2
\usepackage{lastpage}
\jmlrheading{25}{2024}{1-\pageref{LastPage}}{12/20}{N/A}{24-2129}{Jie Wang, Santanu S. Dey and Yao Xie}
\fi
\if\paperversion2
\ShortHeadings{MMD Variable Selection}{Jie Wang, Santanu S. Dey and Yao Xie}
\firstpageno{1}
\fi

\if\paperversion1 
\TITLE{Variable Selection for Kernel Two-Sample Tests}
\ARTICLEAUTHORS{%
\AUTHOR{Jie Wang}
\AFF{School of Industrial and Systems Engineering, Georgia Institute of Technology, Atlanta, GA 30332, \EMAIL{jwang3163@gatech.edu}}
\AUTHOR{Santanu S. Dey}
\AFF{School of Industrial and Systems Engineering, Georgia Institute of Technology, Atlanta, GA 30332, \EMAIL{santanu.dey@isye.gatech.edu}}
\AUTHOR{Yao Xie}
\AFF{School of Industrial and Systems Engineering, Georgia Institute of Technology, Atlanta, GA 30332, \EMAIL{yao.xie@isye.gatech.edu}}
} %
\RUNAUTHOR{Anonymous Authors}
\RUNTITLE{MMD Variable Selection}
\ABSTRACT{
We consider the variable selection problem for two-sample tests, aiming to select the most informative variables to determine whether two collections of samples follow the same distribution. 
To address this, we propose a novel framework based on the kernel maximum mean discrepancy (MMD). Our approach seeks a subset of variables with a pre-specified size that maximizes the variance-regularized kernel MMD statistic. We focus on three commonly used types of kernels: linear, quadratic, and Gaussian. 
From a computational perspective, we derive mixed-integer programming formulations and propose exact and approximation algorithms with performance guarantees to solve these formulations. From a statistical viewpoint, we derive the rate of testing power of our framework under appropriate conditions. These results show that the sample size requirements for the three kernels depend crucially on the number of selected variables, rather than the data dimension. 
Experimental results on synthetic and real datasets demonstrate the superior performance of our method, compared to other variable selection frameworks, particularly in high-dimensional settings.
}
\fi

\begin{document}

\if\paperversion2
\title{Variable Selection for Kernel Two-Sample Tests}

\author{\name Jie Wang \email jwang3163@gatech.edu \\
       \addr School of Industrial and Systems Engineering\\
       Georgia Institute of Technology\\
       Atlanta, GA 30332
       \AND
       \name 
       Santanu S. Dey \email santanu.dey@isye.gatech.edu \\
       \addr School of Industrial and Systems Engineering\\
       Georgia Institute of Technology\\
       Atlanta, GA 30332
       \AND 
       \name  Yao Xie \email yao.xie@isye.gatech.edu\\
       \addr School of Industrial and Systems Engineering\\
       Georgia Institute of Technology\\
       Atlanta, GA 30332
       }

\editor{N/A}
\fi
\maketitle

\if\paperversion2
\begin{abstract}
We consider the variable selection problem for two-sample tests, aiming to select the most informative variables to determine whether two collections of samples follow the same distribution. 
To address this, we propose a novel framework based on the kernel maximum mean discrepancy (MMD). Our approach seeks a subset of variables with a pre-specified size that maximizes the variance-regularized kernel MMD statistic. We focus on three commonly used types of kernels: linear, quadratic, and Gaussian. 
From a computational perspective, we derive mixed-integer programming formulations and propose exact and approximation algorithms with performance guarantees to solve these formulations. From a statistical viewpoint, we derive the rate of testing power of our framework under appropriate conditions. These results show that the sample size requirements for the three kernels depend crucially on the number of selected variables, rather than the data dimension. 
Experimental results on synthetic and real datasets demonstrate the superior performance of our method, compared to other variable selection frameworks, particularly in high-dimensional settings.
\end{abstract}
\fi

\section{Introduction}
\label{submission}

Two-sample test is a classic problem in statistics and an important tool for scientific discovery. Given two sets of observations $\cv x^n:=\{x_i\}_{i=1}^n$ and $\cv y^n=\{y_i\}_{i=1}^n$\footnote{In this paper, we consider the simplified setting where the two sets of samples are the same size; it can be generalized to the setting when the two sets of samples do not have the same size.}, which represent $n$ independent and identically distributed~(i.i.d.) $D$-dimensional samples from distributions $\mu$ and $\nu$, respectively. Using these samples, we aim to decide whether $\mu$ and $\nu$ are distinct. Two-sample test has wide applications: for example, in clinical trials to evaluate the effectiveness of two distinct treatments on patient outcomes; in finance, to compare the performance of two different investment strategies; and in machine learning, to investigate whether the source domain and target domain have significant differences.

Recently, kernel two-sample tests have become a popular approach for modern high-dimensional data (see, e.g., \cite{Gretton12,cheng2021kernel}). Despite the vast literature on kernel two-sample tests, studying variable selection for two-sample testing remains relatively limited. In this context, variable selection seeks the most informative $d$ variables from a pool of $D$ variables (usually $d\ll D$) to differentiate distributions $\mu$ and $\nu$.
On the one hand, finding interpretable variables is crucial for understanding population differences for
domains such as gene expression analysis, where only a small subset of variables elucidates disparities between normal and abnormal populations~\citep{zeisel2015cell}.
On the other hand, the dissimilarities between high-dimensional datasets often exhibit a low-dimensional structure~\citep{wang2022manifold}, and thus, extracting a small set of crucial variables as a pre-processing step may enhance the efficacy of high-dimensional two-sample testing.

Variable selection in the kernel two-sample testing is very different from the widely studied variable selection problem in linear models (notably, Lasso) and generalized linear models  (see, e.g., \cite{hastie2009elements}) because we face completely different objective functions in the optimization formulation.  An interesting aspect of the problem is that depending on the choice of the kernel, the nature of the optimization objective can range from simple to hard. Moreover, the original formulation of the variable selection problem will also lead to the so-called ``subset selection problem'' \cite{miller2002subset}, which leads to an integer program and can be hard to solve directly; it remains a question of how to develop computationally efficient procedures and approximate algorithms. On the other hand, for analyzing statistical performance variable selection for kernel two-sample tests facing finite samples, we need to study the statistical performance for tests (such as the false-detection and detection power), which is characteristic of testing problems that are different from regression type of prediction problems.

In this paper, we provide a novel variable selection framework for two-sample testing by choosing key variables that maximize the variance-regularized kernel maximum mean discrepancy~(MMD) statistic, which in turn (approximately) maximizes the corresponding testing power.
Our focus is on three types of kernels: linear, quadratic, and Gaussian.
The contributions are summarized as follows.
\begin{itemize}
    \item 
From the computational perspective, we leverage mixed-integer programming techniques to solve the MMD optimization problem for variable selection.
For linear kernel, we reformulate the optimization as an \emph{inhomogeneous} quadratic maximization with $\ell_2$ and $\ell_0$ norm constraints~(see Section~\ref{Sec:MMD:linear}), called {\bf S}parse {\bf T}rust {\bf R}egion {\bf S}ubproblem~(STRS).
Despite its NP-hardness, we provide an exact mixed-integer semi-definite programming formulation together with exact and approximation algorithms for solving this problem.
To the best of our knowledge, this study is new in the literature.
For quadratic and Gaussian kernels, the MMD optimization becomes a sparse maximization of a non-concave function~(see Section~\ref{Sec:MMD:general}), which is intractable in general. 
We propose a heuristic algorithm that iteratively optimizes a quadratic approximation of the objective function, which is also a special case of STRS.
    \item
From the statistical perspective, we derive the rate of testing power of our framework under appropriate conditions.
We demonstrate that when the training sample size $n_{\Tr}$ is sufficiently large, the type-II error decays in the order of $n_{\Te}^{-1/2}$, where $n_{\Te}$ denotes the testing sample size.
For the three focused types of kernels, the training sample size requirement is almost independent of the data dimension $D$ but dependent on the number of selected variables $d$:
For linear, quadratic, and Gaussian kernels, to achieve satisfactory performance, the training sample sizes are at least $\Omega(d^2\log\frac{D}{d}), \Omega(d^4\log\frac{D}{d})$, and $\Omega(d\log\frac{D}{d})$, respectively.
    \item
By combining both viewpoints, it becomes evident that there exists a balance between computational tractability and statistical guarantees. While the Gaussian kernel requires a smaller sample size to be statistically powerful, the corresponding MMD optimization is challenging. Conversely, the linear or quadratic kernel may require more samples to be statistically powerful, but the optimization is easier to solve.
\item
Finally, we perform numerical experiments using synthetic and real datasets to showcase the effectiveness of our proposed framework compared to other baseline methods. We utilize synthetic datasets to evaluate the testing power and variable selection performance. Subsequently, we test our approach on real data, including the standard MNIST handwritten digits and more specialized sepsis detection datasets.
\end{itemize}

\noindent{\bf Notations.}
Given a positive integer $n$, define $[n]=\{1,\ldots,n\}$.
Let $\bF=\{0,1\}$, and $\bS_n^+$ denote the collection of $n\times n$ symmetric positive semi-definite matrices.
Given a vector $z\in\bR^D$ and a set $S\subseteq[D]$, we use $z^{(k)}$ denote the $k$-th entry in $z$, and $z^{(S)}$ to denote the subvector with entries indexed by $S$.
Given an $m\times n$ matrix $A$ and two sets $ {S\subseteq[m], t\subseteq[n]}$,  denote $A^{(i,j)}$ the $(i,j)$-th entry in $A$ and denote $A^{(S, T)}$ as the submatrix with rows and columns indexed by $S$ and $T$.
Given a vector $z\in\bR^D$ and a distribution $\mu$ in $\bR^D$, denote $z\circ\mu$ as the distribution of the random variable $\sum_{k\in[D]}z^{(k)}x^{(k)}$ provided that $x\sim \mu$.
Define the norm $\|z\|_{(d)}=\max_{S:~|S|\le d}\left\|z^{(S)}\right\|_2$.
For a $D$-dimensional distribution $\mu$ and $s\in[D]$, let $\mathrm{Proj}_{s\#}\mu$ be the $s$-th marginal distributions of $\mu$.

\subsection*{Related work}

\noindent{\bf Variable selection.}
Classical variable selection approaches seek to extract the most valuable features from a group of high-dimensional data points.
To name a few, sparse PCA seeks to select crucial variables that maximize the sample covariance based on sample sets~\citep{li2020exact, dey2022solving, dey2022using}; truncated SVD aims to formulate a low-rank data matrix with minimum approximation error~\citep{li2020best}, and the maximum entropy sampling or experiment design aims to select a subgroup of samples that reserve information as much as possible~\citep{li2021beyond, li2022d}. 
However, existing literature has paid less attention to {variable selection} for identifying differences between the two groups except~\citep{ide2007computing, ide2009proximity,hara2015consistent, bonferroni1936teoria}.
The references therein mainly rely on parametric assumptions regarding the data-generating distributions.
In particular, \citet{taguchi2000new} assume target distributions are Gaussian and find important variables such that the difference between mean and covariance among two groups is maximized. 
The works~\citep{ide2007computing, ide2009proximity,hara2015consistent} further model distributions as Gaussian graphical models and detect the difference between distributions in correlation and partial correlation.
However, it is undesirable to restrict the analysis to parametric distributions, especially for real-world applications.
Bonferroni method~\citep{bonferroni1936teoria} has been proposed in the two-sample testing context to compare every single feature using statistical tests to obtain representative variables. Still, it may not perform well when correlations exist between variables.

\noindent{\bf Kernel two-sample tests.} 
A popular approach for non-parametric two-sample testing is based on kernel methods~\citep{scholkopf2002learning}: such tests quantify the difference of probability distributions by measuring the difference in \emph{kernel mean embedding}~\citep{berlinet2011reproducing, muandet2017kernel}, which is also called the maximum mean discrepancy (MMD)~\citep{Gretton12, fukumizu2009kernel, kirchler2020two, schrab2021mmd, schrab2022efficient}.
The follow-up works~\citep{liu2020learning, sutherland2016generative} further improve the performance of kernel-based two-sample tests by selecting kernels that maximize the variance-normalized empirical MMD.
We adopt this idea in our variable selection framework.
However, we observe that using this criterion for variable selection results in a fractional program subject to sparsity and norm constraints, which is highly challenging to solve. 
Hence, we are inspired to consider optimizing the variance-regularized empirical MMD statistic as a surrogate~(see \eqref{Eq:formula:MMD:opt:revision}).

\noindent{\bf Other two-sample tests.} 
Many widely-used frameworks employ classification techniques for two-sample testing (see, e.g., \citep{cheng2022classification, kubler2022automl,kubler2022witness}).
It is worth noting that our approach adopts a distinct framework compared to those references: these aforementioned testing methods may not effectively identify interpretable variables capable of distinguishing between two distributions.
One potential alternative is to employ a classifier based on sparse logistic regression~\citep{bertsimas2021sparse} to construct a two-sample test.
However, this approach may not yield satisfactory performance due to the limited flexibility of the parametric form of the classifier, as we will demonstrate in Section~\ref{Sec:classification}.
Recently, Mueller and Jaakkola~\citep{mueller2015principal} proposed to find the optimal subset of features such that the Wasserstein distance between projected distributions in dimension $d=1$ is maximized.
Later Wang et al.~\citep{wang2021two, wang2020kerneltwosample} modified the projection function as the linear mapping with general dimension $d>1$ and nonlinear mapping, respectively, thus improving the flexibility of dimensionality reduction and power of two-sample testing. 
Nevertheless, these references do not impose sparsity constraints when performing dimensionality reduction. Therefore, they cannot select a subset of variables that differentiate the differences between the two groups.

\section{Background}

We first present some background information about maximum mean discrepancy~(MMD).
It measures the discrepancy between two probability distributions by employing test functions within a reproducing kernel Hilbert space~(RKHS), which has been commonly used in two-sample testing area~\citep{Grettonnips12, Gretton09, Gretton12, liu2020learning, JitkrittumME, Chwialkowski15}.
\begin{definition}[Maximum Mean Discrepancy]\label{Def:MMD}
A kernel function $K:~\bR^D\times\bR^D\to\bR$ is called a positive semi-definite kernel if for any finite set of $n$ samples $\{x_i\}_{i=1}^n$ in $\bR^D$ and $\{c_i\}_{i=1}^n$ in $\bR$, it holds that $\sum_{i\in[n]}\sum_{j\in[n]}c_ic_jK(x_i,x_j)\ge0$.
A positive semi-definite kernel $K$ induces a unique RKHS $\cH$.
Given $\cH$ containing a class of candidate testing functions and two distributions $\mu,\nu$, define the corresponding MMD statistic as
\begin{equation}
\tag*{\QDEF}
\MMD(\mu,\nu; K) \triangleq \sup_{f\in\cH, \|f\|_{\cH}\le 1}~\bigg\{
\bE_{\mu}[f] - \bE_{\nu}[f]
\bigg\}.
\end{equation}
\end{definition}
Leveraging reproducing properties of the RKHS, the MMD statistic can be equivalently written as 
\[
\MMD^2(\mu,\nu; K)=\bE_{x,x'\sim\mu}[K(x,x')] +\bE_{y,y'\sim\nu}[K(y,y')] - 2\bE_{x\sim\mu, y\sim\nu}[K(x,y)],
\]
which enables convenient computation and sample estimation.
When the distributions $\mu$ and $\nu$ are not available, one can formulate an estimate of $\MMD^2(\mu,\nu; K)$ based on samples $\cv x^n$ and $\cv y^n$ using the following statistic~\citep{Gretton12}:
\begin{equation}
    \label{Eq:MMD:empirical}
\widehat{\MMD}^2(\cv x^n,\cv y^n; K) \triangleq
\frac{1}{n(n-1)}\sum_{i\in[n], j\in[n], i\ne j}~H_{i,j},
\end{equation}
with
\begin{equation}
H_{i,j}:=K(x_i, x_j) + K(y_i, y_j) - K(x_i,y_j) - K(y_i,x_j).
\label{Eq:H:i:j}
\end{equation}
The choice of kernel function largely influences the performance of {variable selection} for two-sample tests.
To achieve satisfactory performance, we consider the following types of kernel functions, denoted as $K_z(\cdot,\cdot)$.
Here, the coefficient vector $z=(z^{(s)})_{s\in[D]}$ involved in the kernel functions determines which variables to be selected, which is in the domain set
\begin{equation}
\label{Eq:cZ}
\cZ:=\{z\in\bR^D: \|z\|_2=1, \|z\|_0\le d\}.
\end{equation}
\begin{itemize}
    \item
\textbf{Linear Kernel}: 
For each coordinate $s\in[D]$, we specify the scalar-input kernel $k_s:~\mathbb{R}\times\mathbb{R}\to \bR$ and then construct
\begin{equation}%
K_z(x,y) = \sum_{s\in[D]}z^{(s)}k_s\big(x^{(s)}, y^{(s)}\big).
\label{Eq:linear:kernel}
\end{equation}
Those scalar-input kernels $k_s(\cdot,\cdot), s\in[D]$ defined above are used to compare the difference of distributions among each coordinate, which can be chosen as the Gaussian kernel with certain bandwidth hyper-parameter $\tau^2_s$, i.e., $k_s(x,y) = e^{-(x-y)^2/(2\tau^2_s)}$.
    \item
\textbf{Quadratic Kernel}: 
For each coordinate $s\in[D]$, we specify the scalar-input kernel $k_s:~\mathbb{R}\times\mathbb{R}\to \bR$ and then construct
\begin{equation}
K_z(x,y) = \left(\sum_{s\in[D]}z^{(s)}k_s\big(x^{(s)}, y^{(s)}\big)+c\right)^2.
\label{Eq:quadratic:kernel}
\end{equation}
Here $c\ge 0$ is a bandwidth hyper-parameter of the quadratic kernel, and scalar-input kernels $k_s(\cdot,\cdot), s\in[D]$ can be chosen in the same way as defined in the linear kernel case.
    \item
\textbf{(Isotropic) Gaussian Kernel}: 
We first specify the bandwidth hyper-parameter $\sigma^2>0$ and then construct isotropic Gaussian-type kernel
\begin{equation}
K_z(x,y) = \exp\left( 
-\frac{\sum_{s\in[D]}~\left( 
z^{(s)}(x^{(s)} - y^{(s)})
\right)^2}{2\sigma^2}
\right).
\label{Eq:Gaussian:kernel}
\end{equation}
We use isotropic Gaussian kernel, a common choice in the literature on two-sample testing, primarily because it only involves a bandwidth hyper-parameter that is easy to tune.
\end{itemize}

\section{Formulation}

A natural criterion of variable selection is to pick the coefficient vector $z$ such that the empirical MMD statistic is optimized, i.e., it suffices to solve the formulation
\begin{equation}\label{Eq:MMD:opt:naive}
\max_{z\in\cZ}~\widehat{\MMD}^2(\cv x^n, \cv y^n; K_z).
\end{equation}
To motivate this formulation, we start with an example showcasing the nature of the problem: 
the complexity of the problem depends on the choice of the kernel, and the simplest linear kernel leads to the analytical solution. 
Despite that, the linear kernel is known to have limited testing power, and one usually prefers to use a non-linear kernel, whose solution is not analytical. 
Moreover, the test statistic may require normalization by standard deviation, for which case an analytical solution is not available; all these require further algorithm development of algorithm as presented in Section \ref{Sec:STRS}.

Consider Problem~\eqref{Eq:MMD:opt:naive} when the kernel function $K_z$ is a linear kernel in \eqref{Eq:linear:kernel}.
By straightforward calculation, it can be reformulated as the following linear optimization on the domain set $\cZ$:
\begin{equation}
\max_{z\in\cZ}~z\trans a,
\label{Eq:MMD:opt:naive:linear:ker}
\end{equation}
where for $s\in[D]$, the $s$-th entry of the vector $a\in\mathbb{R}^D$ is
\[
a^{(s)} = \widehat{\MMD}^2\Big(\{x_i^{(s)}\}_{i\in[n]}, \{y_i^{(s)}\}_{i\in[n]}, k_s\Big).%
\]
It is easy to check that the optimal solution to \eqref{Eq:MMD:opt:naive:linear:ker} is obtained by taking the non-zero indexes of $z$ to be the indexes of the $d$ largest absolute values of the vector $a$.
In other words, based on the linear kernel and the criterion~\eqref{Eq:MMD:opt:naive}, our framework selects those variables whose MMD discrepancy between two distributions at the corresponding coordinates is as large as possible.
Since this idea only utilizes the information of marginal distributions of a high-dimensional distribution in each coordinate, the linear kernel has limited testing power in practice.

\subsection{Variance regularized MMD optimization}
Although the idea behind the formulation~\eqref{Eq:MMD:opt:naive} is simple, as pointed out in existing literature~\citep{sutherland2016generative, Grettonnips12, kubler2022automl}, directly optimizing the MMD statistic does not result in a powerful two-sample test in practice.
Inspired by these references, we incorporate the variance statistic of MMD in the formulation to achieve more competitive performance.

Specifically, we pick the sparse selection vector $z$ to achieve the most powerful test.
Since the kernel function leading to the most powerful two-sample test approximately maximizes the MMD testing statistic normalized by its standard deviation~\citep{sutherland2016generative}, 
the natural idea is to pick the selection vector $z$ that solves the following fractional optimization problem:
\begin{equation}
\max_{z\in\cZ}~\frac{\widehat{\MMD}^2(\cv x^n,\cv y^n; K_z)}{\Big(\widehat{\sigma}_{\mathcal{H}_1}^2(\cv x^n, \cv y^n; K_z)\Big)^{1/2}},\label{Eq:fractional:program}
\end{equation}
where $\widehat{\MMD}^2(\cv x^n,\cv y^n; K_z)$ and $\widehat{\sigma}_{\mathcal{H}_1}^2(\cv x^n, \cv y^n; K_z)$ are unbiased empirical estimators of the population testing statistic and the variance of testing statistic under alternative hypothesis $\cH_1:~\mu\ne\nu$, respectively.
For fixed samples $\cv x^n, \cv y^n$ and kernel function $K(\cdot,\cdot)$, by \citep{sutherland2016generative}, the variance estimator
\begin{equation}
\widehat{\sigma}_{\mathcal{H}_1}^2(\cv x^n, \cv y^n; K) = \frac{4}{n^3}\sum_{i\in[n]}\left( 
\sum_{j\in[n]}H_{i,j}
\right)^2 - \frac{4}{n^4}\left( 
\sum_{i\in[n]}\sum_{j\in[n]}H_{i,j}
\right)^2,
\label{Eq:MMD:variance:empirical}
\end{equation}
where $H_{i,j}, i,j\in[n]$ are defined in \eqref{Eq:H:i:j}.
Since the optimization over a fraction in \eqref{Eq:fractional:program} is difficult to handle, we introduce a regularization hyper-parameter $\lambda>0$~(which can be tuned by cross-validation in practice) and propose to solve the new optimization problem instead:
\begin{equation}\label{Eq:formula:MMD:opt:revision}
\begin{aligned}
\max_{z\in\cZ}&\quad \Big\{ 
\widehat{F}(z; \cv x^n,\cv y^n):=\widehat{\MMD}^2(\cv x^n,\cv y^n; K_z) - \lambda\widehat{\sigma}_{\mathcal{H}_1}^2(\cv x^n, \cv y^n; K_z)\Big\}.
\end{aligned}
\end{equation}
The rationale behind problem~\eqref{Eq:formula:MMD:opt:revision} is that, by properly tuning the regularizer $\lambda>0$, we balance the trade-off between maximizing the testing statistic and minimizing its variance, which amounts to approximately optimizing the testing power criteria in \eqref{Eq:fractional:program}.

The hyper-parameter $\lambda$ is tuned using the following cross-validation procedure:
We take a set of candidate choices of $\lambda$, denoted as $\{0.1, 0.5, 1, 2, 5\}$, and split the dataset into $50\%$ training and $50\%$ validation datasets.
For each choice of hyper-parameters, we use the training dataset to obtain the optimal coefficient vector and examine its hold-out performance on the validation dataset, which is quantified as the negative of the p-value for two-sample tests between two collections of samples in the validation dataset. We finally specify the hyper-parameter as the one with the highest value of hold-out performance.

Using the proposed variable selection framework, we present a kernel two-sample test as follows. 
The data points are divided into training and testing datasets. Initially, the training set is utilized to obtain the selection coefficient that optimally identifies the differences between the two groups. Next, a permutation test is performed on the testing data points, projected based on the trained selection coefficient. 
The threshold for this permutation test is calibrated by bootstrapping following~\citep[Section~5]{Gretton12} and \citep{arcones1992bootstrap}.
The detailed algorithm is presented in Algorithm~\ref{Alg:permutation:test}. 
This test is guaranteed to control the type-I error~\citep{good2013permutation} because we evaluate the $p$-value of the test via the permutation approach. 
In the following sections, we discuss how to solve the optimization problem~\eqref{Eq:formula:MMD:opt:revision} with linear and quadratic kernels, respectively.
In the subsequent sections, we develop algorithms for solving the MMD optimization problem and then establish statistical testing power guarantees for our proposed framework.

\begin{center}
\begin{algorithm}[!t]
\caption{
A permutation two-sample test using MMD with variable selection
} 
\begin{algorithmic}[1]\label{Alg:permutation:test}
\REQUIRE
{
cardinality $d$, type-I error threshold $\alpha_{\mathrm{level}}$,
bootstrap size $N_p$,
collected samples $\cv x^n$ and $\cv y^n$.
}
\STATE{ 
Split data as $\cv x^n = \cv x^{\Tr}\cup \cv x^{\Te}$ and $\cv y^n = \cv y^{\Tr} \cup \cv y^{\Te}$.
}
\STATE{
Solve \eqref{Eq:formula:MMD:opt:revision} with input data $(\cv x^{\Tr},\cv y^{\Tr})$ to obtain optimal sparse selection vector $z^*$.
}
\STATE{Compute test statistic $T = \widehat{\MMD}^2(\cv x^{\Te}, \cv y^{\Te}; K_{z^*})$.} 
\STATE{\textbf{for $i = 1, \ldots, N_p$ do}}\hfill\COMMENT \texttt{~Step~4-8: Decide threshold using bootstrap}
\STATE{Shuffle $\cv x^{\Te}\cup \cv y^{\Te}$ to obtain $\cv x^{\Te}_{(i)}$ and $\cv y^{\Te}_{(i)}$.} 
\STATE{Compute test statistic for bootstrap samples $T_i=\widehat{\MMD}^2(\cv x^{\Te}_{(i)}, \cv y^{\Te}_{(i)}; K_{z^*}).$
}
\STATE{\textbf{end for}}
\STATE{$t_{\mathrm{thres}}\leftarrow (1-\alpha_{\mathrm{level}})$-quantile of $\{T_i\}_{i\in[N_p]}$.}
\STATE{Reject $\cH_0$ (i.e., decide the two sample distributions are different) if $T>t_{\mathrm{thres}}$.}
\end{algorithmic}
\end{algorithm}
\end{center}

\subsection{Connections with classification-based testing}\label{Sec:classification}
It is worth mentioning that any method for classification can be applied to two-sample testing: 
Given training samples $\cv x^{\Tr}=\{x_i\}_{i\in[|\cv x^{\Tr}|]}$ and $\cv y^{\Tr}=\{y_i\}_{i\in[|\cv y^{\Tr}|]}$, we formulate the feature-label pairs as $\mathcal{D}_{\Tr}=\{(x_i,0)\}_{i\in[|\cv x^{\Tr}|]}\cup\{(y_i,1)\}_{i\in[|\cv y^{\Tr}|]}$.
One can use any classification method to obtain a classifier $\widehat{f}$ based on training dataset $\mathcal{D}_{\Tr}$.
After that, the testing statistic based on testing samples $\cv x^{\Te}$ and $\cv y^{\Te}$ can be computed as
$T=\frac{1}{|\cv x^{\Te}|}\sum_{x\in \cv x^{\Te}}\widehat{f}(x) - \frac{1}{|\cv y^{\Te}|}\sum_{y\in \cv y^{\Te}}\widehat{f}(y).$
If the testing statistic $T$ is greater than a certain threshold, the null hypothesis $\cH_0$ is rejected, and otherwise, it is accepted.

A notable existing variable selection approach for classification is the \emph{sparse logistic regression}~(SLR)~\citep{bertsimas2021sparse}, which uses the classifier $\widehat{f}(\cdot)=\inp{\cdot}{\beta}$ for some sparse vector $\beta$.
The coefficient vector $\beta$ can be obtained by solving a sparse optimization problem using training dataset $\mathcal{D}_{\Tr}$, and its non-zero entries correspond to the selected variables that distinguish the differences between groups $\cv x^{\Tr}$ and $\cv y^{\Tr}$.
Based on samples $\cv x^{\Te}$ and $\cv y^{\Te}$, SLR formulates the following testing statistic and rejects the null hypothesis if it exceeds a certain threshold:
$
T_{\mathrm{SLR}}=\frac{1}{|\cv x^{\Te}|}\sum_{x\in \cv x^{\Te}}~\beta\trans x - \frac{1}{|\cv y^{\Te}|}\sum_{y\in \cv y^{\Te}}~\beta\trans y.
$
Such an approach assumes a parametric assumption that the data distributions $\mu$ and $\nu$ are \emph{linearly separable} since otherwise, the linear predictor may not achieve satisfactory performance.

In contrast, our proposed method can be viewed as a generalized classification-based testing, which consists of two phases:
\begin{enumerate}
    \item 
At the first phase, we choose a suitable kernel function $K(\cdot,\cdot)$ based on training data $\cv x^{\Tr}$ and $\cv y^{\Tr}$ that depends only on a small group of variables leading to satisfactory two-sample testing performance.
Such a variable selection procedure makes our classification model more interpretable.
     \item
At the second phase, we obtain the classifier (also called the witness function in \citep[Section~2.3]{Gretton12}), denoted as $\widehat{f}$, based on validation data $\cv x^{\Te}$ and $\cv y^{\Te}$:
\begin{equation}
\label{Eq:logit}
\widehat{f}(z) \propto \frac{1}{|\cv x^{\Te}|}\sum_{x\in \cv x^{\Te}}K(x,z)
-
\frac{1}{|\cv y^{\Te}|}\sum_{y\in \cv y^{\Te}}K(y,z)
.
\end{equation}
\end{enumerate}
In comparison with the SLR framework, we replace the linear classifier with the kernel-based classifier, which is a more flexible and powerful choice. 
In the following, we provide an example demonstrating that our proposed framework can successfully select useful variables to distinguish the difference between two groups, while the sparse logistic regression cannot finish this task.
\begin{example}[Example when sparse logistic regression cannot identify variables]\label{Eq:example:1}
Consider the example where $\mu= \mathcal{N}(0, I_D)$ and $\nu = \mathcal{N}(0, \mathrm{diag}((1+\epsilon)^2,1,\ldots,1))$ with $\epsilon>0$.
Here, only the first coordinate can differentiate between $\mu$ and $\nu$.
When using the sparse logistic regression, it is clear that for any $\beta$ satisfying $\|\beta\|_0\le 1$, it holds that
the population version of testing statistic $\mathbb{E}[T_{\mathrm{SLR}}]=0$.
This indicates that sparse logistic regression may not achieve satisfactory performance in hypothesis testing or classification. 
In contrast, consider our proposed MMD framework with the linear kernel.
For any $z$ such that $\|z\|_2=1, \|z\|_0\le 1$, it holds that the population version of the objective in \eqref{Eq:formula:MMD:opt:revision} achieves the unique optimal solution $\widehat{z}$ with $\widehat{z}^{(1)}=1$ if the variance regularization $\lambda$ is selected properly.
Specifically, when $\lambda$ is chosen to be smaller than a constant $\bar{\lambda}>0$,
our proposed MMD framework can always select the true useful variable.\footnote{Here we take $\bar{\lambda} = \frac{\MMD^2(\mathcal{N}(0,1), \mathcal{N}(0,(1+\epsilon)^2); k_1)}{\sigma_{\cH_1}^2(\mathcal{N}(0,1), \mathcal{N}(0,(1+\epsilon)^2); k_1)}$ to satisfy the desired result. Specifically, we provide closed-form expressions on those statistics in the following~(see the proof in Appendix~\ref{Appendix:proof:example:1}):
\begin{equation*}%
\begin{aligned}
A&\triangleq \MMD^2(\mathcal{N}(0,1), \mathcal{N}(0,(1+\epsilon)^2); k_1)
=
\sqrt{\frac{\tau_1^2}{\tau_1^2+2}} + \sqrt{\frac{\tau_1^2}{\tau_1^2+2(1+\epsilon)^2}}-2\sqrt{\frac{\tau_1^2}{\tau_1^2+1 + (1+\epsilon)^2}},\\
B&\triangleq \sigma_{\cH_1}^2(\mathcal{N}(0,1), \mathcal{N}(0,(1+\epsilon)^2); k_1)=4C - 4A^2,\\
C&\triangleq \sqrt{\frac{\tau_1^4}{(\tau_1^2+1)(3+\tau_1^2)}} + \sqrt{\frac{4\tau_1^4}{(\tau_1^2+2)(\tau_1^2+2(1+\epsilon)^2)}}\\
&\qquad-\sqrt{\frac{16\tau_1^4}{
2\tau_1^2+1 + (1+\epsilon)^2 + (1+\tau_1^2)((1+\epsilon)^+\tau_1^2)
}}-\sqrt{\frac{16\tau_1^4}{
(\tau_1^2+1+(1+\epsilon)^2)(\tau_1^2+2(1+\epsilon)^2)
}}\\
&\qquad\qquad\qquad\qquad+\sqrt{\frac{16\tau_1^4}{(\tau_1^2+ (1+\epsilon)^2)(\tau_1^2+ (1+\epsilon)^2 + 2)}}+\sqrt{\frac{\tau_1^4}{(\tau_1^2+(1+\epsilon)^2)(\tau_1^2 + 3(1+\epsilon)^2)}}.
\end{aligned}
\end{equation*}}
\QEG
\end{example}

\section{Algorithm}\label{Sec:STRS}

To prepare for solving the MMD optimization problem~\eqref{Eq:formula:MMD:opt:revision}, we first introduce the following Sparse Trust Region Subproblem~(STRS):
\begin{equation}
\tag{STRS}
\max_{z\in\cZ}~\Big\{ 
z\trans Az + z\trans a
\Big\},\label{Eq:general:MIQP}
\end{equation}
where the set $\cZ$ is defined in \eqref{Eq:cZ} and $(A,a)$ are problem coefficients.
This problem extends the standard Trust Region Subproblem~\citep{conn2000trust}, since we require the decision variable to satisfy an extra sparse constraint.
In Section~\ref{Sec:linear:MMD}, we will show that Problem~\eqref{Eq:formula:MMD:opt:revision} can be reformulated as a special case of \eqref{Eq:general:MIQP} for linear kernels.
For generic kernels, the optimization procedure for solving~\eqref{Eq:formula:MMD:opt:revision} involves solving the subproblem~\eqref{Eq:general:MIQP}.
Unfortunately, the set $\cZ=\{z\in\bR^D:~\|z\|_2=1, \|z\|_0\le d\}$ in \eqref{Eq:general:MIQP} involves $\ell_0$-norm constraint, which typically leads to a mixed-integer program reformulation that is NP-hard to solve.
This motivates us to provide efficient optimization algorithms to tackle this challenge in Section~\ref{Sec:challenge}.

Without loss of generality, we assume $A\succeq0$ in \eqref{Eq:general:MIQP}, since otherwise, we can re-write the problem as 
$
\max_{z\in\cZ}~\Big\{ 
z\trans (A - \lambda_{\min}(A)I_D)z + z\trans a
\Big\} + \lambda_{\min}(A),
$
where the shifted matrix $A - \lambda_{\min}(A)I_D\succeq0$, where $\lambda_{\min}$ denotes the smallest eigenvalue of a matrix.
It is worth mentioning that the problem~\eqref{Eq:general:MIQP} reduces to sparse PCA formulation when the coefficient vector $a=0$ (that is, the linear term is zero), which has been studied extensively in the literature~\citep{berk2019certifiably, gally2016computing, moghaddam2005spectral, li2020exact}.
However, the study for general vector $a$ for the problem~\eqref{Eq:general:MIQP} is new.
In Section~\ref{Sec:challenge}, we discuss the exact and approximation algorithms for solving \eqref{Eq:general:MIQP} with generic data matrix $A$ and vector $a$.

\subsection{MMD optimization with different kernels}
\label{Sec:linear:MMD}
In the following, we provide detailed algorithms for solving the MMD optimization problem~\eqref{Eq:formula:MMD:opt:revision} for various kernels considered in \eqref{Eq:linear:kernel}-\eqref{Eq:Gaussian:kernel}.
\subsubsection{Linear kernel}\label{Sec:MMD:linear}
For linear kernel defined in \eqref{Eq:linear:kernel}, one can verify that $H_{i,j}\in \mathbb R$ defined in \eqref{Eq:H:i:j} can be written as a linear function in terms of $z$:
\begin{align*}
H_{i,j}&=\sum_{s\in[D]}~z^{(s)}\Big[k_s(x_i^{(s)}, x_j^{(s)})+k_s(y_i^{(s)}, y_j^{(s)})-k_s(x_i^{(s)}, y_j^{(s)})-k_s(y_i^{(s)}, x_j^{(s)})\Big]
=z\trans h_{i,j},
\end{align*}
where we denote the $D$-dimensional vector 
\begin{equation}
h_{i,j}= \Big\{k_s(x_i^{(s)}, x_j^{(s)})+k_s(y_i^{(s)}, y_j^{(s)})-k_s(x_i^{(s)}, y_j^{(s)})-k_s(y_i^{(s)}, x_j^{(s)})\Big\}_{s\in[D]}.
\end{equation}
Since the empirical MMD estimator $\widehat{\MMD}^2(\cv x^n,\cv y^n; K_z)$ is a linear combination of $\{H_{i,j}\}_{i,j}$ and the empirical variance estimator $\hat{\sigma}_{\mathcal{H}_1}^2(\cv x^n, \cv y^n; K_z)$ is a quadratic function in terms of $\{H_{i,j}\}_{i,j}$, it is clear that the MMD optimization problem~\eqref{Eq:formula:MMD:opt:revision} can be reformulated as the mixed-integer quadratic optimization problem~\eqref{Eq:general:MIQP},
where the data matrix $A\in\bR^{D\times D}$ and vector $a\in\bR^D$ have the following expressions:
\begin{align*}
A^{(s_1,s_2)}&=\frac{4\lambda}{n^3}\sum_{i\in[n]}\left( 
\sum_{j\in[n]}h_{i,j}^{(s_1)}
\right)\left( 
\sum_{j\in[n]}h_{i,j}^{(s_2)}
\right)-\frac{4\lambda}{n^4}
\left( 
\sum_{i,j\in[n]}h_{i,j}^{(s_1)}
\right)
\left( 
\sum_{i,j\in[n]}h_{i,j}^{(s_1)}
\right),\quad \forall s_1,s_2\in[D],
\\
a&=\frac{1}{n(n-1)}\sum_{i\in[n], j\in[n], i\ne j}h_{i,j}.
\end{align*}
Therefore, one can query either the exact or approximation algorithm to solve problem~\eqref{Eq:formula:MMD:opt:revision} with strong optimization guarantees for this linear kernel case. 
In the following remark, we discuss under which conditions will linear kernel MMD may or may not achieve satisfactory performance on the variable selection task.
\begin{remark}[Limitation of Linear Kernel]\label{Remark:linear}
Under the linear kernel choice, it can be shown that the population MMD statistic becomes
$
\MMD^2(\mu,\nu; K_z) = \sum_{s\in[D]}z^{(s)}\MMD^2(\text{Proj}_{s\#}\mu, \text{Proj}_{s\#}\nu; k_s),
$
where $\text{Proj}_{s\#}\mu, \text{Proj}_{s\#}\nu$ are the $s$-th marginal distributions of $\mu,\nu$, respectively.
In other words, the selection coefficient $z$ aims to find a direction to identify the difference between marginal distributions of $\mu$ and $\nu$.
However, under the case where marginal distributions of $\mu$ and $\nu$ are the same, the linear kernel MMD does not have enough power to find informative variables to distinguish those two distributions.\QEG
\end{remark}

\subsubsection{Other kernel choices}\label{Sec:MMD:general}
For other kernel choices, such as the quadratic kernel in \eqref{Eq:quadratic:kernel} and Gaussian kernel in \eqref{Eq:Gaussian:kernel}, the objective for MMD optimization is a nonlinear non-concave function with respect to $z$. This, together with the sparse constraint of the domain set $\cZ$, makes this type of problem very challenging to solve.
In this part, we provide a heuristic algorithm that incorporates simulated annealing~(SA)~\citep{bertsimas1993simulated} and STRS that tries to find a feasible solution of \eqref{Eq:formula:MMD:opt:revision} with high solution quality.
Such a heuristic can also be naturally extended for generic kernel choices.

Here, we outline our SA and STRS-based heuristics.
For notational simplicity, we denote the objective of \eqref{Eq:formula:MMD:opt:revision} as $F(z)$ instead.
Our proposed algorithm is an iterative method that generates a trajectory of feasible solutions $z_1,\ldots,z_{i_{\max}}$.
At the iteration point $z_i$, we generate a candidate solution $\tilde{z}_i$ by optimizing a second-order approximation of the objective $F(z)$ with quadratic penalty regularization around $z_i$:
\begin{equation}
\tilde{z}_i=\argmax_{z\in\cZ}~\left\{ 
F(z_i) + \nabla F(z_i)\trans(z-z_i) + \frac{1}{2}(z-z_i)\trans\nabla^2F(z_i)(z-z_i) - \frac{\tau_i}{2}\|z - z_i\|_2^2
\right\},\label{Eq:update:z_t:can}
\end{equation}
where $\tau_i$ denotes the quadratic regularization value.
Such a problem is a special case of \eqref{Eq:general:MIQP}, where the data matrix $A\in\bR^{D\times D}$ and vector $a\in\bR^D$ have the following expressions:
\begin{align*}
A&=\frac{1}{2}\nabla^2F(z_i)-\frac{\tau_i}{2}I_D,\quad 
a=\nabla^2F(z_i)z_i-\tau_iz_i + \nabla F(z_i).
\end{align*}
Hence, Problem~\eqref{Eq:update:z_t:can} can be solved by querying the exact or approximation algorithm described in Section~\ref{Sec:STRS}.
Let $\Delta_i=F(\tilde{z}_i) - F(z_i)$ denote the residual value for moving from $z_i$ to $\tilde{z}_i$.
The central idea of SA is always to accept moves with positive residual values while not forbidding moves with negative residual values.
Specifically, we assign a certain temperature $\texttt{Tem}$, and update $z_{i+1}$ as $\tilde{z}_i$ according to the probability %
\[
p_i=\left\{
\begin{aligned}
1,&\quad\text{if }\Delta_i\ge 0\\
e^{\Delta_i/\texttt{Tem}},&\quad\text{if }\Delta_i<0.
\end{aligned}
\right.
\]
If the candidate solution $\tilde{z}_i$ is not accepted, we update $z_{i+1}$ as ${z}_i$.
The temperature parameter $\texttt{Tem}$ is a critical hyper-parameter in this algorithm. We assign an initial value of $\texttt{Tem}$ and iteratively decrease it such that in the last iterations, the moves with worse objective values are less and less likely to be accepted.
See our detailed algorithm procedure in Algorithm~\ref{alg:Eq:obj:nonconcave}.
\begin{algorithm}[!t]
   \caption{Heuristic algorithm for solving \eqref{Eq:formula:MMD:opt:revision} with generic kernel}
   \label{alg:Eq:obj:nonconcave}
\begin{algorithmic}[1]
   \STATE {\bfseries Input:} Max iterations $i_{\max}$, initial guess $z_1$, initial temperature $\texttt{Tem}$, cooling parameter $\alpha$, and a set of regularization values $\mathcal{G}$
   \FOR{$i = 1,\ldots, i_{\max}-1$}
   \STATE{Randomly pick the regularization value $\tau_i$ from $\mathcal{G}$.}
   \STATE{
   Obtain $\tilde{z}_i$ by solving a STRS in \eqref{Eq:update:z_t:can}.
   } 
   \STATE{
   Compute residual level $\Delta_i=F(\tilde{z}_i) - F(z_i)$ and probability $p_i=e^{\Delta_i/\texttt{Tem}}$
   }
   \IF{$\mathrm{rand}(0,1)<p_i$}
   \STATE{$z_{i+1} = \tilde{z}_i$}
   \ELSE
   \STATE{$z_{i+1} = z_i$}
   \ENDIF
   \STATE{$\texttt{Tem}=\alpha\cdot \texttt{Tem}$}
   \ENDFOR
   \STATE{\textbf{Return} $z_{i_{\max}}$}
\end{algorithmic}
\end{algorithm}

Finally, we add remarks regarding the tractability and flexibility of quadratic and Gaussian kernels.
\begin{remark}[Quadratic kernel]
For quadratic kernel defined in \eqref{Eq:quadratic:kernel}, it can be shown that the population MMD is given by $\MMD^2(\mu,\nu; K_z) = z\trans \mathcal{A}(\mu,\nu)z + z\trans\mathcal{T}(\mu,\nu),$
where $\mathcal{A}(\mu,\nu)$ is a $\bR^{D\times D}$-valued mapping such that
\begin{multline*}
(\mathcal{A}(\mu,\nu))^{(s_1,s_2)} = 
\bE_{x,x'\sim\mu}[k_{s_1}(x^{(s_1)},x'^{(s_1)})k_{s_2}(x^{(s_2)},x'^{(s_2)})]
\\+
\bE_{y,y'\sim\nu}[
k_{s_1}(y^{(s_1)},y'^{(s_1)})k_{s_2}(y^{(s_2)},y'^{(s_2)})
]
-2
\bE_{x\sim\mu,y\sim\nu}[k_{s_1}(x^{(s_1)},y^{(s_1)})k_{s_2}(x^{(s_2)},y^{(s_2)})],
\end{multline*}
and $\mathcal{T}(\mu,\nu)$ is a $\bR^D$-valued mapping such that $(\mathcal{T}(\mu,\nu))^{(s)} = 2c\MMD^2(\text{Proj}_{s\#}\mu, \text{Proj}_{s\#}\nu; k_s).$
Given two multivariate distributions, the quadratic MMD aims to find a direction $z$ to distinguish the difference in each coordinate and the correlation between the two coordinates the most.
Compared with the linear MMD, which only identifies the difference in each coordinate, the quadratic MMD is a more flexible choice.
However, it can be shown that the objective in \eqref{Eq:formula:MMD:opt:revision} with the quadratic kernel is a $4$-th order non-concave monomial with respect to $z$, which is computationally intractable to optimize.
In practical experiments, we use the heuristic algorithm in Algorithm~\ref{alg:Eq:obj:nonconcave} to obtain a reasonably high-quality solution.\QEG
\end{remark}
\begin{remark}[Gaussian kernel]
One can also re-write the population testing statistic for the Gaussian kernel defined in \eqref{Eq:Gaussian:kernel}.
For notational simplicity, let $K(x,y)=\exp\left(-\frac{\|x-y\|_2^2}{2\sigma^2}\right), \forall x,y\in\bR^d$ be a standard Gaussian kernel with low-dimensional data, and define $z_{\#}\nu$ as a $d$-dimensional distribution such that
\[
z_{\#}\nu=\big( 
z^{(s)}x^{(s)}
\big)_{s\in\mathrm{supp}(z)},\quad \text{where }x\sim\nu.
\]
With these notations, it can be shown that the population MMD statistic becomes $\MMD^2(\mu,\nu; K_z) = \MMD^2(z_{\#}\mu, z_{\#}\nu; K).$
Since the kernel $K$ satisfies the universal property~\citep{micchelli2006universal}, our proposed Gaussian kernel distinguishes the difference between $\mu$ and $\nu$ as long as there exists a $d$-size sub-group of coordinates of $\mu$ and $\nu$ that cause the difference.
Compared with linear and quadratic kernels, the Gaussian kernel is a more flexible choice.
Unfortunately, the computation burden of the Gaussian kernel is heavier than the other two simple kernels because the objective in \eqref{Eq:formula:MMD:opt:revision} can be viewed as a non-concave $\infty$-degree monomial with respect to $z$, whereas the second-order approximation scheme in \eqref{Eq:update:z_t:can} may not provide reliable performance for optimization.\QEG
\end{remark}

\subsection{Tackling the %
challenge of solving \texorpdfstring{\eqref{Eq:general:MIQP}}{}}
\label{Sec:challenge}

There are two challenges in solving~\eqref{Eq:general:MIQP} in particular for large-scale problems.
First, since the objective function is non-concave in $z$, it is difficult to develop exact algorithms directly for solving \eqref{Eq:general:MIQP}. Instead, 
we provide a mixed-integer \emph{convex} programming reformulation, which motivates us to develop exact algorithms in Section~\ref{Sec:MISDP}.
Second, this problem is NP-hard even if the coefficient vector $a=0$, as pointed out in~\citep{magdon2017np}.
When the problem is large-scale, we provide approximation algorithms with provable performance guarantees.

\subsubsection{Exact mixed-integer SDP~(MISDP) reformulation}\label{Sec:MISDP}
We first provide an exact MISDP reformulation of \eqref{Eq:general:MIQP}.
When the coefficient vector $a=0$, similar reformulation results have been developed in the sparse PCA literature~\citep{li2020exact, bertsimas2022solving}.
However, such a reformulation for $a\ne 0$ is new in the literature.
For notational simplicity, we define the following block matrix of size $(D+1)\times(D+1)$:
\[
\tilde{A} = \begin{pmatrix}
0&\frac{1}{2}a\trans\\
\frac{1}{2}a&A
\end{pmatrix}.
\]
\begin{theorem}[MISDP Reformulation of {\eqref{Eq:general:MIQP}}]
\label{Thm:Eq:MIQP:quad}
Problem~\eqref{Eq:general:MIQP} can be equivalently formulated as the following MISDP
\begin{subequations}
\label{Eq:MISDP:quad}
\begin{align}
\max_{
\substack{
Z\in\bS^+_{D+1}, q\in\cQ\\
}}&\quad \inp{\tilde{A}}{Z}\\
\mbox{s.t.}&\quad 
Z^{(i,i)}\le q^{(i)},\quad i\in[D],\label{Eq:MISDP:quad:b}\\
&\quad Z^{(0,0)}=1, \text{Tr}(Z)=2,\label{Eq:MISDP:quad:c}
\end{align}
\end{subequations}
where the set 
\begin{equation}
\label{Eq:cQ}
\cQ = \left\{ 
q\in\bF^D:~\sum_{k\in[D]}q^{(k)}\le d
\right\},
\end{equation}
and we assume the indices of $Z, \tilde{A}\in\bS_{D+1}^+$ are both over $[0:D]\times[0:D]$.
The continuous relaxation value of \eqref{Eq:MISDP:quad} equals $w_{\mathrm{rel}} = \max_{z:~\|z\|_2=1}~\left\{
z\trans Az + z\trans t
\right\}.$
\end{theorem}

The proof idea of Theorem~\ref{Thm:Eq:MIQP:quad} is to express the problem~\eqref{Eq:general:MIQP} as a \emph{rank-$1$ constrained SDP} problem.
Leveraging well-known results on rank-constrained optimization (see, e.g., \citep{polik2007survey, dey2020convexifications, li2022exactness}), one can remove the rank constraint without changing the optimal value of the original SDP problem.
Although \eqref{Eq:MISDP:quad} is equivalent to \eqref{Eq:general:MIQP}, the fact that its continuous relaxation value is equal to $w_{\mathrm{rel}}$ suggests that it may be a weak formulation.
Inspired from \citep{li2020exact, bertsimas2022solving}, we propose the additional two valid inequalities to strengthen the formulation~\eqref{Eq:MISDP:quad} in Corollary~\ref{Corollary:strong:MISDP}.

\begin{corollary}[Stronger MISDP Reformulation of {\eqref{Eq:general:MIQP}}]\label{Corollary:strong:MISDP}
The problem~\eqref{Eq:general:MIQP} reduces to the following stronger MISDP formulation:
\begin{subequations}\label{Eq:MISDP:quad:strong:sum}
\begin{align}\label{Eq:MISDP:quad:strong}
\max_{
\substack{
Z\in\bS^+_{D+1}, q\in\cQ\\
}}&\quad \inp{\tilde{A}}{Z}\\
\mbox{s.t.}&\quad 
\eqref{Eq:MISDP:quad:c},
\sum_{j\in[D]}(Z^{(i,j)})^2\le Z^{(i,i)}q^{(i)}, 
\left( 
\sum_{j\in[D]}|Z^{(i,j)}|
\right)^2\le dZ^{(i,i)}q^{(i)}, \quad \forall i\in[D].\label{Eq:MISDP:quad:strong:d}%
\end{align}
\end{subequations}

\end{corollary}

It is worth noting that two distinct references~\citep{li2020exact, bertsimas2022solving} have independently introduced two valid inequalities to enhance the performance of solving the sparse PCA problem, which is a special instance of \eqref{Eq:general:MIQP} for $a=0$.
However, one of the valid inequalities in \eqref{Eq:MISDP:quad:strong:d} proposed in \citep{bertsimas2022solving} is dominated by a valid inequality proposed in \citep{li2020exact}. In contrast, the other valid inequality has been proposed simultaneously in these two references.
This motivates us to incorporate two valid inequalities from~\citep{li2020exact} into our formulation, as outlined in Corollary~\ref{Corollary:strong:MISDP}.
On the one hand, the resulting formulation~\eqref{Eq:MISDP:quad:strong:sum} can be directly solved via some exact MISDP solvers such as YALMIP~\citep{lofberg2004yalmip}.
On the other hand, it enables us to develop a customized, exact algorithm to solve this formulation based on Benders decomposition since the binary vector $q$ can be separated from other decision variables.

To develop the exact algorithm, we first reformulate the problem~\eqref{Eq:MISDP:quad:strong:sum} as a max-min saddle point problem so that it can be solved based on the outer approximation technique~\citep{fletcher1994solving,bonami2008algorithmic}.
\begin{theorem}[Saddle Point Reformulation of {\eqref{Eq:MISDP:quad:strong:sum}}]
\label{Thm:ref:Eq:MISDP:quad}
Problem~\eqref{Eq:MISDP:quad:strong:sum} shares the same optimal value as the following problem:
\begin{equation}
\begin{aligned}
\max_{q\in\cQ}~\left\{ 
f(q)\triangleq\max_{Z\in\bS_{D+1}^+}~\left\{ 
\inp{\tilde{A}}{Z}: s.t. ~\eqref{Eq:MISDP:quad:strong:d}
\right\}
\right\}.
\end{aligned}
\label{Eq:one:dim:q}
\end{equation}
Here the function $f(q)$ is concave in $q$ over the domain $\overline{\cQ} := \mathrm{conv}(\cQ)$, and equivalently, is the optimal value to the following problem:
\begin{equation}\label{Eq:optval:fq}
\begin{aligned}
\min_{\substack{
\lambda,\lambda_0,\nu_1,\nu_2,
\Lambda, \beta,\mu,W_1,W_2
}}&\quad \lambda_0 + 2\lambda + 
q\trans\left[ 
\frac{d}{2}(\nu_1-\nu_2) + \frac{1}{2}(\mu - \diag(\Lambda))
\right]\\
\mbox{s.t.}&\quad \begin{pmatrix}
-\lambda_0&\quad\frac{1}{2}t\trans\\
\frac{1}{2}t&\quad A-\lambda I_D + W_1-W_2 + \Lambda + \frac{1}{2}\diag(\nu_1 + \nu_2)
\end{pmatrix}\preceq0,\quad  W_1 + W_2 - \diag(\beta)\le 0,\\
&\quad \sum_j(\Lambda^{(i,j)})^2\le (\mu^{(i)})^2,\quad (\beta^{(i)})^2 + (\nu_2^{(i)})^2\le (\nu_1^{(i)})^2,\quad i\in[D],\\
&\quad 
\nu_1,\beta, \mu\in\bR_+^D,\quad W_1,W_2\in\bR_+^{D\times D},\quad \lambda,\lambda_0\in\bR,
\quad \nu_2\in\bR^D, \Lambda\in\bR^{D\times D}. %
\end{aligned}
\end{equation}
For fixed $q$, the sup-gradient of $f$ with respect to $q$ can be computed as
\[
\partial f(q) = \frac{d}{2}(\nu_1^* - \nu_2^*) + \frac{1}{2}(\mu^* - \diag(\Lambda^*)),
\]
where $(\nu_1^*, \nu_2^*, \mu^*,\Lambda^*)$ is an optimal solution to the optimization problem above.
\end{theorem}

By Theorem~\ref{Thm:ref:Eq:MISDP:quad}, we find that given a reference direction $\hat{q}$, 
$f(q)\le \bar{f}(q; \hat{q})\triangleq f(\hat{q}) + g_{\hat{q}}\trans(q - \hat{q}),$
where $g_{\hat{q}}$ is a sup-gradient of $f$ at $\hat{q}$.
    \begin{algorithm}[!t]
   \caption{Exact Algorithm for solving \eqref{Eq:general:MIQP}}
   \label{alg:exact:quad}
\begin{algorithmic}[1]
   \STATE {\bfseries Input:} Max iterations $i_{\max}$, initial guess $q_1$, tolerance $\epsilon$.
   \FOR{$i = 1,\ldots, i_{\max}-1$}
   \STATE{Compute $q_{i+1}$ as the optimal solution from
   \[
   \max_{q\in\cQ}~\left\{ \bar{f}^i(q)\triangleq\min_{1\le j\le i}~
\bar{f}(q; q_j)\right\}
   \]}
   \STATE{Compute $f(q_{i+1})$ and $g_{q_{i+1}}\in\partial f(q_{i+1})$}
   \STATE{{\bf Break} if $f(q_{i+1}) - \bar{f}^i(q_{i+1})<\epsilon$} 
   \ENDFOR
   \STATE{\textbf{Return} $q_{i_{\max}}$}
\end{algorithmic}
\end{algorithm}
Based on this observation, we use the common outer-approximation technique, which is widely used for general mixed-integer nonlinear programs~\citep{fletcher1994solving,bonami2008algorithmic}, to solve the problem: at iterations $i=1,2,\ldots,i_{\max}-1$,
we maximize and refine a piecewise linear upper-bound of $f(q)$: $\bar{f}^i(q) = \min_{1\le j\le i}~
\bar{f}(q; q_j).$
The algorithm is summarized in Algorithm~\ref{alg:exact:quad}. 

By the reference~\citep{fletcher1994solving}, it can be shown that this algorithm yields a non-increasing sequence of overestimators $\{\bar{f}^i(q)\}_{i=1}^{i_{\max}}$, which converge to the optimal value of $f(q)$ within a finite
number of iterations $i_{\max}\le \binom{D}{1}+\cdots+\binom{D}{d}$.

\subsubsection{Convex relaxation algorithm}
Inspired by Theorem~\ref{Thm:ref:Eq:MISDP:quad}, a natural idea of approximately solving the problem~\eqref{Eq:MISDP:quad} is to consider the following problem, in which we replace the nonconvex constraint $q\in\cQ$ by a set of linear constraints, which forms its convex hull:
\begin{equation}
\max_{q\in\overline{\cQ}}~f(q),\quad \text{where }\overline{\cQ}= \mathrm{conv}(\cQ) = \left\{q\in[0,1]^D:~\sum_iq^{(i)}\le d\right\}.
\label{Eq:cvx:relax}
\end{equation}
Since the problem~\eqref{Eq:cvx:relax} is a convex program, it can be solved in polynomial time.
Besides, one can obtain a high-quality feasible solution to the problem~\eqref{Eq:MISDP:quad}, using a greedy rounding scheme:
We first solve \eqref{Eq:cvx:relax} to obtain its optimal solution $\tilde{q}$, and then project it onto $\overline{\cQ}$ to obtain $q$.
Next, we solve the problem~\eqref{Eq:MISDP:quad} by fixing the variable $q$ and optimizing $Z$ only.

In the following theorem, we provide the approximation ratio regarding the SDP formulation above.
The proof adopts similar techniques as in \citep[Theorem~5]{li2020exact}, but we extend the analysis for inhomogeneous quadratic maximization formulation.

\begin{theorem}[Approximation Gap for Convex Relaxation]
\label{Thm:optval:relaxation:ratio}
Denote by \textsf{optval}\eqref{Eq:cvx:relax} and \textsf{optval}\eqref{Eq:MISDP:quad} the optimal values of problem~\eqref{Eq:cvx:relax} and \eqref{Eq:MISDP:quad}, respectively.
Then, it holds that $    \textsf{optval}\eqref{Eq:MISDP:quad} \le \textsf{optval}\eqref{Eq:cvx:relax}\le \|a\|_2+ 
\min\bigg\{ 
D/d\cdot\textsf{optval}\eqref{Eq:MISDP:quad}, d\cdot\textsf{optval}\eqref{Eq:MISDP:quad}-\min_k|a^{(k)}|
\bigg\}.$
\end{theorem}

Despite the convexity of problem~\eqref{Eq:cvx:relax}, it is challenging to solve, especially for high-dimensional scenarios.
References~\citep{bertsimas2022solving, li2020exact} solved a special case of problem~\eqref{Eq:cvx:relax} when $a=0$ based on the interior point method~(see, e.g., \citep{alizadeh1995interior, boyd2004convex, todd2001semidefinite}).
Unfortunately, since the constraint set of \eqref{Eq:cvx:relax} involves the intersection of a semidefinite cone and a large number of second-order cones, re-writing it as a standard conic program and using off-the-shelf solvers to solve this problem spends lots of time.
The work \citep{ma2013alternating} designed a novel variable-splitting technique and proposed a first-order Alternating Direction Method of Multipliers~\citep{boyd2011distributed}~(ADMM) algorithm to solve a special convex relaxation of sparse PCA.
Unlike this reference that only considers the simplest convex relaxation of sparse PCA without adding strong inequalities, our problem~\eqref{Eq:cvx:relax} has considerably complicated constraints.

Inspired by the reference~\citep{ma2013alternating}, we use a similar variable-splitting technique to split the second-order conic constraints and all the other constraints in two blocks of variables and then propose an ADMM algorithm to optimize the augmented Lagrangian function.
The advantage is that each subproblem in iteration update involves only second-order conic constraints or other constraints that are easy to deal with, which results in considerably fast computational speed. 
We provide a detailed implementation of the proposed algorithm for solving~\eqref{Eq:cvx:relax} in Appendix~\ref{Appendix:ADMM}.

\subsubsection{Truncation algorithms with tighter approximation gap}

Unfortunately, the SDP relaxation formulation is still challenging to solve for extremely high-dimension scenarios, which motivates us to develop the following computationally cheap truncation approximation algorithms.
Compared with the approximation ratio of relaxed SDP formulation in Theorem~\ref{Thm:optval:relaxation:ratio} (i.e., $\min(D/d, d)+\mathcal{O}(1)$), the ratio for our proposed algorithm is tighter (i.e., $\min(D/d, \sqrt{d})+\mathcal{O}(1)$).
First, we introduce the definition of a normalized sparse truncation operator.
\begin{definition}[Normalized Sparse Truncation]\label{Def:nor:sparse}
For a vector $z\in\bR^D$ and an integer $d\in[D]$, we say $\bar{z}$ is a $d$-sparse truncation of $z$ if
\[
\bar{z}^{(i)}=\left\{ 
\begin{aligned}
z^{(i)},&\quad\text{if $|z^{(i)}|$ is one of the $d$ largest (in absolute value) entries in $z$}\\
0,&\quad\text{otherwise}.
\end{aligned}
\right.
\]
Besides, the vector $\widehat{z} = \bar{z}/ \|\bar{z}\|_2$ is said to be the normalized $d$-sparse truncation of $z$.
\end{definition}
Now, we introduce the following two truncation algorithms:\\
\noindent{\bf Truncation Algorithm~(I):}
Let $A^{(:,i)}$ be the $i$-th column of $A$ for $i\in[D]$, and denote by $\hat{z}_i$ the normalized $d$-sparse truncation of $A^{(:,i)}$.
Then return the estimated optimal solution as the best over all $\hat{z}_i$'s and $e_i$'s for $i\in[D]$, where $e_i$ denotes the $i$-th standard basis vector.\\
\noindent{\bf Truncation Algorithm~(II):}
Relax the $\ell_0$-norm constraint in \eqref{Eq:general:MIQP} and solve the trust region problem $\max_{z:~\|z\|_2\le 1}~\{z\trans Az + z\trans a\}$ to obtain the optimal primal solution $v$.
Then, return the estimated optimal solution $z$ as the normalized $d$-sparse truncation of $v$.

We summarize the approximation ratios of these two truncation algorithms in Theorem~\ref{Thm:ratio:truncation}.
Its proof technique is adopted from \citep{chan2016approximability}.
The difference is that the authors consider the approximation ratio under the case $a=0$, while we adopt the structure of in-homogeneous quadratic function maximization to extend the case for the general coefficient vector $a$.
\begin{theorem}[Approximation Gap for Truncation Algorithm]\label{Thm:ratio:truncation}
\begin{enumerate}
    \item\label{Thm:ratio:truncation:I}
Truncation Algorithm~$\mathrm{(I)}$ returns a feasible solution of \eqref{Eq:general:MIQP} with objective value $V_{\mathrm{(I)}}$ such that $\textsf{optval}\eqref{Eq:general:MIQP}
\ge 
V_{\mathrm{(I)}}\ge \frac{1}{\sqrt{d}}\textsf{optval}\eqref{Eq:general:MIQP} - 2\|a\|_{(d+1)}.$
    \item\label{Thm:ratio:truncation:II}
Truncation Algorithm~$\mathrm{(II)}$ returns a feasible solution of \eqref{Eq:general:MIQP} with objective value $V_{\mathrm{(II)}}$ such that $\textsf{optval}\eqref{Eq:general:MIQP}
\ge 
V_{\mathrm{(II)}}\ge \frac{d}{D}\cdot\textsf{optval}\eqref{Eq:general:MIQP} - \frac{d}{D}\cdot\|a\|_2 - \left( 
1 + \sqrt{\frac{d}{D}}
\right)\cdot\|a\|_{(d)}.$
\end{enumerate}
\end{theorem}

We return the best over the output from Truncation Algorithm~(I) and (II) as the estimated optimal solution.
By Theorem~\ref{Thm:ratio:truncation}, we find the returned solution approximates the optimal solution up to approximation ratio $\min(D/d,\sqrt{d})+\mathcal{O}(1)$.
It has been shown in Chan et al.~\citep{chan2016approximability} that it is NP-hard to implement any algorithm with \emph{constant} approximation ratio.
Therefore, it is of research interest to explore polynomial-time approximation algorithms with approximation ratio that has milder dependence on $D$ and $d$.
Instead of trying this direction, in the next subsection, we propose another approximation algorithm such that, though NP-hard to solve, it achieves a higher approximation ratio.

\subsubsection{Approximation algorithm via convex integer programming}
In this part, we propose an approximation algorithm based on convex integer programming.
We first consider the following $\ell_1$-norm relaxation of the problem~\eqref{Eq:general:MIQP}, which plays a key role in developing our algorithm:
\begin{equation}
\max~\left\{ 
z\trans Az + z\trans a:~\|z\|_2\le 1, \|z\|_1\le \sqrt{d}
\right\}.
\label{Eq:general:MIQP:noncvx:relax}
\end{equation}
This problem is a relaxation of problem~\eqref{Eq:general:MIQP} because constraints $\|z\|_2\le 1, \|z\|_0\le d$ imply $\|z\|_1\le \sqrt{d}$.
Following the similar proof technique as in \citep[Theorem~1]{dey2022using}, we show that solving this new problem results in a constant approximation ratio.
The difference is that the authors therein only consider the special case of \eqref{Eq:general:MIQP} with $a=0$, while we extend their analysis for general inhomogeneous quadratic objective functions.

\begin{theorem}[Approximation Gap for $\ell_1$-Norm Relaxation]\label{Pro:relax:noncvx}
Let $\rho = 1+\sqrt{d/(d+1)}$. Then we have that $\textsf{optval}\eqref{Eq:general:MIQP}\le \textsf{optval}\eqref{Eq:general:MIQP:noncvx:relax}\le 
\rho^2\textsf{optval}\eqref{Eq:general:MIQP}.$
\end{theorem}
Although the problem~\eqref{Eq:general:MIQP:noncvx:relax} is a relaxation of \eqref{Eq:general:MIQP}, it is still intractable to solve due to the non-concavity of the objective function (recall that $A\succeq0$).
We adopt techniques from \citep[Section~2.2]{dey2022using} to derive a further convex integer program that serves as a further relaxation of the relaxation problem~\eqref{Eq:general:MIQP:noncvx:relax}.
Before proceeding, we define the following notations.
For $i\in[D]$, denote by $(\lambda_i, v_i)$ the $i$-th eigen-pair of the matrix $A$, denote 
$\theta_i:=\max\{z\trans v_i:~\|z\|_2\le 1, \|z\|_1\le \sqrt{d}\},$
and let $\gamma_i^{[-N:N]}$ be the set of partition points of the domain $[-\theta_i,\theta_i]$, i.e.,
$\gamma_i^j = \frac{j}{N}\theta_i,\quad j=-N,\ldots,N.$
Let $\lambda_0\in\mathbb{R}_+$ be a fixed number such that $\lambda_0\le \textsf{optval}\eqref{Eq:general:MIQP}$.
\begin{proposition}[Convex Integer Programming Relaxation of {\eqref{Eq:general:MIQP:noncvx:relax}}]\label{Proposition:CVX:IP:reformulate}
\begin{subequations}\label{Eq:CVXIP}
Consider the convex integer program:
\begin{equation}
\begin{array}{ll}
\mbox{Maximize}&\quad \lambda_0 + \sum_{i:~\lambda_i>\lambda_0}(\lambda_i-\lambda_0)\xi_i - s
\end{array}
\end{equation}
that is subject to the following constraints:
\begin{align*}
&\left\{ 
\begin{aligned}
g_i&=z\trans v_i,\\
|g_i|&\le \theta_i,
\end{aligned}\quad i\in[D],
\right.
\qquad\qquad\qquad\qquad\qquad\qquad
\left\{ 
\begin{aligned}
\sum_{i\in[D]}y_i&\le\sqrt{d},\\
y_i&\ge \left| z^{(i)} \right|, i\in[D],
\end{aligned}
\right.\\
\\ 
&\left\{ 
\begin{aligned}
g_i&=\sum_{j\in[-N,N]}\gamma_i^j\eta_i^j,\\
\xi_i&=\sum_{j\in[-N,N]}(\gamma_i^j)^2\eta_i^j,\\ 
\eta_i^{[-N,N]}&\in\mathrm{SOS\text{-}2},
\end{aligned}\quad i\in\{i:~\lambda_i>\lambda_0\},
\right.\qquad 
\left\{ 
\begin{aligned}
\sum_{i\in[D]}(z^{(i)})^2&\le1,\\
\sum_{i:~\lambda_i>\lambda_0}\left(\xi_i - \frac{\theta_i^2}{4N^2}\right)
+\sum_{i:~\lambda_i\le \lambda_0}g_i^2
&\le 1,\\
\sum_{i:~\lambda_i<\lambda_0}-(\lambda_i-\lambda_0)g_i^2 - z\trans a\le s,
\end{aligned}
\right.
\end{align*}
with $\mathrm{SOS\text{-}2}$ denoting the special ordered set of type-2~\citep{beale1970special},
and involves the following decision variables: 
\[
\begin{multlined}
\{g_i\}_{i=1}^D\in\mathbb{R}^{D},\quad \{\xi_i\}_{i\in\{i:~\lambda_i>\lambda_0\}}\in\mathbb{R}^{|\{i:~\lambda_i>\lambda_0\}|},\quad  
\{\eta_i^j\}_{i\in\{i:~\lambda_i>\lambda_0\}, j\in[-N:N]}\in\mathbb{R}^{(2N+1)|\{i:~\lambda_i>\lambda_0\}|},\\ 
\{y_i\}_{i=1}^D\in\mathbb{R}^{D},\quad s\in\mathbb{R}, \quad z\in\mathbb{R}^{D}.
\end{multlined}
\]
\end{subequations}
This problem is a relaxation of the $\ell_1$-norm relaxed problem~\eqref{Eq:general:MIQP:noncvx:relax}.
Besides, it holds that $\textsf{optval}\eqref{Eq:general:MIQP}
\le 
\mathsf{optval}\eqref{Eq:CVXIP}\le \rho^2\textsf{optval}\eqref{Eq:general:MIQP} + \frac{1}{4N^2}\sum_{i:~\lambda_i>\lambda_0}(\lambda_i - \lambda_0)\theta_i^2,$
where the constant $\rho>0$ is defined in Theorem~\ref{Pro:relax:noncvx}.
\end{proposition}

The convex integer program \eqref{Eq:CVXIP} seems appealing because it only requires solving $O((2N)^{|\{i:~\lambda_i>\lambda_0\}|})$ number of finite-dimensional convex optimization problems to obtain its optimal solution.
In practice, the choice of $\lambda_0$ influences the computational traceability of problem~\eqref{Eq:CVXIP}, and the choice of $N$ influences the quality of the approximation.
We follow the heuristic described in \citep[Section~4.3.1]{dey2022using} to select $\lambda_0$ and $N$.
After solving the problem~\eqref{Eq:CVXIP}, one obtains the decision variable $z$ that may not be feasible in $\cZ$. Then, one can use the greedy rounding scheme to project $z$ onto $\cZ$ to obtain a primal feasible solution.

Finally, we acknowledge a recent concurrent study~\citep{mitsuzawa2023variable} that employed $\ell_1$-norm relaxation as a heuristic for MMD-based variable selection. In contrast to this literature, which presented a heuristic approach, our work is the first study that provides both $\ell_1$ relaxation methodology and its theoretical performance guarantees.

\section{Statistical Properties}\label{Sec:test:power}
In this section, we provide statistical performance guarantees for the variance-regularized MMD statistics in \eqref{Eq:formula:MMD:opt:revision}, specialized for our proposed kernels in \eqref{Eq:linear:kernel}, \eqref{Eq:quadratic:kernel}, and \eqref{Eq:Gaussian:kernel}.
In addition, we develop the guarantees for a generic kernel in Appendix~\ref{Sec:extension:test:power}.

First, we derive concentration properties to show that the empirical estimators $S^2(\cv x^n,\cv y^n; K_z)$ and $\widehat{\sigma}^2_{\mathcal{H}_1}(\cv x^n, \cv y^n; K_z)$ uniformly converge to their population version as the sample size $n$ increases. Such a property is useful for showing the testing consistency and the rate of testing power of our MMD framework.
\begin{proposition}[Non-asymptotic Concentration Properties]\label{Thm:con:prop:special}
For Gaussian kernel in \eqref{Eq:Gaussian:kernel}, we assume the sample space $\Omega\subseteq \{x\in\mathbb{R}^D:~\|x\|_{\infty}\le R\}$ for some constant $R>0$. 
With probability at least $1-\delta$,
(i) the bias approximation error can be bounded as
\begin{align*}
\sup_{z\in\cZ}
&\Big|
S^2(\cv x^n,\cv y^n; K_z)
-\MMD^2(\mu,\nu;K_z)
\Big|
\le \epsilon_{n,\delta}^1=
\left\{ 
\begin{aligned}
&\widetilde{\mathcal{O}}(dn^{-1/2}),&&\quad \text{for linear kernel},\\
&\widetilde{\mathcal{O}}(d^{3/2}n^{-1/2}),&&\quad \text{for quadratic kernel},\\
&\widetilde{\mathcal{O}}(d^{1/2}n^{-1/2}),&&\quad \text{for Gaussian kernel},
\end{aligned}
\right. 
\end{align*}
where $\widetilde{\mathcal{O}}(\cdot)$ hides a multiplicative factor $(\log n + \log(D/d) + \log\frac{1}{\delta})^{1/2}$ and other constants that are independent to parameters $D,d,n$.
(ii) and the variance approximation error can be bounded as
\[
\begin{aligned}
&\sup_{z\in\cZ}
\Big|
\widehat{\sigma}^2_{\mathcal{H}_1}(\cv x^n, \cv y^n; K_z)
-
\mathbb{E}_{\cv x^n\sim\mu,\cv y^n\sim\nu}[\widehat{\sigma}^2_{\mathcal{H}_1}(\cv x^n, \cv y^n; K_z)]
\Big|
\le \epsilon_{n,\delta}^2
\end{aligned}
\]
where 
$\epsilon_{n,\delta}^2$ shares the same order of decaying rate as $\epsilon_{n,\delta}^1$.
\end{proposition}

Based on the concentration properties above, we are ready to derive the asymptotic distribution of the testing statistic.
Furthermore, we impose specific assumptions on data distributions when defining linear, quadratic, or Gaussian kernels. The technical assumption regarding the Gaussian kernel is the most lenient, followed by the quadratic kernel, which is less lenient, whereas the assumption for the linear kernel is the most restrictive, reflecting a gradual decrease in flexibility across these kernels.
\begin{assumption}[Structure of Data]\label{Assumption:data}
Assume the following conditions hold:
\begin{enumerate}
    \item
For linear kernel, there exists $s^*\in[D]$ such that $\mathrm{Proj}_{s^\#}\mu\ne \mathrm{Proj}_{s^\#}\nu$.
    \item
For quadratic kernel, there exists $s^*\in[D]$ such that $\mathrm{Proj}_{s^\#}\mu\ne \mathrm{Proj}_{s^\#}\nu$ or $(\mathcal{A}(\mu,\nu))^{(s^*,s^*)}>0$.
    \item
For Gaussian kernel, the sample space $\Omega\subseteq \{x\in\mathbb{R}^D:~\|x\|_{\infty}\le R\}$ for some constant $R>0$, and there exists $S\subseteq[D]$ with $|S|\le d$ such that $\mathrm{Proj}_{S\#}\mu\ne \mathrm{Proj}_{S\#}\nu$.
\end{enumerate}
\end{assumption}
\begin{proposition}[Asymptotic Distribution of Testing Statistic]\label{Thm:asymptotic}
Under Assumption~\ref{Assumption:data}, suppose the hyper-parameter $\lambda>0$ is properly selected (see its detailed range in xxx), and 
let $\hat{z}_{\mathrm{Tr}}$ be the obtained sparse coefficient by solving~\eqref{Eq:formula:MMD:opt:revision} from training dataset $(\cv x^{\Tr}, \cv y^{\Tr})$ with $|\cv x^{\Tr}|=|\cv y^{\Tr}|=n_{\Tr}$, 
$T_{n_{\Te}}$ be the testing statistic evaluated on testing dataset $(\cv x^{\Te}, \cv y^{\Te})$ with $|\cv x^{\Te}|=|\cv y^{\Te}|=n_{\Te}$.
Then, it holds that 
\begin{enumerate}
    \item 
Under alternative hypothesis $\cH_1:~\mu\ne\nu$, for training sample size
\begin{equation}
n_{\Tr}=\left\{ 
\begin{aligned}
\Omega\left(d^2\log (D/d)\right),\quad &\text{ for linear kernel},\\
\Omega\left(d^4\log(D/d)\right),\quad &\text{ for quadratic kernel},\\
\Omega\left(d\log(D/d)\right),\quad &\text{ for Gaussian kernel},
\end{aligned}
\right.\label{Eq:n:Tr}
\end{equation}
it holds that $\mathbb{E}[T_{n_{\Te}}\mid\cH_1]>0$ with high probability.
    \item\label{Thm:asymptotic:II}
Under null hypothesis $\cH_0:~\mu=\nu$, it holds that $n_{\Te}T_n\to \sum_i\sigma_i(Z_i^2-2)$, with $\sigma_i$ denoting the eigenvalues of the $\mu$-covariance operator of the centered kernel; 
under alternative hypothesis $\cH_1:~\mu\ne\nu$, it holds that  $\sqrt{n_{\Te}}(T_n - \mathbb{E}[T_{n_{\Te}}\mid\cH_1])\to \mathcal{N}(0, \sigma^2_{\mathcal{H}_1}(\mu,\nu;K_{\hat{z}_{\mathrm{Tr}}}))$.
\end{enumerate}
\end{proposition}
It is worth mentioning that the order of training sample size in \eqref{Eq:n:Tr} depends only on factors $\text{poly}(d)$ and $\log(D/d)$, indicating the statistical guarantees of our proposed variable selection framework do not suffer from the curse of dimensionality. 
Based on Proposition~\ref{Thm:asymptotic}, we finally present the consistency and rate of testing power of our framework.
\begin{theorem}[Consistency]\label{Thm:consistent}
Under the same assumption as in Proposition~\ref{Thm:asymptotic}, specify the training sample size $n_{\Tr}$ as in \eqref{Eq:n:Tr}.
Let $\alpha\in(0,1)$ denote the level of two-sample test and take $\tau$ as the $(1-\alpha)$-quantile of the limiting distribution $\sum_i\sigma_i(Z_i^2-2)$ defined in Proposition~\ref{Thm:asymptotic}\ref{Thm:asymptotic:II}, and let the threshold of the test be $t_{\mathrm{thres}}:=\frac{\tau}{n_{\Te}}.$
As a consequence, $\bP\big(T_{n_{\Te}}>t_{\mathrm{thres}}\mid\cH_0\big)\to\alpha$ and $ \bP\big(T_{n_{\Te}}\le t_{\mathrm{thres}}\mid\cH_1\big)\to 0.$
\end{theorem}
\begin{theorem}[Testing Power]\label{Thm:non:power}
Under the same assumption as in Proposition~\ref{Thm:asymptotic}, and the same choices of training sample size and threshold as in Theorem~\ref{Thm:consistent}, we additionally assume $\mathbb{E}[|T_1|^3\mid\cH_1]<\infty$.
When the testing sample size $n_{\Te}$ is sufficiently large so that 
\[
t_{\mathrm{thres}} + \frac{\Phi^{-1}(1-n_{\Te}^{-1/2})}{\sqrt{n_{\Te}}}
=\frac{\tau}{n_{\Te}} + 
\sqrt{
\frac{\ln\frac{n_{\Te}}{2\pi} - \ln\ln\frac{n_{\Te}}{2\pi}}
{n_{\Te}}
}(1+o(1))
\]
is sufficiently small, where $\Phi(x)=\frac{1}{\sqrt{2\pi}}\int_{-\infty}^x e^{-t^2/2}\diff t$ denotes the error function, it holds that $\bP\big(T_{n_{\Te}}>t_{\mathrm{thres}}\mid\cH_0\big)\le \alpha + \cO(n_{\Te}^{-1/2})$ and $\bP\big(T_{n_{\Te}}\le t_{\mathrm{thres}}\mid\cH_1\big)\le \cO(n_{\Te}^{-1/2})$,
where $\cO(\cdot)$ hides constant related to parameters $\mathbb{E}[|T_1|^3\mid\cH_1]$ and $\sigma_{\cH_1}^2:=\mathbb{E}_{\cv x^n\sim\mu,\cv y^n\sim\nu}[\widehat{\sigma}^2_{\mathcal{H}_1}(\cv x^n, \cv y^n; K_{\hat{z}_{\mathrm{Tr}}})]$.
\end{theorem}
Theorem~\ref{Thm:non:power} indicates that 
under alternative hypothesis $\mu\ne\nu$, 
as long as the testing sample size $n_{\Te}$ is sufficiently large, and the training sample size is chosen according to \eqref{Eq:n:Tr},
the testing power approaches $1$ with error rate $\cO(n_{\Te}^{-1/2})$.

\section{Simulated numerical examples}

We first consider synthesized data sets to examine the performance of our proposed variable selection framework.
We consider the following four cases:
\begin{enumerate}
    \item(Gaussian Mean Shift): 
Data distribution $\mu=\mathcal{N}(0, \Sigma)$ with the covariance matrix $\Sigma^{(s_1,s_2)} = \rho^{|s_1-s_2|}$ for some correlation level $\rho\in(0,1)$.
Data distribution $\nu=\mathcal{N}(\mu,\Sigma)$ with the mean vector $\mu^{(s)}=\tau/s, \forall s\in[\dtrue]$ for some scalar $\tau>0$ and otherwise $\mu^{(s)}=0$.
   \item(Gaussian Covariance Shift):
Data distribution $\mu=\mathcal{N}(0, \Sigma)$ with $\Sigma$ specified the same as in Part~(I), and $\nu=\mathcal{N}(0, \tilde{\Sigma})$, with $\tilde{\Sigma}^{(s_1,s_2)}=\tau{\Sigma}^{(s_1,s_2)}, \forall s_1,s_2\in[\dtrue]$ for some scalar $\tau>1$ and otherwise $\tilde{\Sigma}^{(s_1,s_2)}={\Sigma}^{(s_1,s_2)}$.
    \item(Gaussian versus Laplacian):
Data distribution $\mu=\mathcal{N}(0, I_D)$.
The first $\dtrue$ coordinates of $\nu$ are independent Laplace distributions with zero mean and standard deviation $0.8$.
The remaining coordinates of $\nu_Y$ are independent Gaussian distributions $\mathcal{N}(0,1)$.
    \item(Gaussian versus Gaussian Mixture):
Data distribution $\mu=\mathcal{N}(0, I_D)$.
The first $d$ coordinates of $\nu$ are Gaussian mixture distribution $\frac{1}{2}\mathcal{N}(-\mu, I_{\dtrue}) + \frac{1}{2}\mathcal{N}(\mu,I_{\dtrue})$ while the remaining coordinates are independent Gaussian distributions $\mathcal{N}(0,1)$.
Here the mean vector $\mu^{(s)} = \tau/s, \forall s\in[\dtrue]$ for some scalar $\tau>0$.
\end{enumerate}
Unless otherwise specified, we take hyper-parameters in Case~(I) as $\tau=1,\rho=0.5$, in Case~(II) as $\tau=2,\rho=0.5$, and in Case~(IV) as $\tau=2$.
We quantify the performance in terms of hypothesis testing metrics rather than the prediction accuracy metrics used in the literature.
Besides, we also measure the quality of variable selection using \emph{false-discovery proportion}~(FDP) and the \emph{non-discovery proportion}~(NDP) defined in~\citep{bajwa2012group}:
\begin{equation}
\text{FDP}(I) = \frac{|I\setminus I^*|}{|I|},\quad \text{NDP}(I) = \frac{|I^*\setminus I|}{|I^*|},
\label{Eq:FDP:NDP}
\end{equation}
where $I^*$ denotes the ground truth feature set and $I$ denotes the set obtained by  {variable selection} algorithms.
The smaller the FDP or NDP is, the better performance the obtained feature set has.

For simplicity of implementation, we chose the bandwidth hyper-parameter $\tau_s^2$ for the kernel $k_s(x,y)$ using the median heuristic, i.e., we specify it as the median among all pairwise distances for data points in the $s$-th coordinate.
Similarly, we take bandwidth $\sigma^2$ of Gaussian kernel as the median among all pairwise distances for data points~(over all coordinates), and bandwidth of quadratic kernel as $c=\sqrt{\sigma^2}$.
Users are also recommended to tune those hyper-parameters based on the cross-validation technique, 
which tends to return near-optimal hyper-parameter choices for large sample sizes.
\subsection{Numerical performance for solving~\texorpdfstring{\eqref{Eq:general:MIQP}}{}}

\begin{figure}[!ht]
    \centering
    \if\paperversion1
    \includegraphics[width=0.45\textwidth]{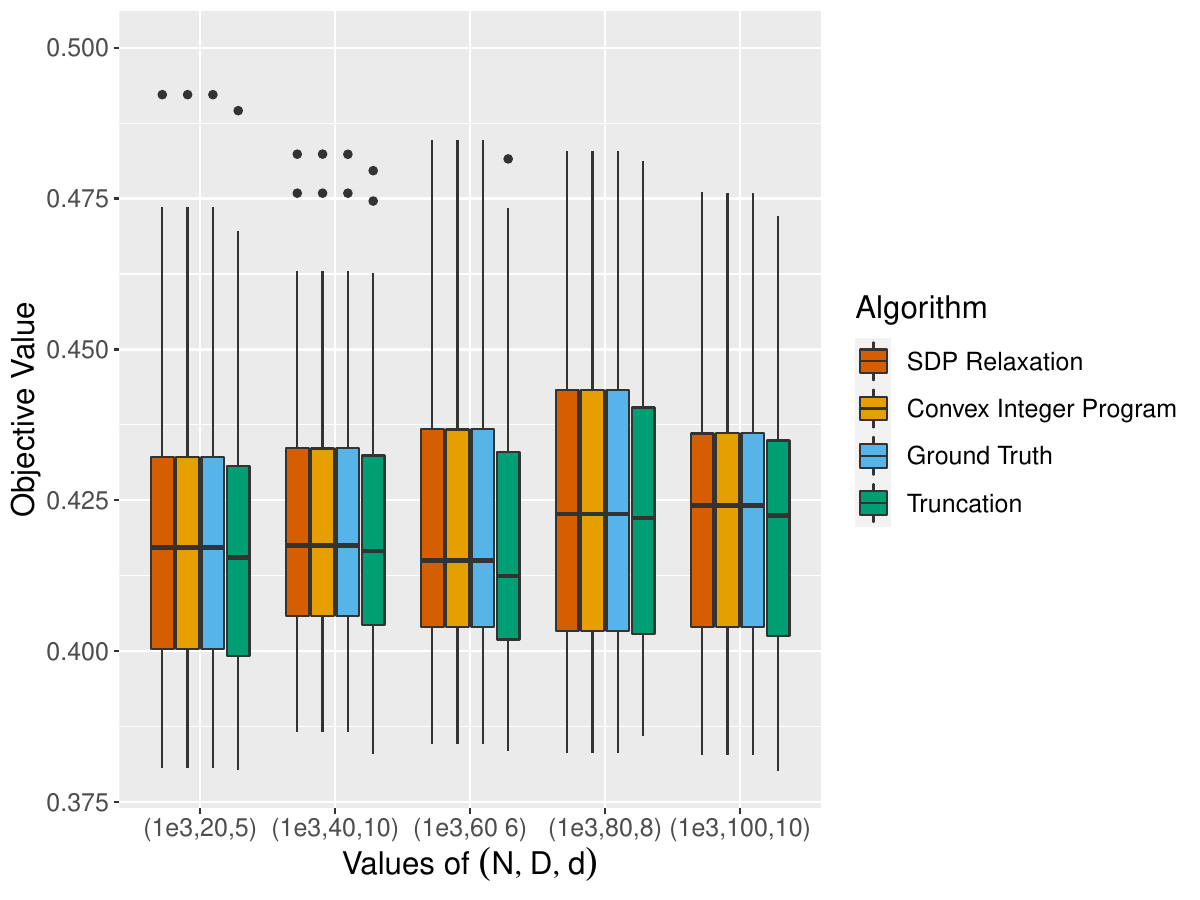}
     \includegraphics[width=0.45\textwidth]{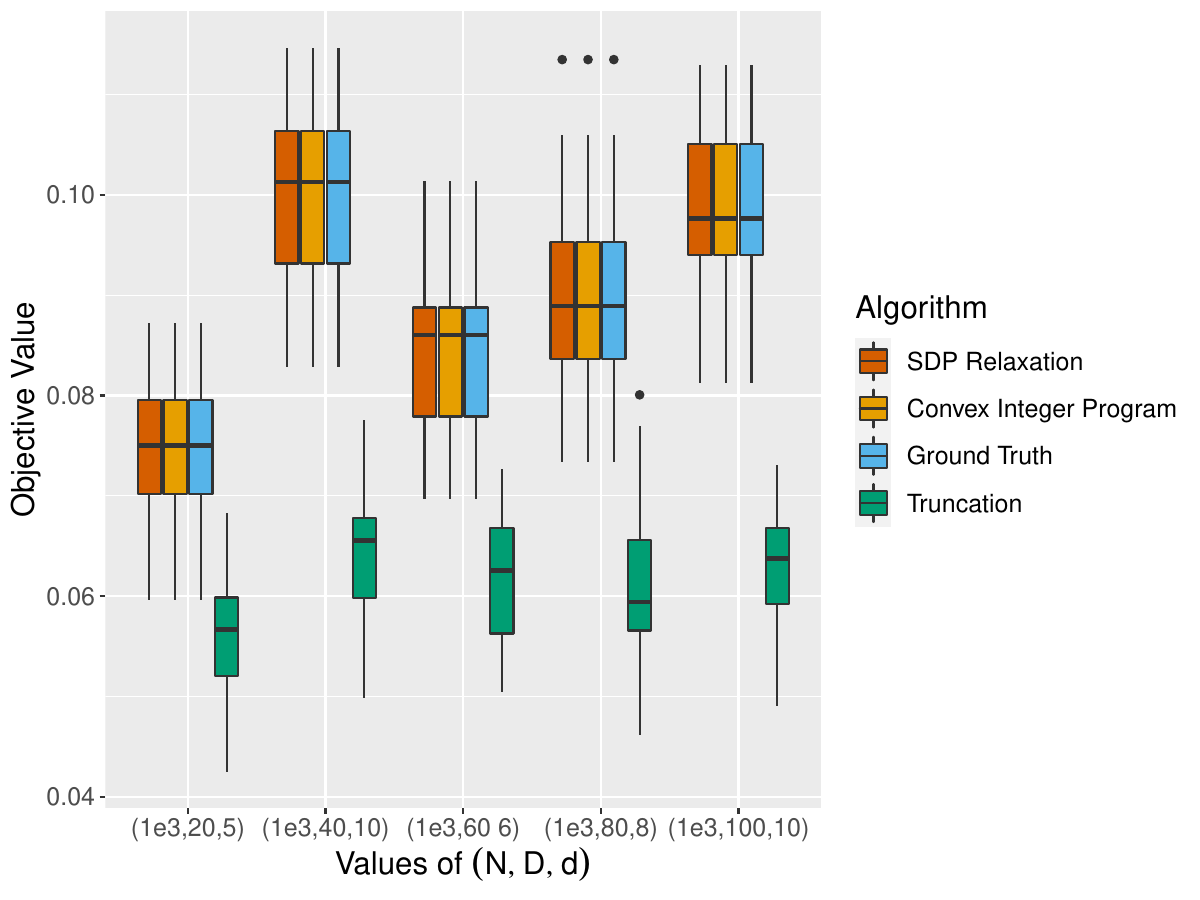}
      \includegraphics[width=0.45\textwidth]{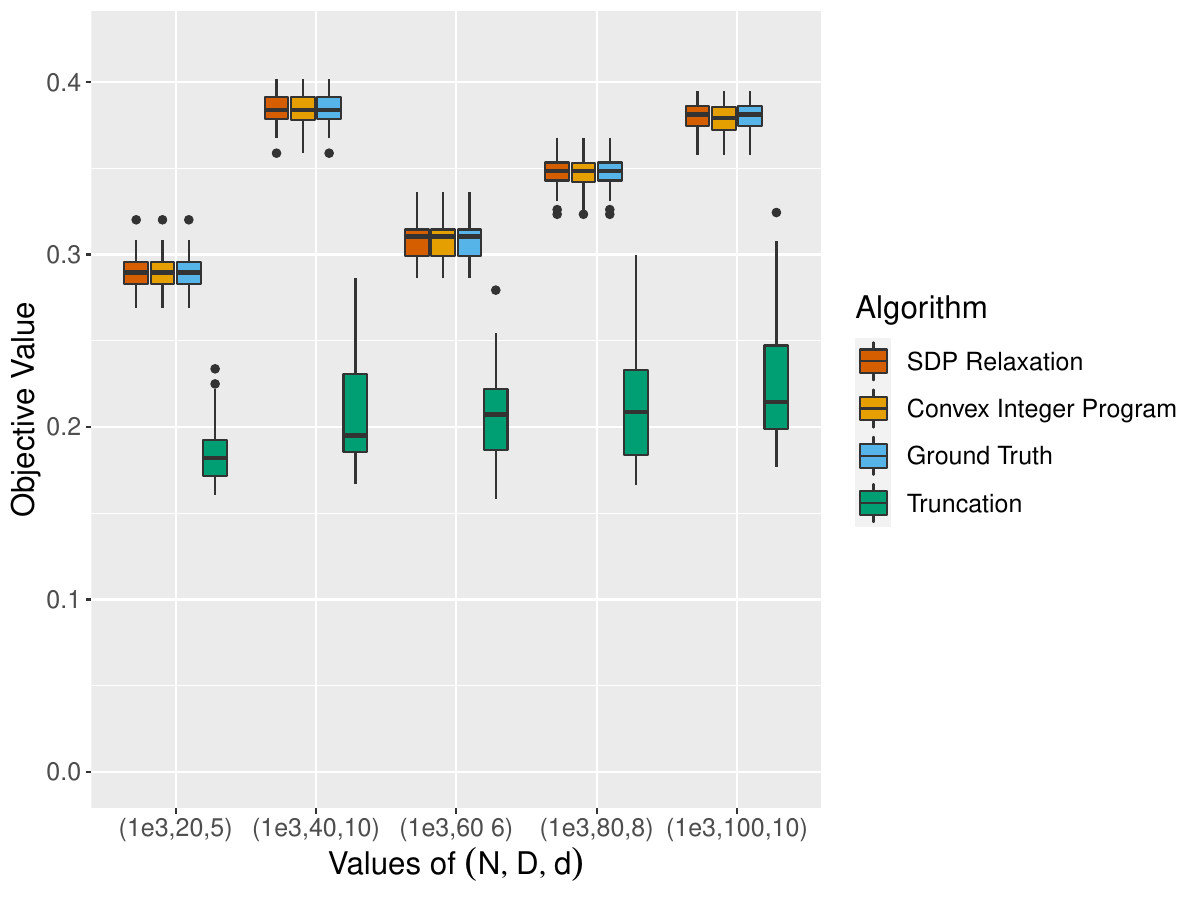}
       \includegraphics[width=0.45\textwidth]{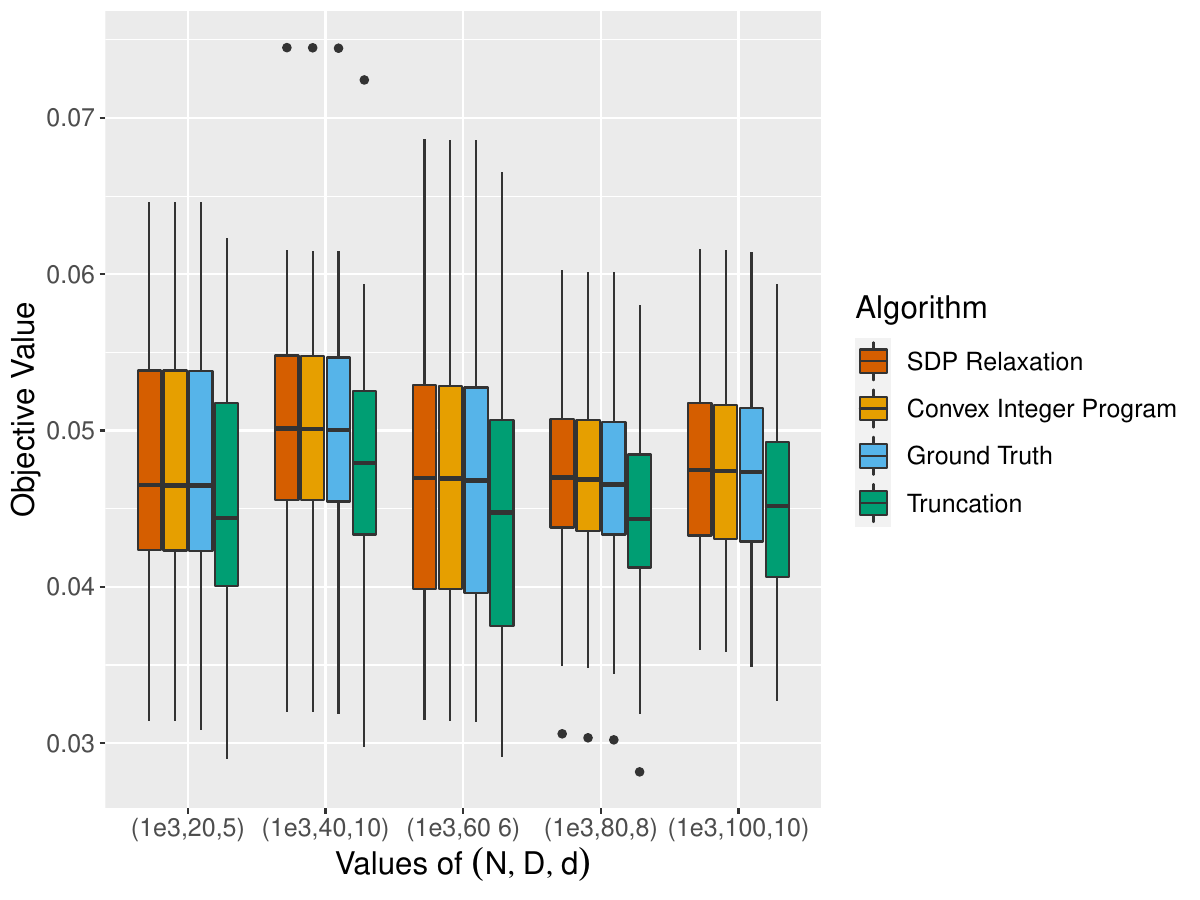}
       \fi
       \if\paperversion2
\includegraphics[width=0.35\textwidth]{Testing_performance_1008_case1}
     \includegraphics[width=0.35\textwidth]{Testing_performance_1008_case2}
      \includegraphics[width=0.35\textwidth]{Testing_performance_1008_case3}
       \includegraphics[width=0.35\textwidth]{Testing_performance_1008_case4}
       \fi
    \caption{Box plots on the performance of various approximation algorithms for solving \eqref{Eq:general:MIQP}. The $x$-axis corresponds to various choices of $(N,D,d)$, and $y$-axis corresponds to the estimated objective value of \eqref{Eq:general:MIQP}. Plots from top to bottom correspond to four types of synthetic datasets.}
    \label{fig:box:performance:app}
\end{figure}
We first examine the numerical performance of various approximation algorithms for solving \eqref{Eq:general:MIQP}, by taking the MMD optimization with linear kernel (see the reformulation in Section~\ref{Sec:MMD:linear}) as a numerical example.
For each of the four synthetic datasets, we try various choices of parameters $(N,D,d)$ from the set
$\{(\textsf{1e3},20,5), (\textsf{1e3}, 40, 10), (\textsf{1e3},60,6), (\textsf{1e3},80,8), (\textsf{1e3},100,10)\}.$
We also specify different hyper-parameters $\lambda\in\{0.8,0.7,0.6,0.5\}$ when using these four different datasets, respectively.
Since those approximation algorithms may return a solution that is infeasible to the constraint $\cZ$, we estimate the corresponding feasible solution by performing the normalized sparse truncation (see Definition~\ref{Def:nor:sparse}).

Figure~\ref{fig:box:performance:app} reports the objective value obtained from the feasible solution based on those approximation algorithms, where the error bars are generated using $100$ independent trials.
The larger the objective value is, the better performance the designed algorithm has.
From the plot, we can see that semidefinite relaxation and convex integer programming algorithms perform nearly optimally compared with the ground truth. In contrast, the performance truncation algorithm is slightly worse compared with those approaches.
Table~\ref{Tab:time:var:app} reports the corresponding computation time of those approximation algorithms, from which we identify that the truncation algorithm has the fastest computational speed. In contrast, SDP relaxation has the slowest speed.
Since the convex integer programming algorithm has satisfactory performance with relatively fast computational speed, we recommend using this approximation algorithm when solving~\eqref{Eq:general:MIQP}.

\begin{table}[!ht]
	\setlength{\tabcolsep}{2pt} %
	\renewcommand{\arraystretch}{1.2} %
	\caption{Averaged computational time of various approximation algorithms for solving \eqref{Eq:general:MIQP}.}\label{Tab:time:var:app}
	\label{table1}
	\begin{center}
			\begin{tabular}{c |r r r|c| c| c }
				\hline
				\multirow{2}{*}{Data Type} & \multicolumn{3}{c|}{Parameters} & \multicolumn{3}{c}{Averaged Computational Time(s) of Approximation Algorithms}  \\
    \cline{5-7}
				& $n$ & $D$ & $d$ & \multicolumn{1}{c|}{Truncation Algorithm} & \multicolumn{1}{c|}{SDP Relaxation} &\multicolumn{1}{c}{Convex Integer Programming} \\
				\hline
				Gaussian &   \textsf{1e3}& 20 & 5 & 2.13e-3 & 1.18 & 1.25e-1 \\
				Mean Shift & \textsf{1e3}& 40 & 10 & 6.08e-3 & 2.55 & 2.29e-1\\
                            & \textsf{1e3}& 60 & 6 & 1.30e-2 & 4.80 & 4.47e-1\\
                            & \textsf{1e3}& 80 & 8 & 3.08e-2 & 6.19 & 6.87e-1\\
                             & \textsf{1e3}& 100 & 10 & 6.87e-2 & 9.47 & 8.77e-1\\
				\hline
    \hline
				Gaussian &   \textsf{1e3}& 20 & 5 & 2.33e-3 & 1.18 & 1.24e-1 \\
				Covariance Shift & \textsf{1e3}& 40 & 10 & 5.86e-3 & 2.57 & 3.11e-1\\
                            & \textsf{1e3}& 60 & 6 & 1.32e-2 & 4.80 & 4.02e-1\\
                            & \textsf{1e3}& 80 & 8 & 3.07e-2 & 6.46 & 9.38e-1\\
                             & \textsf{1e3}& 100 & 10 & 6.76e-2 & 9.73 & 1.23\\
				\hline
    \hline
				Gaussian &   \textsf{1e3}& 20 & 5 & 2.28e-3 & 1.29 & 1.69e-1 \\
				versus Laplacian & \textsf{1e3}& 40 & 10 & 6.39e-3 & 2.85 & 5.65e-1\\
                            & \textsf{1e3}& 60 & 6 & 1.44e-2 & 5.20 & 5.14e-1\\
                            & \textsf{1e3}& 80 & 8 & 3.31e-2 & 6.79 & 1.15\\
                             & \textsf{1e3}& 100 & 10 & 6.95e-2 & 1.02e+1 & 2.10\\
				\hline
    \hline
				Gaussian &   \textsf{1e3}& 20 & 5 & 2.17e-3 & 1.16 & 1.17e-1 \\
				versus  & \textsf{1e3}& 40 & 10 & 6.38e-3 & 2.57 & 2.14e-1\\
                    Gaussian Mixture        & \textsf{1e3}& 60 & 6 & 1.42e-2 & 4.72 & 3.94e-1\\
                            & \textsf{1e3}& 80 & 8 & 3.31e-2 & 6.07 & 5.91e-1\\
                             & \textsf{1e3}& 100 & 10 & 7.25e-2 & 9.44 & 8.67e-1\\
				\hline
			\end{tabular}
	\end{center}
\end{table}

\subsection{Impact of sample size and data dimension}

In this subsection, we compare the performance of {variable selection} based on the following approaches:
(I) Linear kernel MMD;
(II) Quadratic kernel MMD;
(III) Gaussian kernel MMD; 
(IV) Sparse Logistic Regression: a framework that trains the projection vector with $\ell_0$-norm constraint to minimize the logistic loss~\cite{bertsimas2021sparse};
and
(V) Projected Wasserstein: {variable selection} framework using projected Wasserstein distance~\citep{mueller2015principal}.
For baselines (I)-(III), we also compare the performance of standard MMD testing without the variable selection technique.
We quantify the performance using the testing power metric with controlled type-I error $\alpha_{\textrm{level}}=0.05$, and take the training/testing sample sizes as $n_{\mathrm{Tr}}=n_{\mathrm{Te}}=n$. 

\begin{figure}[!ht]
    \centering
    \if\paperversion1
    \includegraphics[width=0.49\textwidth]{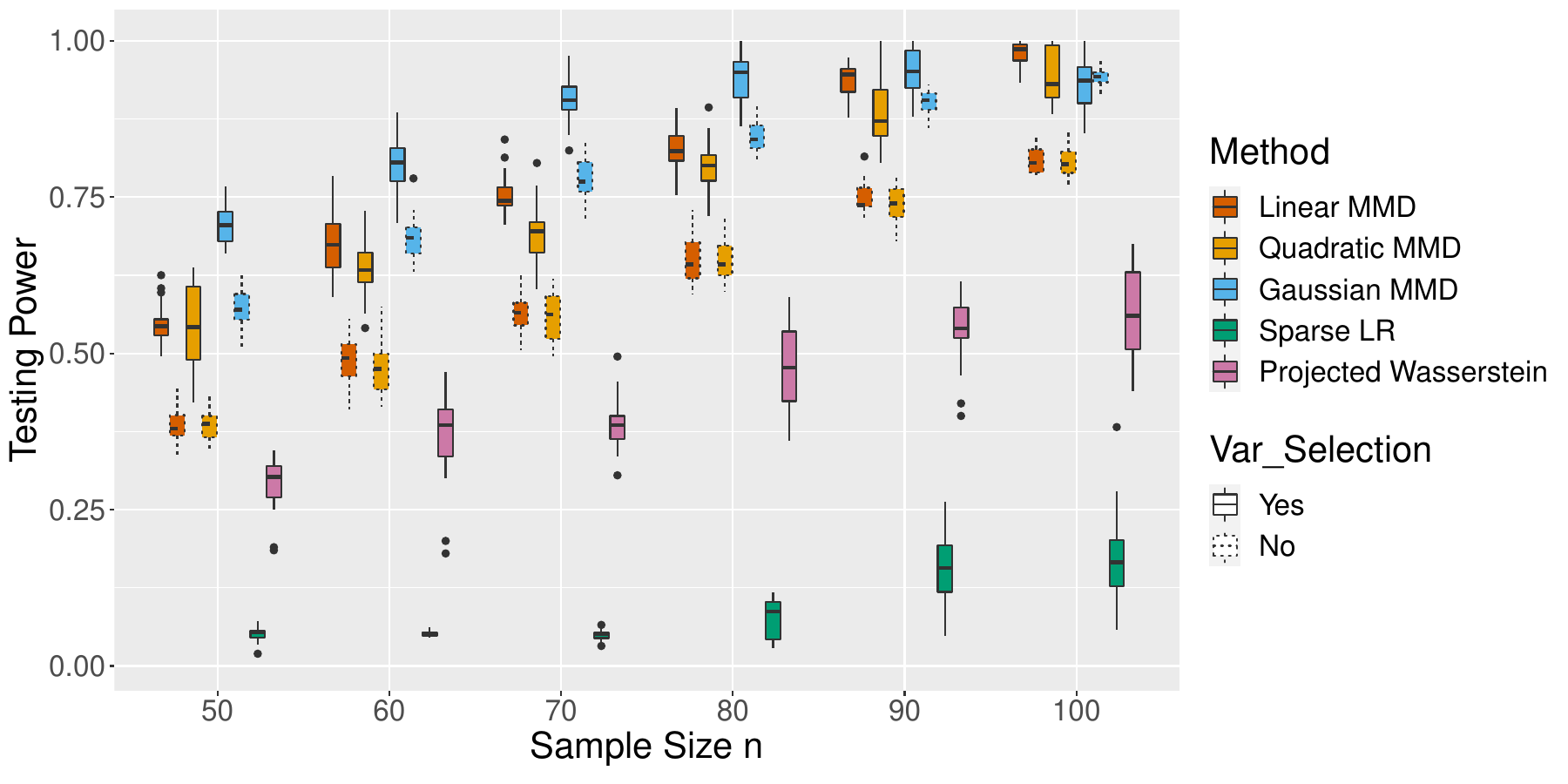}
    \includegraphics[width=0.49\textwidth]{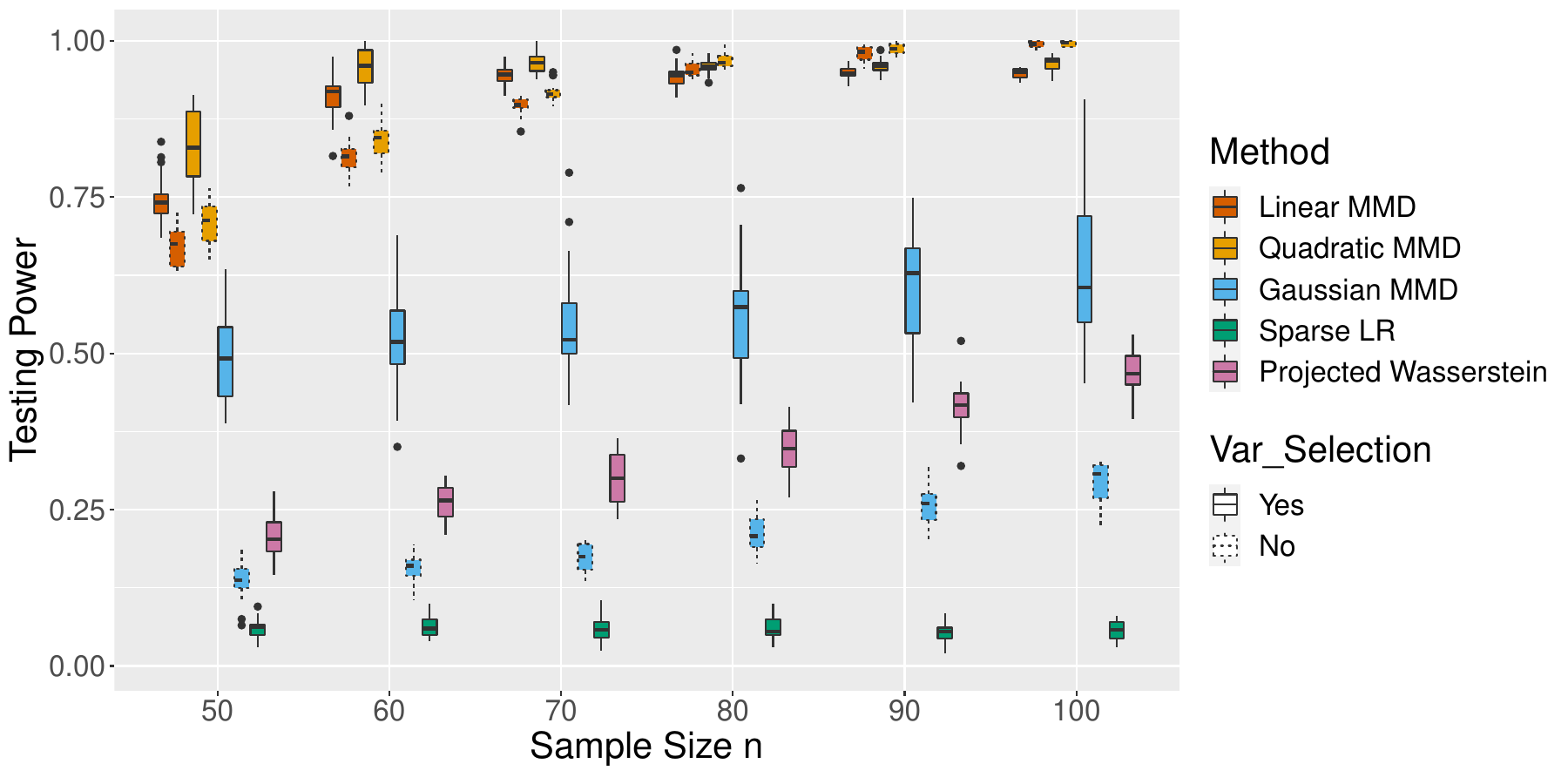}
    \includegraphics[width=0.49\textwidth]{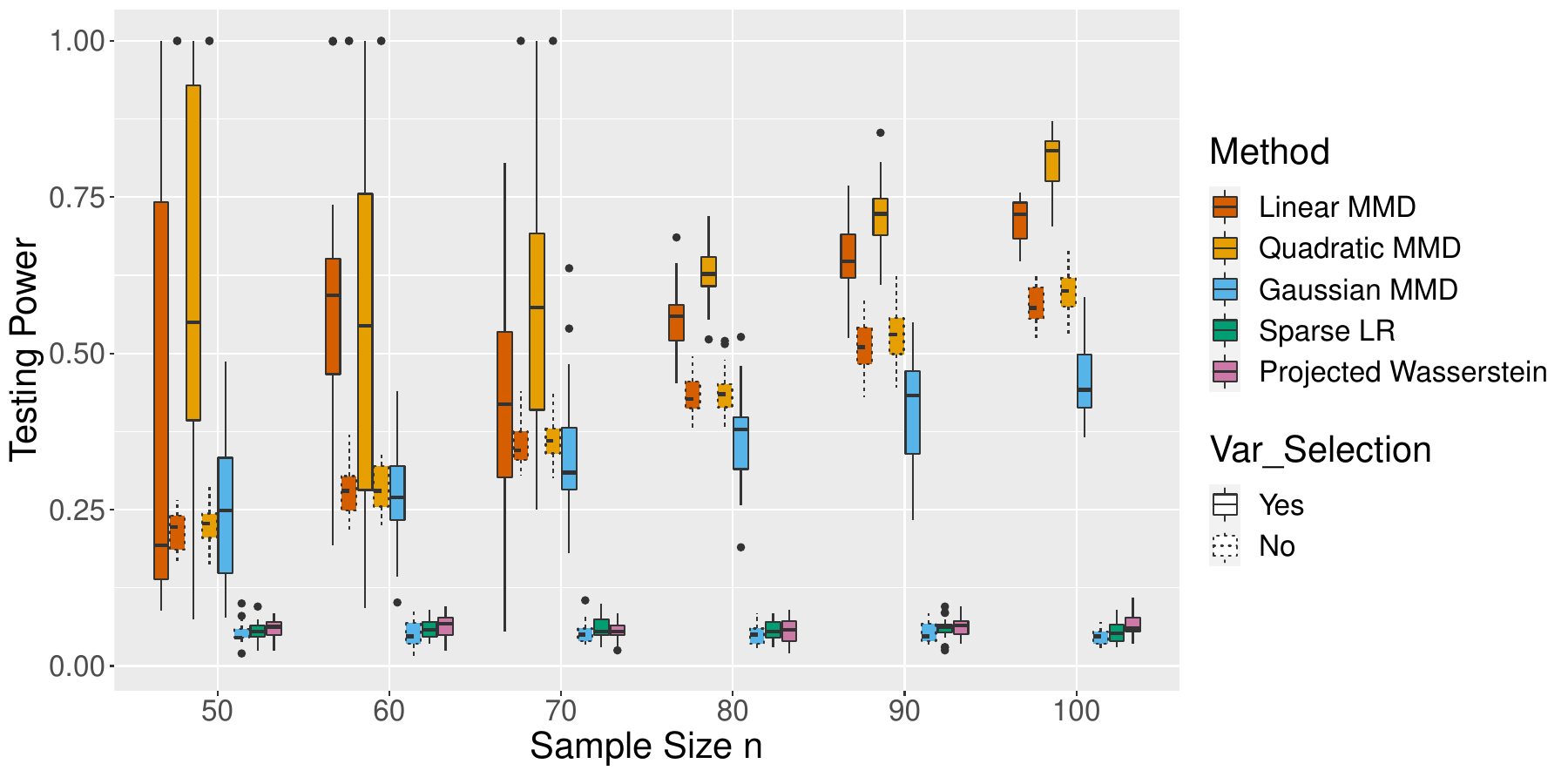}
    \includegraphics[width=0.49\textwidth]{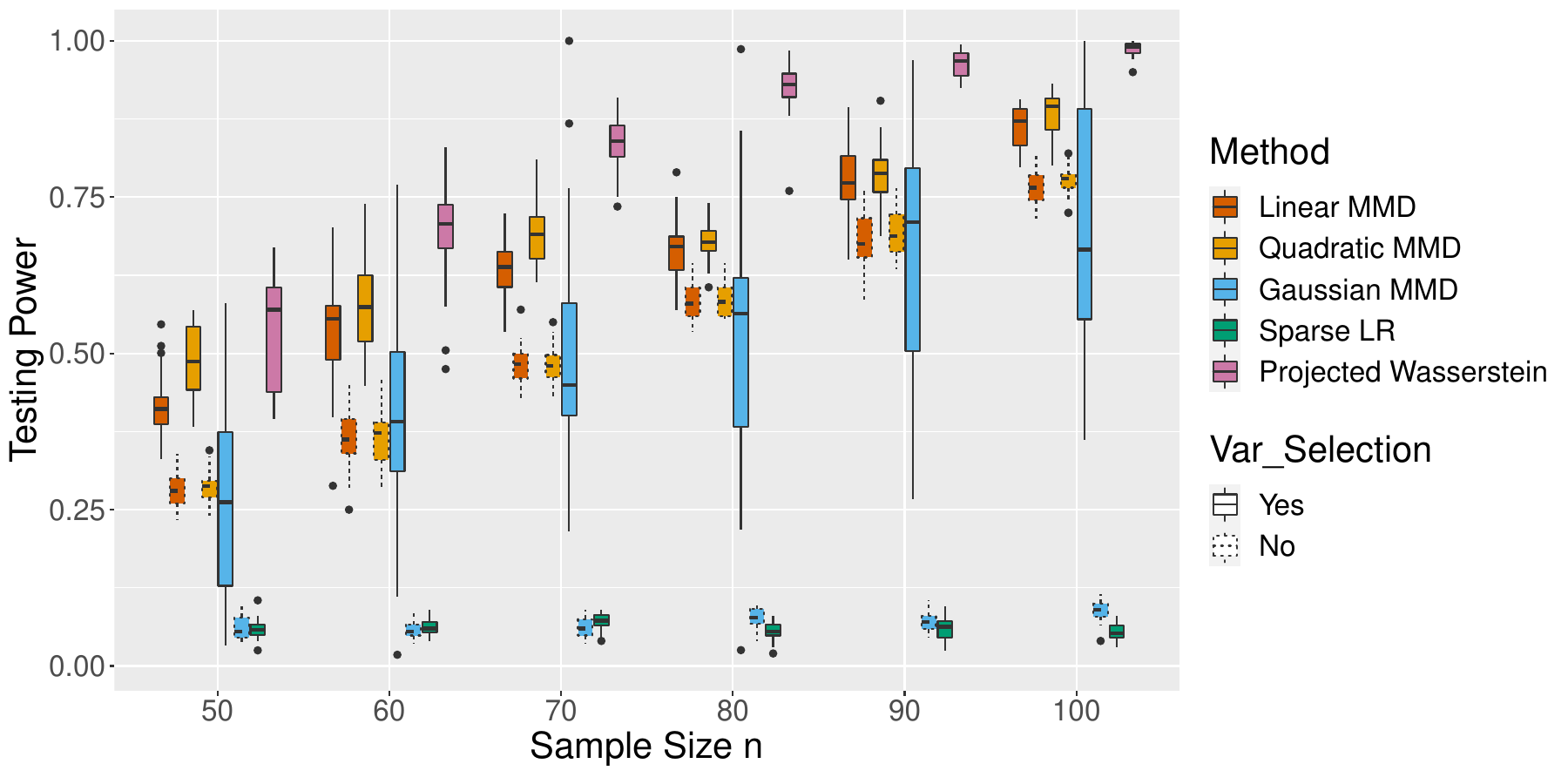}
    \fi
    \if\paperversion2
 \includegraphics[width=0.42\textwidth]{Power_Sample_case1.pdf}
    \includegraphics[width=0.42\textwidth]{Power_Sample_case2.pdf}
    \includegraphics[width=0.42\textwidth]{Power_Sample_case3.pdf}
    \includegraphics[width=0.42\textwidth]{Power_Sample_case4.pdf}
    \fi 
    \caption{Testing power of various two-sample tests with different choices of sample size $n$.
    Here we fix parameters $D=100, 
~d_{\mathrm{true}}=20,~d=20$ and control the type-I error $\alpha_{\mathrm{level}}=0.05$.
    Plots from top to bottom correspond to four different types of synthetic datasets.}
    \label{fig:power:sample:size}
\end{figure}
Figure~\ref{fig:power:sample:size} reports a numerical study on the impact of sample size $n$ with data dimension $D=100$, number of different variables $d_{\mathrm{true}}=20$ and sparsity level $d=20$.
The error bars are generated using 20 independent trials.
From these plots, we find that the sparse logistic regression does not have a competitive performance in general.
The explanation is that a linear classifier is not flexible enough to distinguish the distributions from two groups. 
Following the similar argument from Example~\ref{Eq:example:1}, one can check the testing statistic of this baseline always equals zero as long as the mean vectors of two distributions are the same, which explains why this baseline has nearly zero power for the synthetic dataset of case~(II)-(IV).
The testing power for the other two-sample testing methods increases with respect to the sample size.
We can see the variable selection technique improves the performance of the standard MMD framework.
For the first three synthetic datasets, the linear or quadratic MMD testing with variable selection achieves superior performance than other baselines. At the same time, for the last example, the projected Wasserstein distance has the best performance.
One possible explanation is that the MMD testing framework may not be good at detecting distribution changes for Gaussian mixture distributions.

\begin{figure}[!ht]
    \centering
    \includegraphics[width=0.49\textwidth]{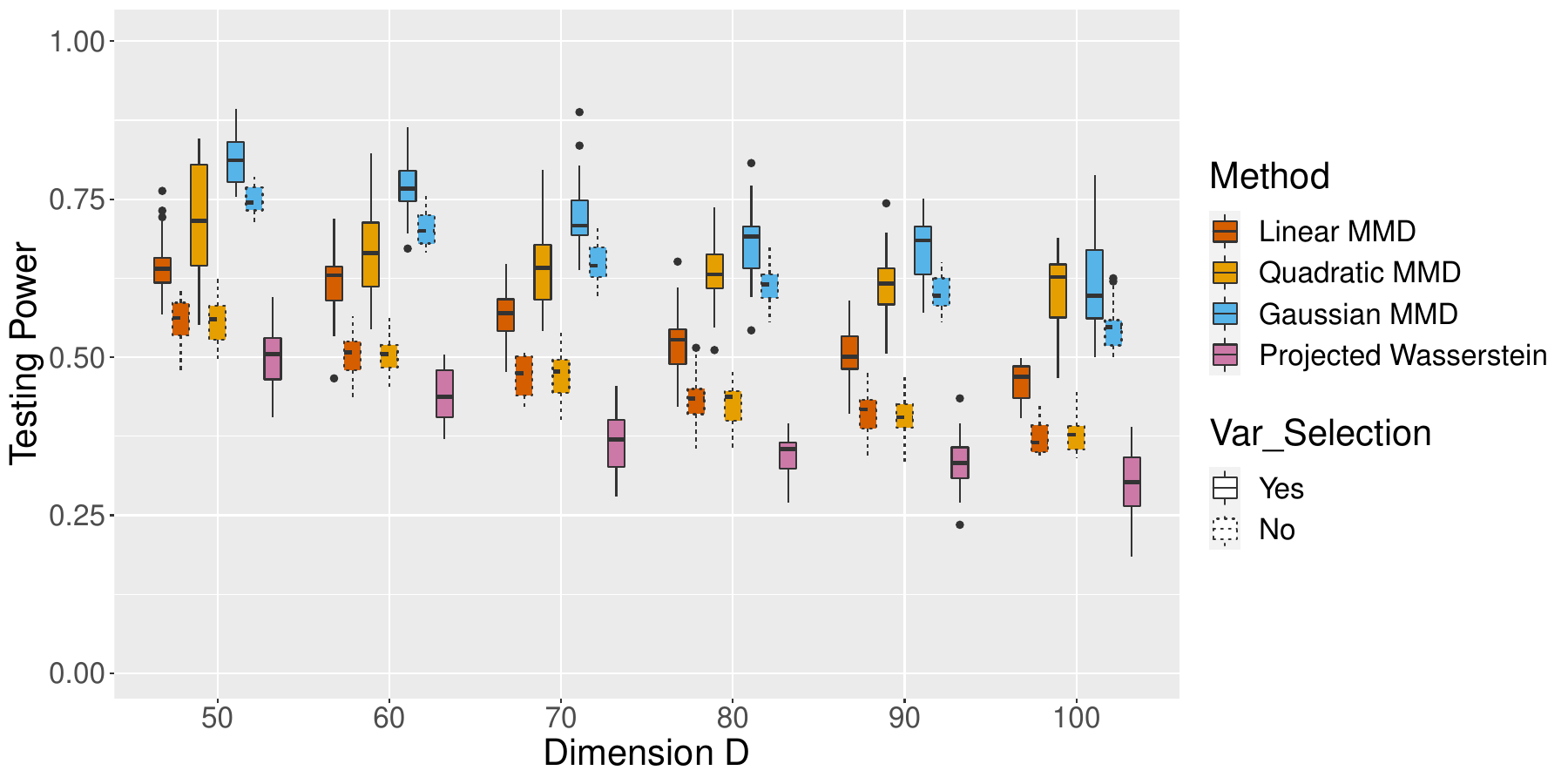}
    \includegraphics[width=0.49\textwidth]{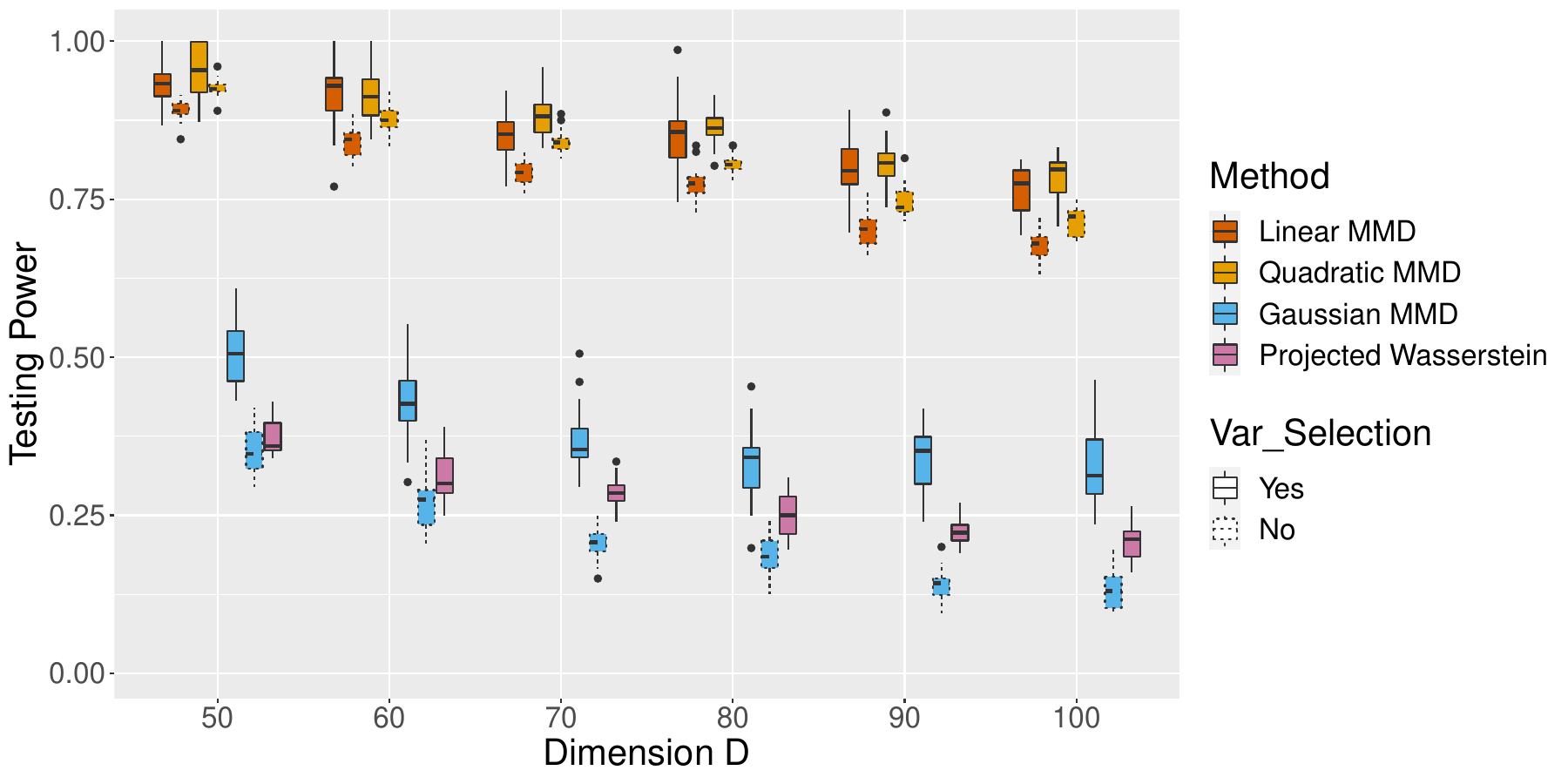}
    \includegraphics[width=0.49\textwidth]{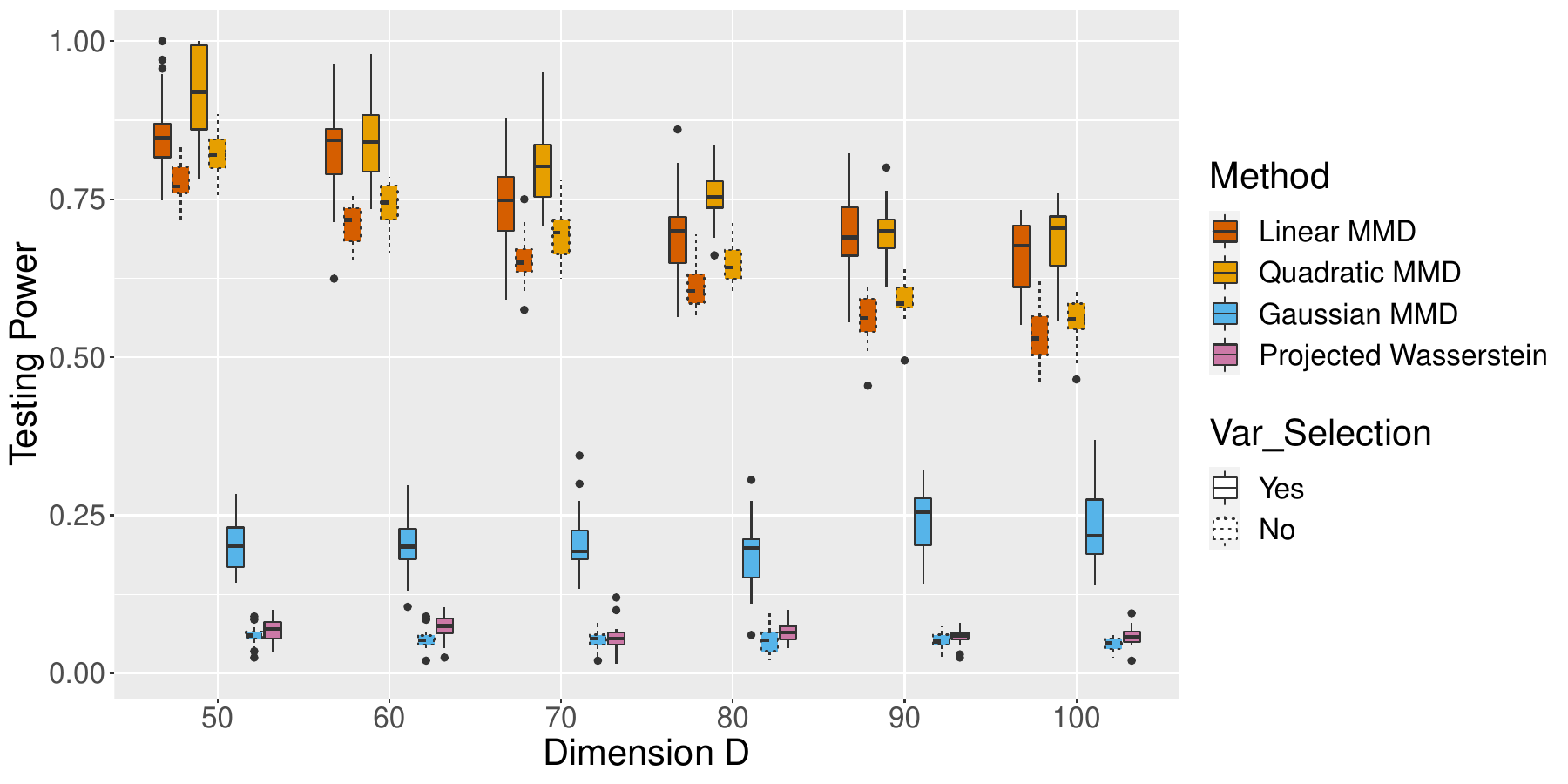}
    \includegraphics[width=0.49\textwidth]{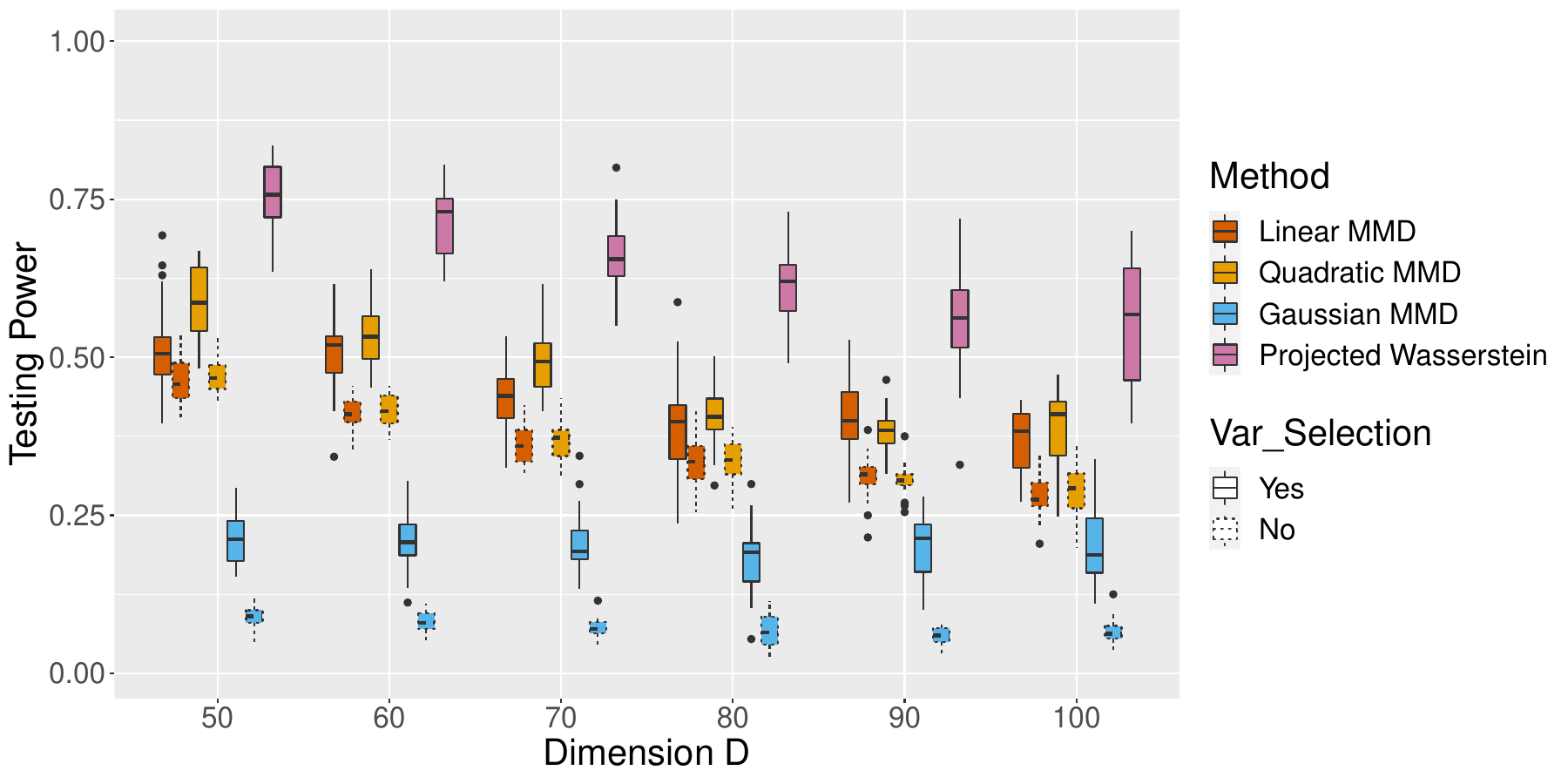}
    \caption{Testing power of various two-sample tests with different choices of data dimension $D$.
    Here we fix parameters $n=50,~d_{\mathrm{true}}=20,~d=20$ and control the type-I error $\alpha_{\mathrm{level}}=0.05$.
    Plots from top to bottom correspond to four different types of synthetic datasets.}
    \label{fig:power:data:dimension}
\end{figure}

Next, we examine the impact of the data dimension $D$ with fixed $n=50, d_{\mathrm{true}}=20, d=20$ in Figure~\ref{fig:power:data:dimension}.
We omit the performance of the sparse logistic regression baseline because it does not achieve satisfactory testing performance as studied before.
From those plots, we find that as the data dimension increases, all methods tend to have decreasing testing power. However, the decaying rate of MMD testing with the variable selection procedure seems slower than that of standard MMD testing.
For the synthetic dataset of case~(I), the Gaussian kernel has the best performance, while for case~(II)-(III), the linear or quadratic kernel has the best performance.
A possible explanation is that one can optimize the linear kernel with strong performance guarantees, whereas we only use quadratic approximation heuristics to optimize other types of kernel functions.
Since the quadratic approximation of the objective for the quadratic kernel seems to be tight, it is intuitive to see the performance of the quadratic kernel is also consistently good.

\subsection{Results on support recovery}
In this subsection, we demonstrate the performance of support recovery for various variable selection methods, evaluated using the FDP and NDP metrics defined in \eqref{Eq:FDP:NDP}. Since our goal is to assess consistent performance across both metrics, we also compute the average of the two as a combined measure. We set the parameters to $D=100$ and $d_{\mathrm{true}}=20$, while varying the sparsity level $d=1,\ldots,20$. The sample size $n=150$ remains fixed for Cases~(II) and (III), while for Case~(I) and Case~(IV), we adjust the parameters to $(\tau,n)=(2,200)$ and $(\tau,n)=(4,200)$, respectively, to ensure reliable support recovery.

\begin{figure}[!ht]
  \begin{center}
  \if\paperversion2
  \includegraphics[width=0.75\textwidth]{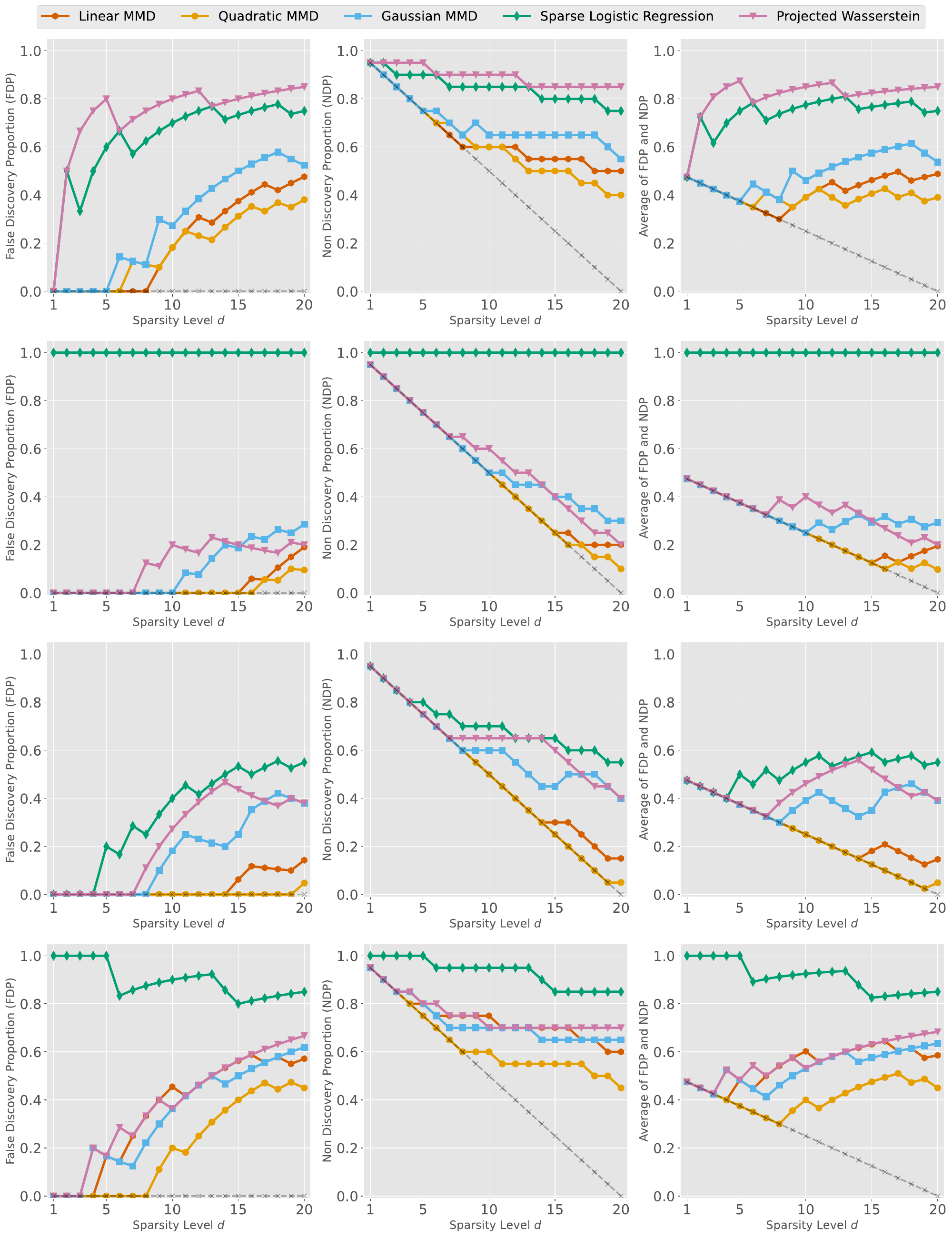}
  \fi
  \if\paperversion1
 \includegraphics[width=0.95\textwidth]{fig_revision_0905.pdf}
  \fi
  \end{center}
 \caption{FDP and NDP metrics obtained by various approaches for different choices of sparsity level $d$. The dimension of the problem is $D=100$; the true sparsity level is 20. %
 Figures for different columns correspond to different synthetic datasets.
 For each subplot, the dashed gray line denotes the performance for the ideal case where ground truth variables are selected successfully; the closer to the dashed gray line, the better.
 }
    \label{fig:FDP:NDP}
\end{figure}

Figure~\ref{fig:FDP:NDP} presents the numerical results for support recovery: each row corresponds to a different dataset, from Case~(I) to (IV), while each column represents a different performance measure (FDP, NDP, or their average). Each subplot illustrates the performance of various variable selection methods at different sparsity levels. From these plots, we observe that our proposed variable selection framework, whether using a linear or quadratic kernel, consistently outperforms the alternatives across all four cases, as reflected by the lowest FDP and NDP values. This finding aligns with the testing power performance discussed in the previous subsection. Additionally, it is noteworthy that for Cases~(II) and (III), our methods achieve near-optimal performance, as indicated by the nearly horizontal FDP lines and the almost straight NDP lines.

\section{Real data examples}

In this section, we present additional numerical studies with real-world datasets.
Specifically, we demonstrate a visualization of variable selection based on the MNIST handwritten digits image dataset in Section~\ref{Sec:visual}.
Next, we show that the variable selection approach can help identify key variables for disease diagnosis in Section~\ref{Sec:medical}.

\subsection{Visualization on MNIST image datasets}\label{Sec:visual}
In this part, we demonstrate a visualization of our variable selection framework by taking the classification of MNIST image datasets, consisting of $28\times28$ gray-scale handwritten images for digits from $0$ to $9$, as toy examples. 
We take the training sample size $n_{\mathrm{Tr}}=20$ and testing sample size $n_{\mathrm{Te}}=5$.
We pre-process the MNIST images by performing a 2d convolutional operator using the kernel of size $9\times9$.
The pre-processed samples have dimension $D=169$, and
we take the number of selected variables $d=20$.
We construct four types of data distributions $(\mu,\nu)$ for two-sample testing: $\mu$ and $\nu$ are distributions of images corresponding to digits $0$ and $6$, $8$ and $9$, $3$ and $8$, or $7$ and $9$, respectively.
\begin{figure}[!ht]
    \centering
\includegraphics[width=0.19\textwidth]{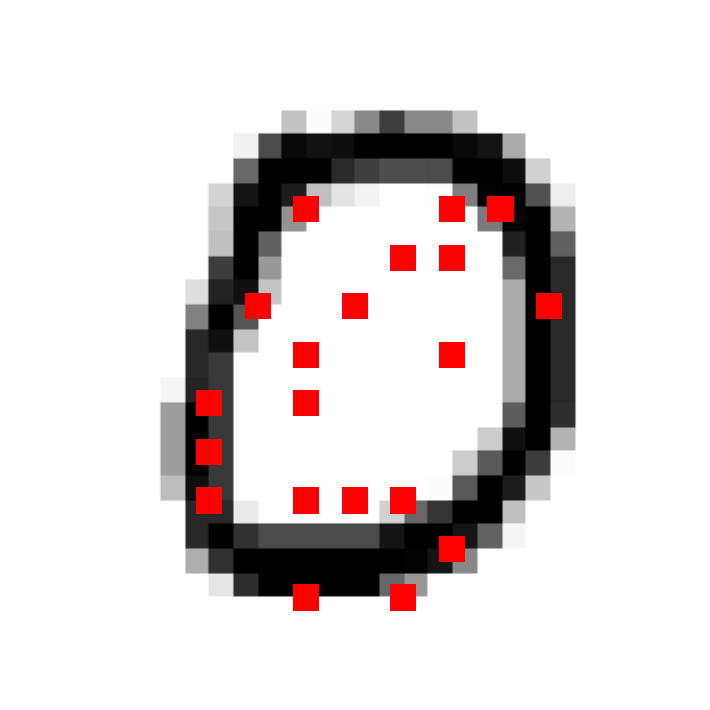}
\includegraphics[width=0.19\textwidth]{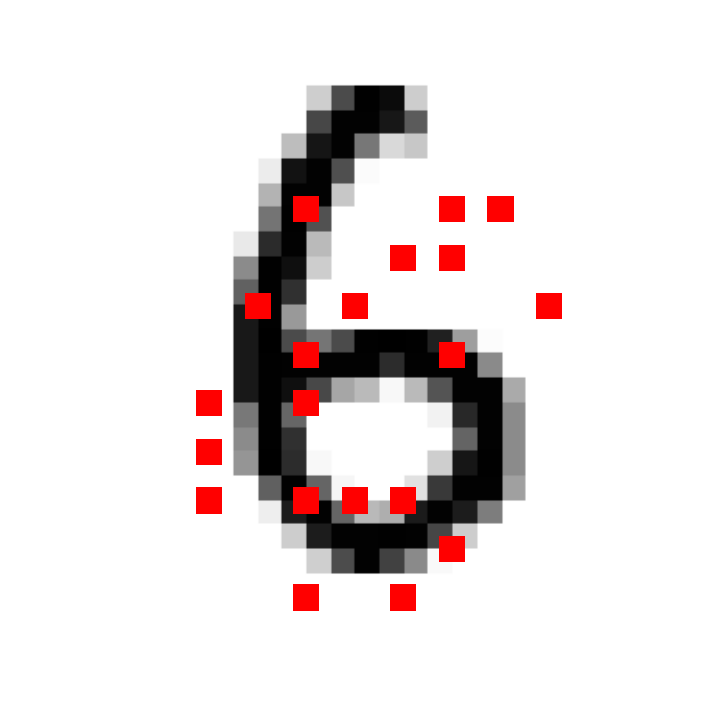}
\includegraphics[width=0.19\textwidth]{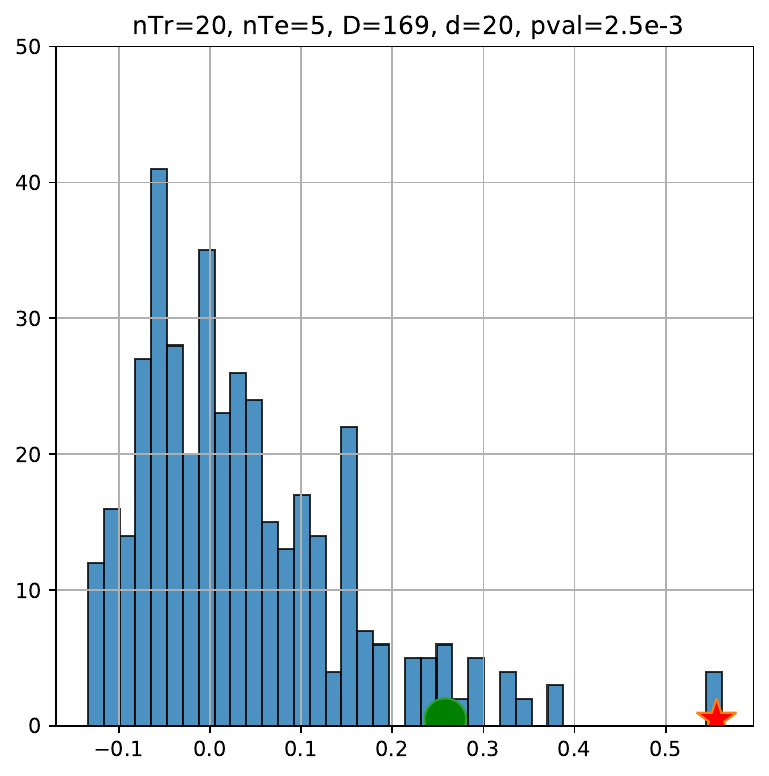}
\includegraphics[width=0.19\textwidth]{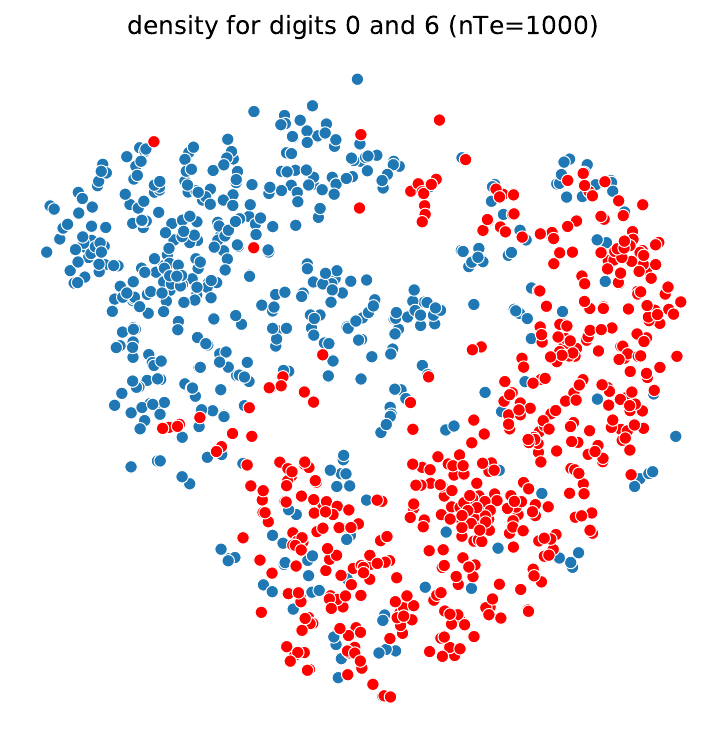}
\includegraphics[width=0.19\textwidth]{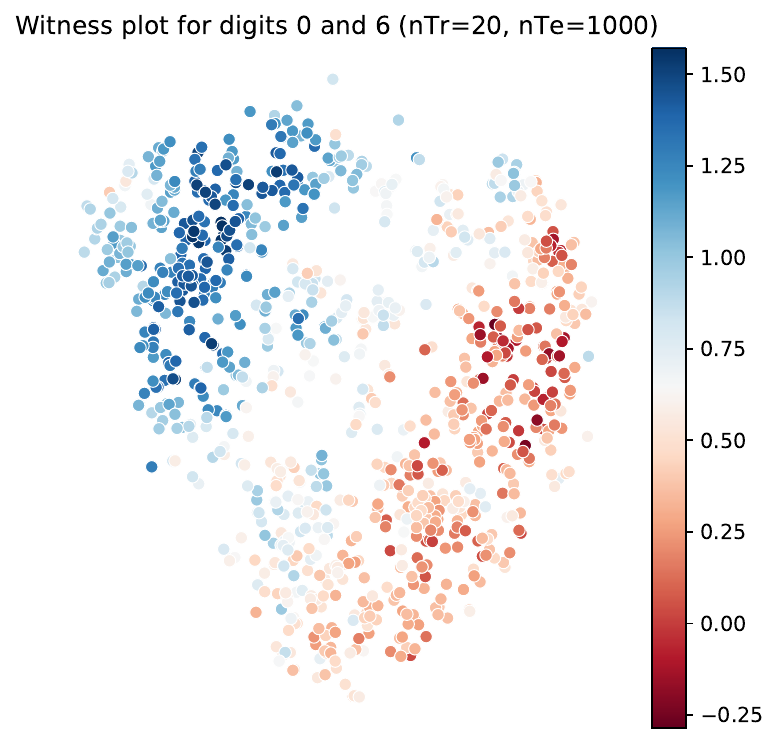}
\includegraphics[width=0.19\textwidth]{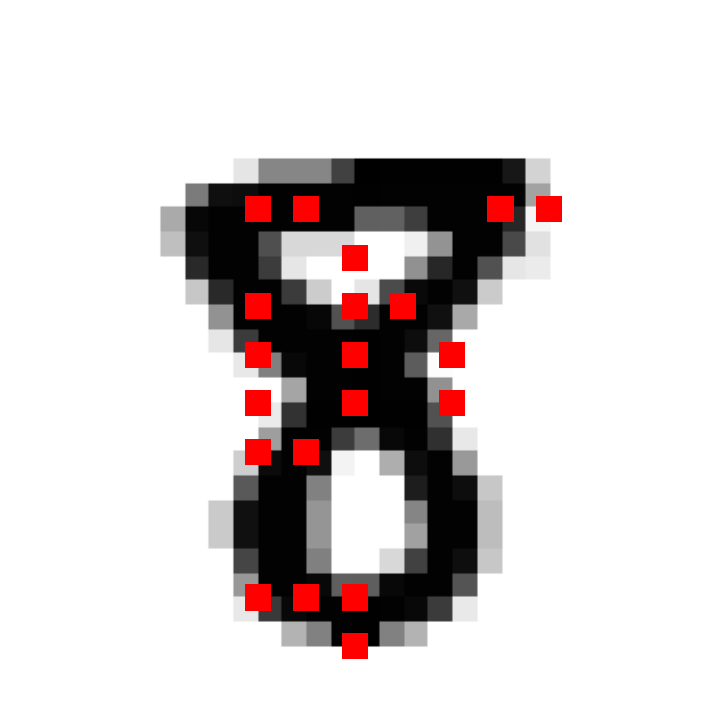}
\includegraphics[width=0.19\textwidth]{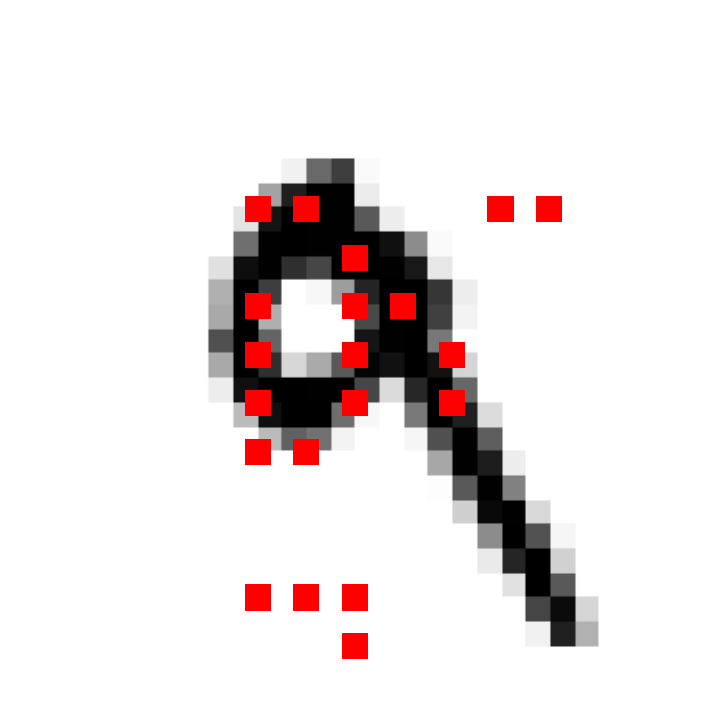}
\includegraphics[width=0.19\textwidth]{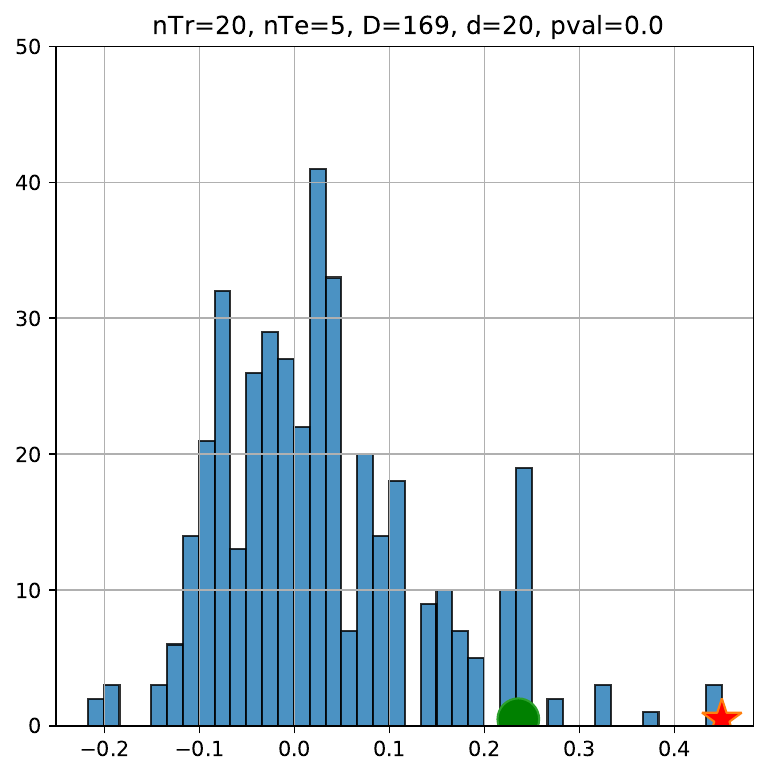}
\includegraphics[width=0.19\textwidth]{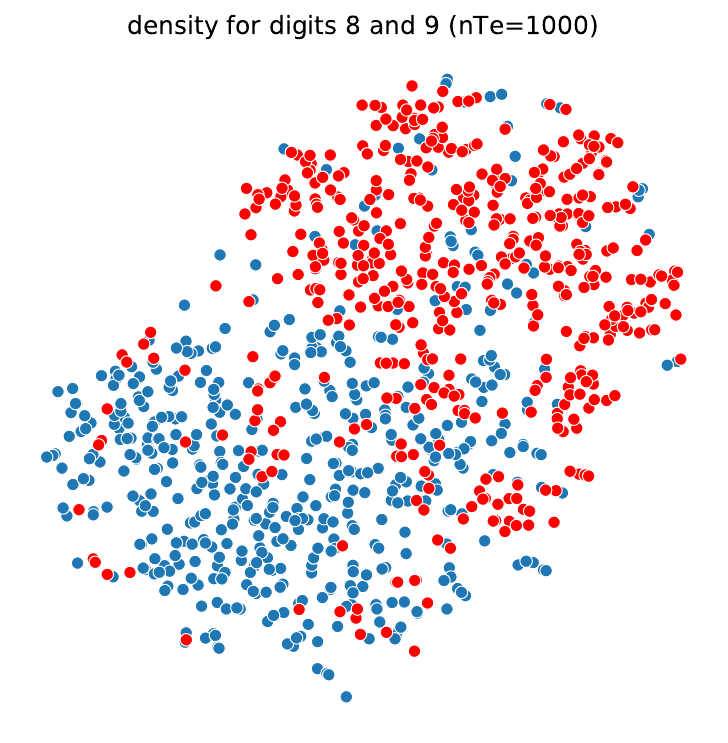}
\includegraphics[width=0.19\textwidth]{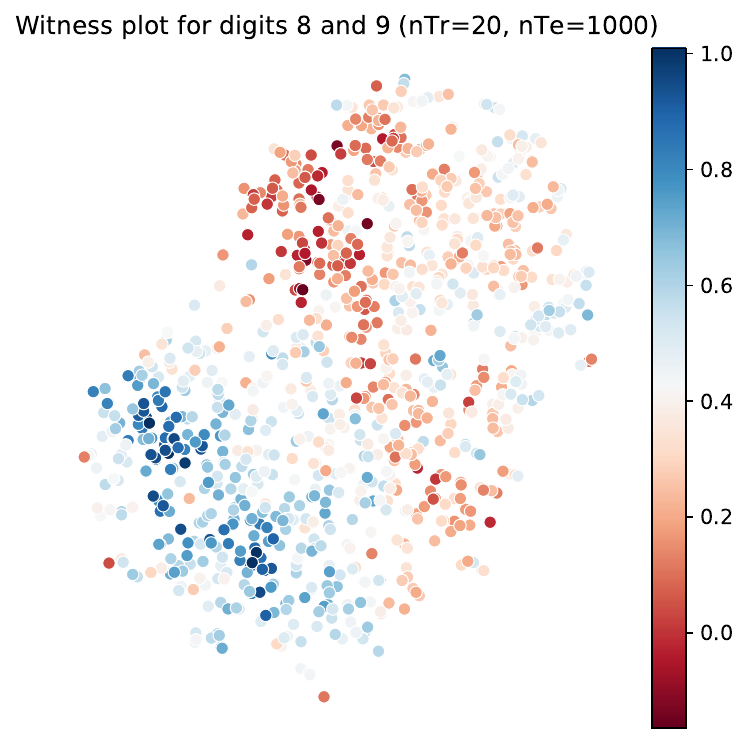}
\includegraphics[width=0.19\textwidth]{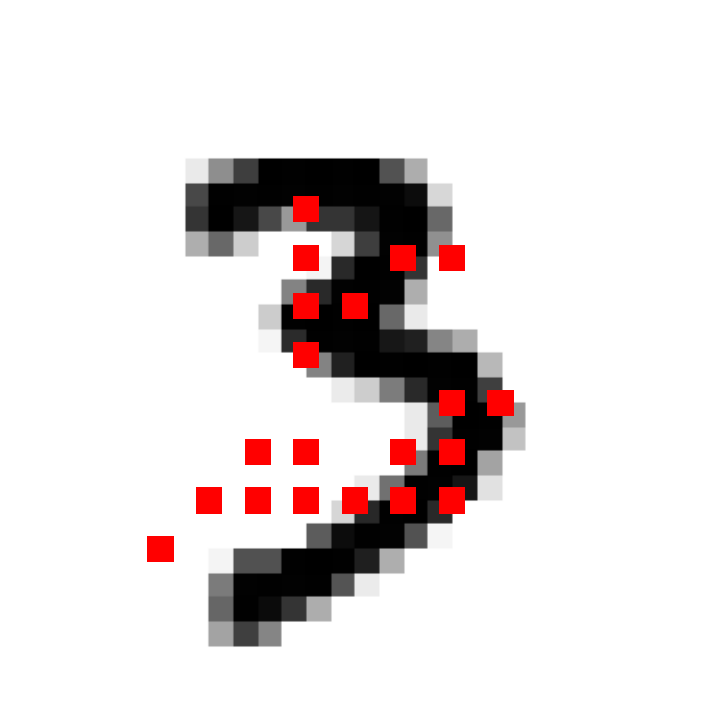}
\includegraphics[width=0.19\textwidth]{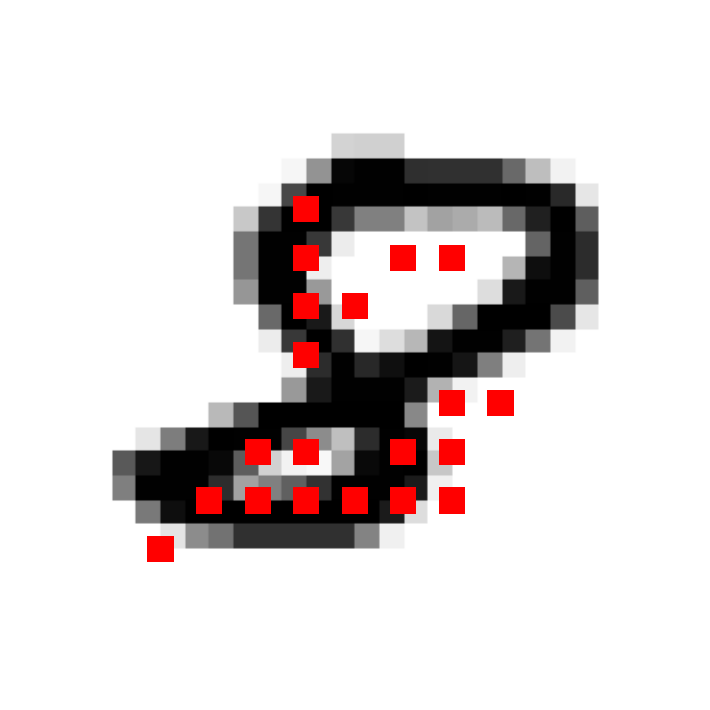}
\includegraphics[width=0.19\textwidth]{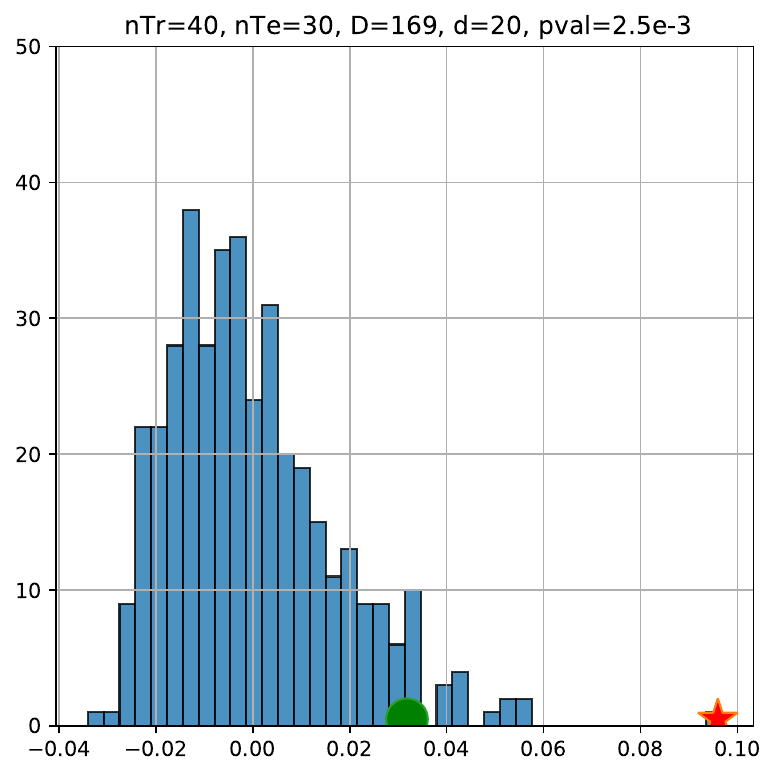}
\includegraphics[width=0.19\textwidth]{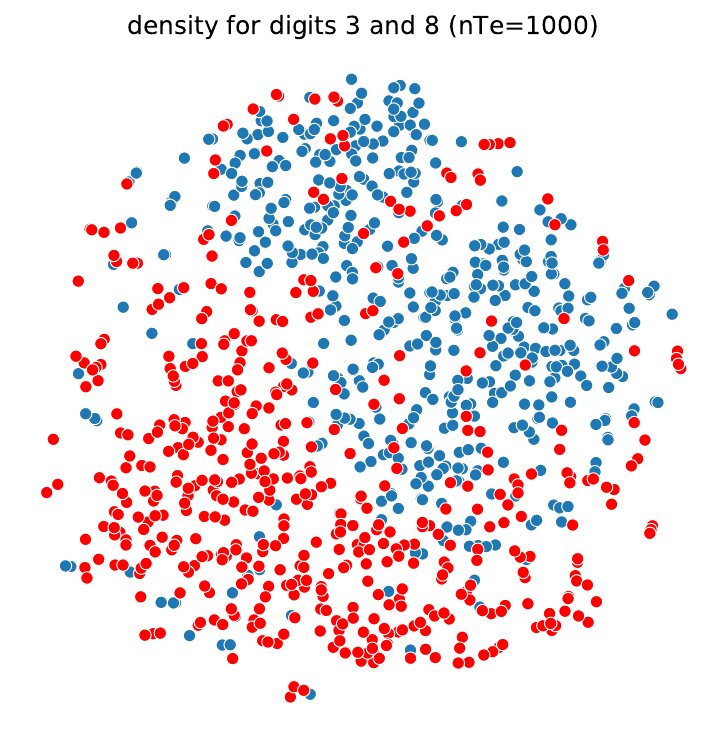}
\includegraphics[width=0.19\textwidth]{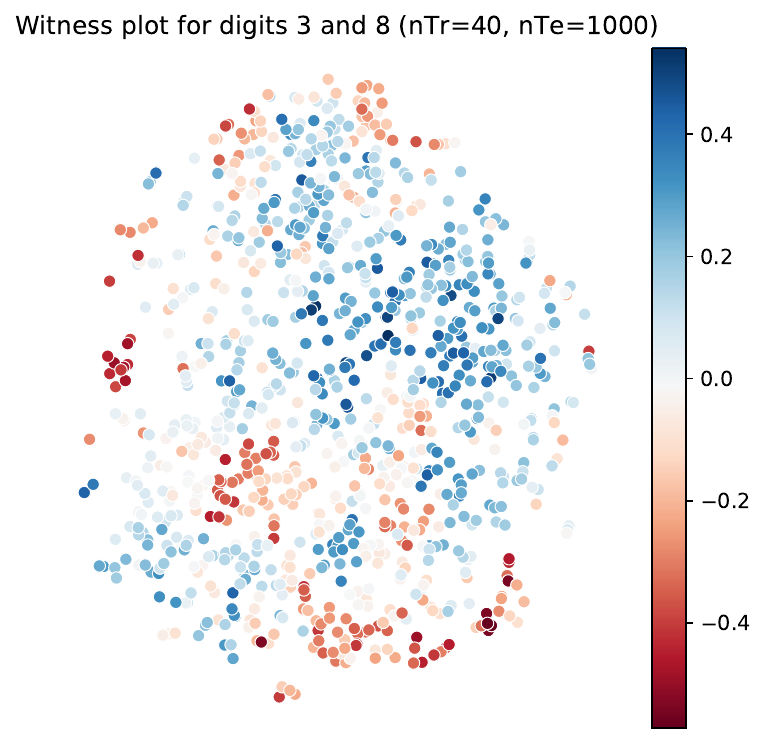}
\includegraphics[width=0.19\textwidth]{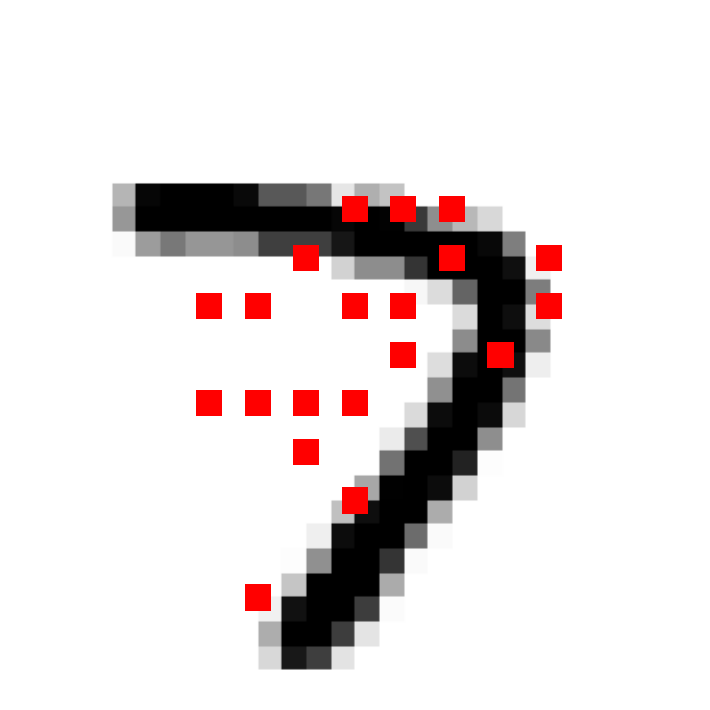}
\includegraphics[width=0.19\textwidth]{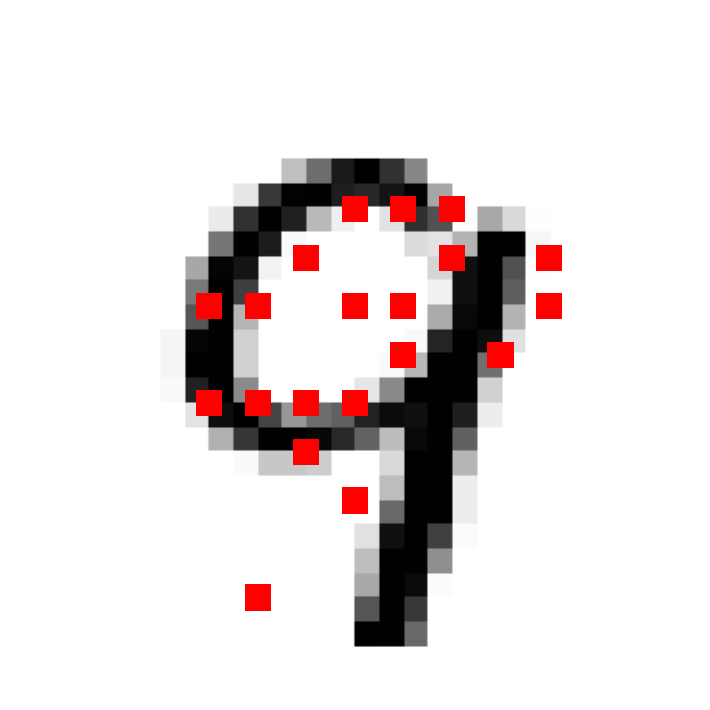}
\includegraphics[width=0.19\textwidth]{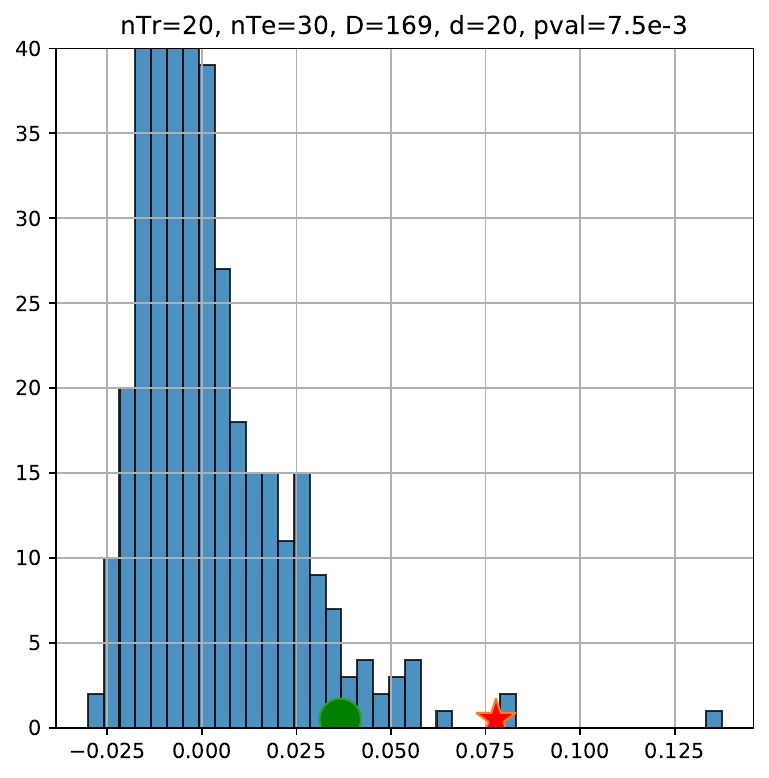}
\includegraphics[width=0.19\textwidth]{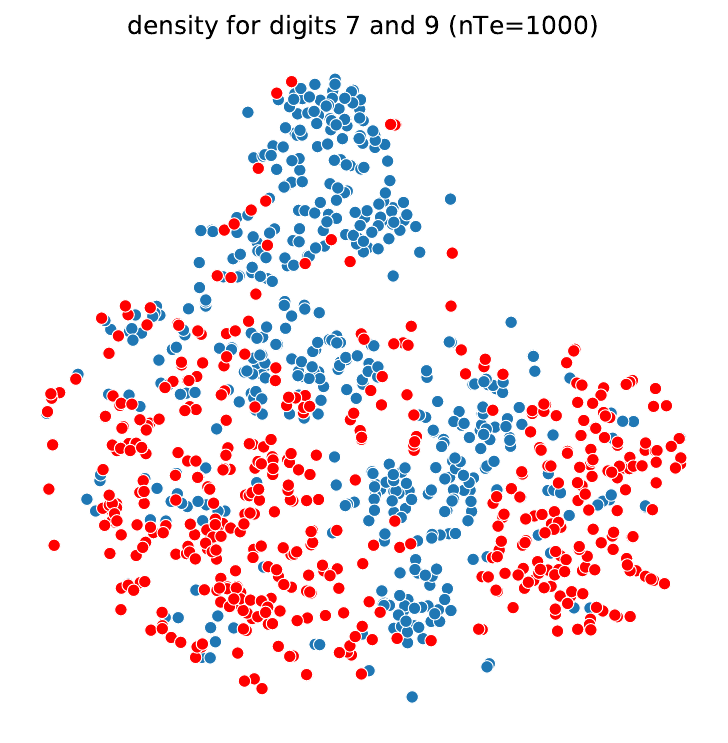}
\includegraphics[width=0.19\textwidth]{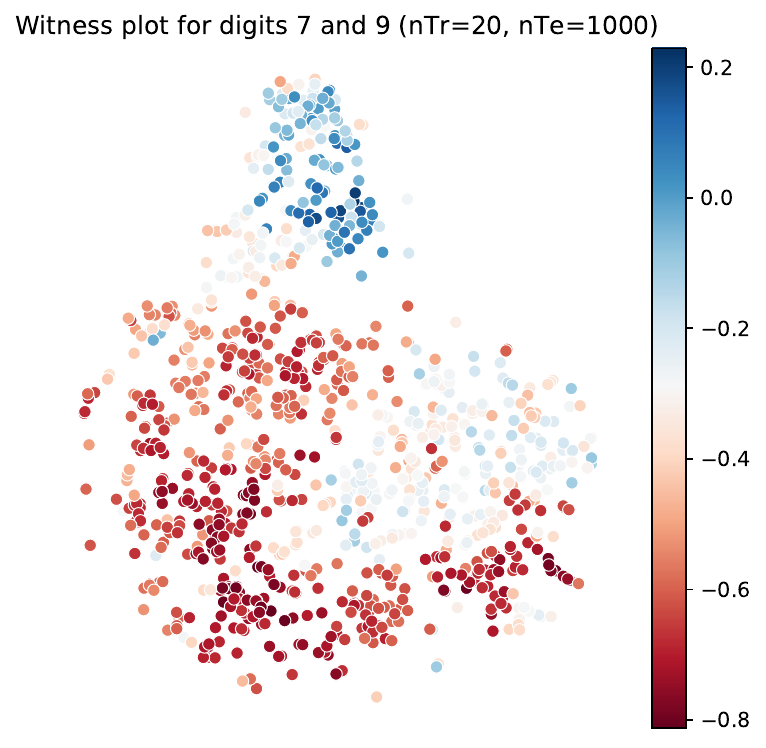}
    \caption{
    Different rows correspond to two-sample testing with different MNIST digits.
    The first two column plots visualize the selected pixels based on the variable selection framework.
    The third column plots visualize the MMD statistic together with the empirical distribution under $H_0$ that is estimated via bootstrapping (the green circle markers correspond to the bootstrap threshold for rejecting $H_0$, and red star markers correspond to the testing statistics).
    The fourth column plots visualize the distribution of MNIST digits after variable selection embedded in 2D.
    The fifth column plots visualize the estimated witness function~~(defined in \eqref{Eq:logit}) for MMD.}
    \label{fig:box:performance:app:visual}
\end{figure}
We show the visualization results in Figure~\ref{fig:box:performance:app:visual}.
Specifically, different rows correspond to different data distributions for two-sample testing.
Plots in the left two columns visualize the selected pixels (highlighted with red square markers) on two different image samples based on our linear kernel variable selection framework, from which we can see that our proposed method identifies the difference between two digits correctly.
Plots in the third column report the MMD statistic compared with the empirical distribution under $H_0$ via test-only bootstrap, where the green circle markers correspond to the bootstrap threshold for rejecting $H_0$ and red star markers correspond to the testing statistics.
From these plots, we find our proposed framework has satisfactory testing power even with small training and testing sample sizes.
Plots on the fourth column report the visualization of the distribution of the MNIST dataset after variable selection embedded in 2D generated by tSNE~\citep{van2014accelerating}, which is estimated based on $1000$ testing samples.
In comparison, we plot the estimated witness function~(defined in \eqref{Eq:logit}) as a color field over those samples in the fifth column.
From those plots, we can see that the estimated witness function identifies the region of the distribution change for all of these four two-sample testing tasks.

\subsection{Healthcare datasets}\label{Sec:medical}

Finally, we study the performance of  {variable selection} on a healthcare dataset~\citep{wang2022improving} that records information for healthy people and Sepsis patients. 
This dataset consists of $D=39$ features from $m=20771$ healthy people and $n=2891$ Sepsis patients.
We take training samples with sample sizes $m_{\text{Tr}}=20000, n_{\text{Tr}}=2000$ and specify the remaining as validation samples.
We quantify the performance of  {variable selection} as the testing power on testing samples with sample size $m_{\text{Te}}=n_{\text{Te}}=100$.
The testing power is computed based on randomly selected samples and their associated labels from the validation sample sets, with a significance level of $0.05$.
We repeat the testing procedure for $2000$ independent trials and report the average testing power in Table~\ref{tab:MSRC:full}.

\begin{figure}[!ht]
    \centering
    \if\paperversion2
     \includegraphics[width=0.8\textwidth]{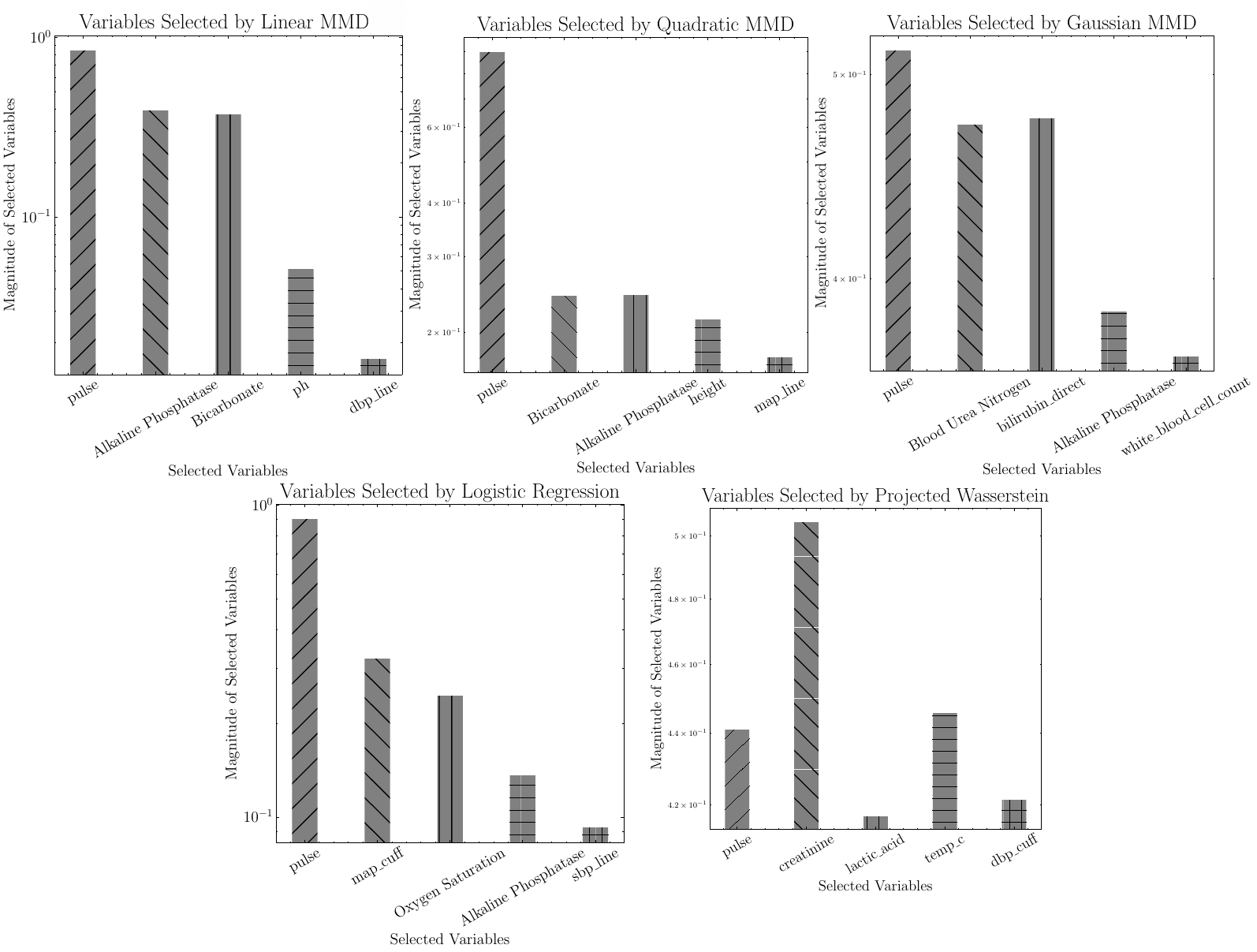}
     \fi
     \if\paperversion1
\includegraphics[width=0.9\textwidth]{Health_10111.pdf}
     \fi 
    \caption{Top $5$ {variables} selected by various approaches in the healthcare dataset.}
    \label{fig:my_label}
\end{figure}

We report the top $5$ features selected by various approaches based on the training samples in Figure~\ref{fig:my_label}.
From the Table, we can see that methods \textsf{Quadratic MMD} and \textsf{Linear MMD} perform the best, and the intersection of those selected features are~
pulse,~Bicarbonate,~Alkaline Phosphatase.
\begin{table}[!ht]
  \centering
  \caption{
  Averaged testing power for the sepsis prediction at a significance level $\alpha=0.05$ (i.e., the threshold is set such that the Type-I error when the two distributions are the same is set to 0.05.)
  }
    \begin{tabular}{ccccc}
    \toprule
  $\begin{array}{l}
\mbox{Linear}\\\mbox{MMD}
  \end{array}$ &  
  $\begin{array}{l}
\mbox{Quadratic}\\\mbox{MMD}
  \end{array}$
  &  $\begin{array}{l}
\mbox{Gaussian}\\\mbox{MMD}
  \end{array}$ &  $\begin{array}{l}
\mbox{Logistic}\\\mbox{Regression}
  \end{array}$ &  $\begin{array}{l}
\mbox{Projected}\\\mbox{Wasserstein}
  \end{array}$ \\
    \midrule
    0.835 & 0.915 & 0.784   & 0.771 &  0.749\\
    \bottomrule
    \end{tabular}%
  \label{tab:MSRC:full}%
\end{table}

\section{Conclusion}
We studied  {variable selection} for the kernel-based two-sample testing problem, which can be formulated as mixed-integer programming problems.
We developed exact and approximate algorithms with performance guarantees to solve those formulations.
Theoretical properties for the proposed frameworks are provided.
Finally, we validated the power of this approach in synthetic and real datasets.
In the meantime, several interesting research topics are left for future work. For example, providing theoretical analysis on the optimal choice of kernel hyper-parameters and support recovery for variable selection is of future research interest. Additionally, it holds great significance in developing more efficient algorithms for variable selection when working with different types of kernels.

\theendnotes
\if\paperversion1
\bibliographystyle{informs2014} 
\fi
{
\if\paperversion2
\footnotesize
\fi
\bibliography{shortbib}

\begin{thebibliography}{67}
\providecommand{\natexlab}[1]{#1}
\providecommand{\url}[1]{\texttt{#1}}
\providecommand{\urlprefix}{URL }

\bibitem[{Adachi et~al.(2017)Adachi, Iwata, Nakatsukasa, \protect\BIBand{}
  Takeda}]{adachi2017solving}
Adachi S, Iwata S, Nakatsukasa Y, Takeda A (2017) Solving the trust-region
  subproblem by a generalized eigenvalue problem. \emph{SIAM Journal on
  Optimization} 27(1):269--291.

\bibitem[{Alizadeh(1995)}]{alizadeh1995interior}
Alizadeh F (1995) Interior point methods in semidefinite programming with
  applications to combinatorial optimization. \emph{SIAM journal on
  Optimization} 5(1):13--51.

\bibitem[{Arcones \protect\BIBand{} Gine(1992)}]{arcones1992bootstrap}
Arcones MA, Gine E (1992) On the bootstrap of u and v statistics. \emph{The
  Annals of Statistics} 655--674.

\bibitem[{Bajwa \protect\BIBand{} Mixon(2012)}]{bajwa2012group}
Bajwa WU, Mixon DG (2012) Group model selection using marginal correlations:
  The good, the bad and the ugly. \emph{2012 50th Annual Allerton Conference on
  Communication, Control, and Computing}, 494--501.

\bibitem[{Beale \protect\BIBand{} Tomlin(1970)}]{beale1970special}
Beale EML, Tomlin JA (1970) Special facilities in a general mathematical
  programming system for non-convex problems using ordered sets of variables.
  \emph{Operations Research} 69(447-454):99.

\bibitem[{Berk \protect\BIBand{} Bertsimas(2019)}]{berk2019certifiably}
Berk L, Bertsimas D (2019) Certifiably optimal sparse principal component
  analysis. \emph{Mathematical Programming Computation} 11:381--420.

\bibitem[{Berlinet \protect\BIBand{}
  Thomas-Agnan(2011)}]{berlinet2011reproducing}
Berlinet A, Thomas-Agnan C (2011) \emph{Reproducing kernel Hilbert spaces in
  probability and statistics} (Springer Science \& Business Media).

\bibitem[{Bertsimas et~al.(2022)Bertsimas, Cory-Wright, \protect\BIBand{}
  Pauphilet}]{bertsimas2022solving}
Bertsimas D, Cory-Wright R, Pauphilet J (2022) Solving large-scale sparse {PCA}
  to certifiable (near) optimality. \emph{Journal of Machine Learning Research}
  23:13--1.

\bibitem[{Bertsimas et~al.(2021)Bertsimas, Pauphilet, \protect\BIBand{}
  Van~Parys}]{bertsimas2021sparse}
Bertsimas D, Pauphilet J, Van~Parys B (2021) Sparse classification: a scalable
  discrete optimization perspective. \emph{Machine Learning} 110:3177--3209.

\bibitem[{Bertsimas \protect\BIBand{}
  Tsitsiklis(1993)}]{bertsimas1993simulated}
Bertsimas D, Tsitsiklis J (1993) Simulated annealing. \emph{Statistical
  science} 8(1):10--15.

\bibitem[{Bonami et~al.(2008)Bonami, Biegler, Conn, Cornu{\'e}jols, Grossmann,
  Laird, Lee, Lodi, Margot, Sawaya et~al.}]{bonami2008algorithmic}
Bonami P, Biegler LT, Conn AR, Cornu{\'e}jols G, Grossmann IE, Laird CD, Lee J,
  Lodi A, Margot F, Sawaya N, et~al. (2008) An algorithmic framework for convex
  mixed integer nonlinear programs. \emph{Discrete optimization} 5(2):186--204.

\bibitem[{Bonferroni(1936)}]{bonferroni1936teoria}
Bonferroni C (1936) Teoria statistica delle classi e calcolo delle probabilita.
  \emph{Pubblicazioni del R Istituto Superiore di Scienze Economiche e
  Commericiali di Firenze} 8:3--62.

\bibitem[{Boyd et~al.(2011)Boyd, Parikh, Chu, Peleato, Eckstein
  et~al.}]{boyd2011distributed}
Boyd S, Parikh N, Chu E, Peleato B, Eckstein J, et~al. (2011) Distributed
  optimization and statistical learning via the alternating direction method of
  multipliers. \emph{Foundations and Trends{\textregistered} in Machine
  learning} 3(1):1--122.

\bibitem[{Boyd \protect\BIBand{} Vandenberghe(2004)}]{boyd2004convex}
Boyd SP, Vandenberghe L (2004) \emph{Convex optimization} (Cambridge university
  press).

\bibitem[{Chan et~al.(2016)Chan, Papailliopoulos, \protect\BIBand{}
  Rubinstein}]{chan2016approximability}
Chan SO, Papailliopoulos D, Rubinstein A (2016) On the approximability of
  sparse {PCA}. \emph{Conference on Learning Theory}, 623--646.

\bibitem[{Cheng \protect\BIBand{} Cloninger(2022)}]{cheng2022classification}
Cheng X, Cloninger A (2022) Classification logit two-sample testing by neural
  networks for differentiating near manifold densities. \emph{IEEE Transactions
  on Information Theory} 68(10):6631--6662.

\bibitem[{Cheng \protect\BIBand{} Xie(2021)}]{cheng2021kernel}
Cheng X, Xie Y (2021) Kernel two-sample tests for manifold data. \emph{arXiv
  preprint arXiv:2105.03425} .

\bibitem[{Chwialkowski et~al.(2015)Chwialkowski, Ramdas, Sejdinovic,
  \protect\BIBand{} Gretton}]{Chwialkowski15}
Chwialkowski KP, Ramdas A, Sejdinovic D, Gretton A (2015) Fast two-sample
  testing with analytic representations of probability measures. \emph{Advances
  in Neural Information Processing Systems}, volume~28.

\bibitem[{Conn et~al.(2000)Conn, Gould, \protect\BIBand{}
  Toint}]{conn2000trust}
Conn AR, Gould NI, Toint PL (2000) \emph{Trust region methods} (SIAM).

\bibitem[{Dey et~al.(2019)Dey, Kocuk, \protect\BIBand{}
  Santana}]{dey2020convexifications}
Dey SS, Kocuk B, Santana A (2019) Convexifications of rank-one-based
  substructures in qcqps and applications to the pooling problem. \emph{Journal
  of Global Optimization} 77(2):227--272.

\bibitem[{Dey et~al.(2022{\natexlab{a}})Dey, Mazumder, \protect\BIBand{}
  Wang}]{dey2022using}
Dey SS, Mazumder R, Wang G (2022{\natexlab{a}}) Using $\ell_1$-relaxation and
  integer programming to obtain dual bounds for sparse {PCA}. \emph{Operations
  Research} 70(3):1914--1932.

\bibitem[{Dey et~al.(2022{\natexlab{b}})Dey, Molinaro, \protect\BIBand{}
  Wang}]{dey2022solving}
Dey SS, Molinaro M, Wang G (2022{\natexlab{b}}) Solving sparse principal
  component analysis with global support. \emph{Mathematical Programming}
  1--39.

\bibitem[{Dominici*(2003)}]{dominici2003inverse}
Dominici* DE (2003) The inverse of the cumulative standard normal probability
  function. \emph{Integral Transforms and Special Functions} 14(4):281--292.

\bibitem[{Fletcher \protect\BIBand{} Leyffer(1994)}]{fletcher1994solving}
Fletcher R, Leyffer S (1994) Solving mixed integer nonlinear programs by outer
  approximation. \emph{Mathematical programming} 66:327--349.

\bibitem[{Fukumizu et~al.(2009)Fukumizu, Gretton, Lanckriet, Sch{\"o}lkopf,
  \protect\BIBand{} Sriperumbudur}]{fukumizu2009kernel}
Fukumizu K, Gretton A, Lanckriet G, Sch{\"o}lkopf B, Sriperumbudur BK (2009)
  Kernel choice and classifiability for rkhs embeddings of probability
  distributions. \emph{Advances in neural information processing systems} 22.

\bibitem[{Gally \protect\BIBand{} Pfetsch(2016)}]{gally2016computing}
Gally T, Pfetsch ME (2016) Computing restricted isometry constants via
  mixed-integer semidefinite programming. \emph{Preprint Available at
  Optimization Online} .

\bibitem[{Good(2013)}]{good2013permutation}
Good P (2013) \emph{Permutation tests: a practical guide to resampling methods
  for testing hypotheses} (Springer Science $\&$ Business Media).

\bibitem[{Gretton et~al.(2012{\natexlab{a}})Gretton, Borgwardt, Rasch,
  Sch\"{o}lkopf, \protect\BIBand{} Smola}]{Gretton12}
Gretton A, Borgwardt KM, Rasch MJ, Sch\"{o}lkopf B, Smola A
  (2012{\natexlab{a}}) A kernel two-sample test. \emph{Journal of Machine
  Learning Research} 13:723--773.

\bibitem[{Gretton et~al.(2009)Gretton, Fukumizu, Harchaoui, \protect\BIBand{}
  Sriperumbudur}]{Gretton09}
Gretton A, Fukumizu K, Harchaoui Z, Sriperumbudur BK (2009) A fast, consistent
  kernel two-sample test. \emph{Advances in Neural Information Processing
  Systems}, volume~22.

\bibitem[{Gretton et~al.(2012{\natexlab{b}})Gretton, Sejdinovic, Strathmann,
  Balakrishnan, Pontil, Fukumizu, \protect\BIBand{}
  Sriperumbudur}]{Grettonnips12}
Gretton A, Sejdinovic D, Strathmann H, Balakrishnan S, Pontil M, Fukumizu K,
  Sriperumbudur BK (2012{\natexlab{b}}) Optimal kernel choice for large-scale
  two-sample tests. \emph{Advances in neural information processing systems},
  1205--1213.

\bibitem[{Hara et~al.(2015)Hara, Morimura, Takahashi, Yanagisawa,
  \protect\BIBand{} Suzuki}]{hara2015consistent}
Hara S, Morimura T, Takahashi T, Yanagisawa H, Suzuki T (2015) A consistent
  method for graph based anomaly localization. \emph{Artificial intelligence
  and statistics}, 333--341.

\bibitem[{Hastie et~al.(2009)Hastie, Tibshirani, Friedman, \protect\BIBand{}
  Friedman}]{hastie2009elements}
Hastie T, Tibshirani R, Friedman JH, Friedman JH (2009) \emph{The elements of
  statistical learning: data mining, inference, and prediction}, volume~2
  (Springer).

\bibitem[{Id{\'e} et~al.(2009)Id{\'e}, Lozano, Abe, \protect\BIBand{}
  Liu}]{ide2009proximity}
Id{\'e} T, Lozano AC, Abe N, Liu Y (2009) Proximity-based anomaly detection
  using sparse structure learning. \emph{Proceedings of the 2009 SIAM
  international conference on data mining}, 97--108.

\bibitem[{Id{\'e} et~al.(2007)Id{\'e}, Papadimitriou, \protect\BIBand{}
  Vlachos}]{ide2007computing}
Id{\'e} T, Papadimitriou S, Vlachos M (2007) Computing correlation anomaly
  scores using stochastic nearest neighbors. \emph{Seventh IEEE international
  conference on data mining}, 523--528.

\bibitem[{Jitkrittum et~al.(2016)Jitkrittum, Szab\'{o}, Chwialkowski,
  \protect\BIBand{} Gretton}]{JitkrittumME}
Jitkrittum W, Szab\'{o} Z, Chwialkowski K, Gretton A (2016) Interpretable
  distribution features with maximum testing power. \emph{Proceedings of the
  30th International Conference on Neural Information Processing Systems}.

\bibitem[{Kirchler et~al.(2020)Kirchler, Khorasani, Kloft, \protect\BIBand{}
  Lippert}]{kirchler2020two}
Kirchler M, Khorasani S, Kloft M, Lippert C (2020) Two-sample testing using
  deep learning. \emph{International Conference on Artificial Intelligence and
  Statistics}, 1387--1398.

\bibitem[{K{\"u}bler et~al.(2022{\natexlab{a}})K{\"u}bler, Jitkrittum,
  Sch{\"o}lkopf, \protect\BIBand{} Muandet}]{kubler2022witness}
K{\"u}bler JM, Jitkrittum W, Sch{\"o}lkopf B, Muandet K (2022{\natexlab{a}}) A
  witness two-sample test. \emph{International Conference on Artificial
  Intelligence and Statistics}, 1403--1419.

\bibitem[{K{\"u}bler et~al.(2022{\natexlab{b}})K{\"u}bler, Stimper, Buchholz,
  Muandet, \protect\BIBand{} Sch{\"o}lkopf}]{kubler2022automl}
K{\"u}bler JM, Stimper V, Buchholz S, Muandet K, Sch{\"o}lkopf B
  (2022{\natexlab{b}}) Automl two-sample test. \emph{Advances in Neural
  Information Processing Systems} 35:15929--15941.

\bibitem[{Li et~al.(2024)Li, Fampa, Lee, Qiu, Xie, \protect\BIBand{}
  Yao}]{li2022d}
Li Y, Fampa M, Lee J, Qiu F, Xie W, Yao R (2024) D-optimal data fusion: Exact
  and approximation algorithms. \emph{INFORMS Journal on Computing}
  36(1):97--120.

\bibitem[{Li \protect\BIBand{} Xie(2020)}]{li2020exact}
Li Y, Xie W (2020) Exact and approximation algorithms for sparse {PCA}.
  \emph{arXiv preprint arXiv:2008.12438} .

\bibitem[{Li \protect\BIBand{} Xie(2022)}]{li2022exactness}
Li Y, Xie W (2022) On the exactness of dantzig-wolfe relaxation for rank
  constrained optimization problems. \emph{arXiv preprint arXiv:2210.16191} .

\bibitem[{Li \protect\BIBand{} Xie(2023)}]{li2021beyond}
Li Y, Xie W (2023) Beyond symmetry: Best submatrix selection for the sparse
  truncated {SVD}. \emph{Mathematical Programming} 1--50.

\bibitem[{Li \protect\BIBand{} Xie(2024)}]{li2020best}
Li Y, Xie W (2024) Best principal submatrix selection for the maximum entropy
  sampling problem: scalable algorithms and performance guarantees.
  \emph{Operations Research} 72(2):493--513.

\bibitem[{Liu et~al.(2020)Liu, Xu, Lu, Zhang, Gretton, \protect\BIBand{}
  Sutherland}]{liu2020learning}
Liu F, Xu W, Lu J, Zhang G, Gretton A, Sutherland DJ (2020) Learning deep
  kernels for non-parametric two-sample tests. \emph{International Conference
  on Machine Learning}, 6316--6326.

\bibitem[{Lofberg(2004)}]{lofberg2004yalmip}
Lofberg J (2004) Yalmip: A toolbox for modeling and optimization in matlab.
  \emph{2004 IEEE international conference on robotics and automation},
  284--289.

\bibitem[{Ma(2013)}]{ma2013alternating}
Ma S (2013) Alternating direction method of multipliers for sparse principal
  component analysis. \emph{Journal of the Operations Research Society of
  China} 1:253--274.

\bibitem[{Magdon-Ismail(2017)}]{magdon2017np}
Magdon-Ismail M (2017) Np-hardness and inapproximability of sparse {PCA}.
  \emph{Information Processing Letters} 126:35--38.

\bibitem[{Micchelli et~al.(2006)Micchelli, Xu, \protect\BIBand{}
  Zhang}]{micchelli2006universal}
Micchelli CA, Xu Y, Zhang H (2006) Universal kernels. \emph{Journal of Machine
  Learning Research} 7(12).

\bibitem[{Miller(2002)}]{miller2002subset}
Miller A (2002) \emph{Subset selection in regression} (chapman and hall/CRC).

\bibitem[{Mitsuzawa et~al.(2023)Mitsuzawa, Kanagawa, Bortoli, Grossi,
  \protect\BIBand{} Papotti}]{mitsuzawa2023variable}
Mitsuzawa K, Kanagawa M, Bortoli S, Grossi M, Papotti P (2023) Variable
  selection in maximum mean discrepancy for interpretable distribution
  comparison. \emph{arXiv preprint arXiv:2311.01537} .

\bibitem[{Moghaddam et~al.(2005)Moghaddam, Weiss, \protect\BIBand{}
  Avidan}]{moghaddam2005spectral}
Moghaddam B, Weiss Y, Avidan S (2005) Spectral bounds for sparse {PCA}: Exact
  and greedy algorithms. \emph{Advances in neural information processing
  systems} 18.

\bibitem[{Muandet et~al.(2017)Muandet, Fukumizu, Sriperumbudur, Sch{\"o}lkopf
  et~al.}]{muandet2017kernel}
Muandet K, Fukumizu K, Sriperumbudur B, Sch{\"o}lkopf B, et~al. (2017) Kernel
  mean embedding of distributions: A review and beyond. \emph{Foundations and
  Trends{\textregistered} in Machine Learning} 10(1-2):1--141.

\bibitem[{Mueller \protect\BIBand{} Jaakkola(2015)}]{mueller2015principal}
Mueller J, Jaakkola T (2015) Principal differences analysis: Interpretable
  characterization of differences between distributions. \emph{Advances in
  Neural Information Processing Systems}, volume~28.

\bibitem[{P{\'o}lik \protect\BIBand{} Terlaky(2007)}]{polik2007survey}
P{\'o}lik I, Terlaky T (2007) A survey of the {S}-lemma. \emph{SIAM review}
  49(3):371--418.

\bibitem[{Sch{\"o}lkopf \protect\BIBand{} Smola(2002)}]{scholkopf2002learning}
Sch{\"o}lkopf B, Smola AJ (2002) \emph{Learning with kernels: support vector
  machines, regularization, optimization, and beyond} (MIT press).

\bibitem[{Schrab et~al.(2023)Schrab, Kim, Albert, Laurent, Guedj,
  \protect\BIBand{} Gretton}]{schrab2021mmd}
Schrab A, Kim I, Albert M, Laurent B, Guedj B, Gretton A (2023) {MMD}
  aggregated two-sample test. \emph{Journal of Machine Learning Research}
  24(194):1--81.

\bibitem[{Schrab et~al.(2022)Schrab, Kim, Guedj, \protect\BIBand{}
  Gretton}]{schrab2022efficient}
Schrab A, Kim I, Guedj B, Gretton A (2022) Efficient aggregated kernel tests
  using incomplete $ u $-statistics. \emph{Advances in Neural Information
  Processing Systems} 35:18793--18807.

\bibitem[{Shalev-Shwartz \protect\BIBand{} Singer(2006)}]{shalev2006efficient}
Shalev-Shwartz S, Singer Y (2006) Efficient learning of label ranking by soft
  projections onto polyhedra. \emph{Journal of Machine Learning Research} .

\bibitem[{Sutherland et~al.(2017)Sutherland, Tung, Strathmann, De, Ramdas,
  Smola, \protect\BIBand{} Gretton}]{sutherland2016generative}
Sutherland DJ, Tung HY, Strathmann H, De S, Ramdas A, Smola A, Gretton A (2017)
  Generative models and model criticism via optimized maximum mean discrepancy.
  \emph{International Conference on Learning Representations}.

\bibitem[{Taguchi \protect\BIBand{} Rajesh(2000)}]{taguchi2000new}
Taguchi G, Rajesh J (2000) New trends in multivariate diagnosis.
  \emph{Sankhy{\=a}: The Indian Journal of Statistics, Series B} 233--248.

\bibitem[{Todd(2001)}]{todd2001semidefinite}
Todd MJ (2001) Semidefinite optimization. \emph{Acta Numerica} 10:515--560.

\bibitem[{Van Der~Maaten(2014)}]{van2014accelerating}
Van Der~Maaten L (2014) Accelerating t-sne using tree-based algorithms.
  \emph{The journal of machine learning research} 15(1):3221--3245.

\bibitem[{Wang et~al.(2023)Wang, Chen, Zhao, Liao, \protect\BIBand{}
  Xie}]{wang2022manifold}
Wang J, Chen M, Zhao T, Liao W, Xie Y (2023) A manifold two-sample test study:
  integral probability metric with neural networks. \emph{Information and
  Inference: A Journal of the IMA} 12(3):iaad018.

\bibitem[{Wang et~al.(2021)Wang, Gao, \protect\BIBand{} Xie}]{wang2021two}
Wang J, Gao R, Xie Y (2021) Two-sample test using projected {W}asserstein
  distance. \emph{2021 IEEE International Symposium on Information Theory},
  3320--3325.

\bibitem[{Wang et~al.(2022{\natexlab{a}})Wang, Gao, \protect\BIBand{}
  Xie}]{wang2020kerneltwosample}
Wang J, Gao R, Xie Y (2022{\natexlab{a}}) Two-sample test with kernel projected
  {W}asserstein distance. \emph{Proceedings of The 25th International
  Conference on Artificial Intelligence and Statistics}, volume 151 of
  \emph{Proceedings of Machine Learning Research}, 8022--8055.

\bibitem[{Wang et~al.(2022{\natexlab{b}})Wang, Moore, Xie, \protect\BIBand{}
  Kamaleswaran}]{wang2022improving}
Wang J, Moore R, Xie Y, Kamaleswaran R (2022{\natexlab{b}}) Improving sepsis
  prediction model generalization with optimal transport. \emph{Machine
  Learning for Health}, 474--488.

\bibitem[{Zeisel et~al.(2015)Zeisel, Mu{\~n}oz-Manchado, Codeluppi,
  L{\"o}nnerberg, La~Manno, Jur{\'e}us, Marques, Munguba, He, Betsholtz
  et~al.}]{zeisel2015cell}
Zeisel A, Mu{\~n}oz-Manchado AB, Codeluppi S, L{\"o}nnerberg P, La~Manno G,
  Jur{\'e}us A, Marques S, Munguba H, He L, Betsholtz C, et~al. (2015) Cell
  types in the mouse cortex and hippocampus revealed by single-cell rna-seq.
  \emph{Science} 347(6226):1138--1142.

\end{thebibliography}
}

\clearpage

\if\paperversion2
\appendix
\fi
\if\paperversion1
\ECSwitch
\ECHead{Supplementary for \emph{``Variable Selection for Kernel Two-Sample Tests''}}
\fi
\section{ADMM for Solving SDP Problem~\texorpdfstring{\eqref{Eq:cvx:relax}}{}}\label{Appendix:ADMM}

Define the domain sets 
\begin{align*}
\mathcal{C}&=\Big\{ 
Z\in\bS_{D+1}^+:~Z^{(0,0)}=1, \text{Tr}(Z)=2
\Big\},\\
\mathcal{B}&=\Bigg\{ 
(Z,q):~\sum_{j\in[D]}~\left( 
Z^{(i,j)}
\right)^2\le Z^{(i,i)}q^{(i)}, 
\left(\sum_{j\in[D]}~|Z^{(i,j)}|\right)^2\le dZ^{(i,i)}q^{(i)}, \forall i\in[D], q\in\overline{\cQ} 
\Bigg\}.
\end{align*}
Let $\mathcal{I}_{\mathcal{A}}(\cdot)$ denote the indicator function of set $\mathcal{A}$.
Then problem~\eqref{Eq:cvx:relax} can be reformulated as
\[
\min_{Z,q}~\Big\{ 
-\inp{\tilde{A}}{Z} + \mathcal{I}_{\mathcal{C}}(Z) + \mathcal{I}_{\mathcal{B}}(Z,q)
\Big\}.
\]
By introducing a new variable $Y$, the problem above can be written as
\begin{equation}
\label{Eq:reformulate:cvx:relax}
\min_{Z,Y,q}~\Big\{ 
-\inp{\tilde{A}}{Z} + \mathcal{I}_{\mathcal{C}}(Z) + \mathcal{I}_{\mathcal{B}}(Y,q):~Z=Y
\Big\}.
\end{equation}
The augmented Lagrangian function for problem~\eqref{Eq:reformulate:cvx:relax} is defined as
\begin{align*}
\mathcal{L}_{\mu}(Z,Y,q;\Lambda)&=
-\inp{\tilde{A}}{Z} + \mathcal{I}_{\mathcal{C}}(Z) + \mathcal{I}_{\mathcal{B}}(Y,q)
-\inp{\Lambda}{Z-Y} + \frac{1}{2\tau}\|Z-Y\|_F^2,
\end{align*}
where $\tau>0$ is a penalty parameter.
The ADMM approach produces the following iterations:
\begin{subequations}
\begin{align}
Z_{k+1}&=\argmin_{Z}~\mathcal{L}_{\mu}(Z,Y_k,q_k;\Lambda_k),\label{Eq:ADMM:a}\\
(Y_{k+1}, q_{k+1})&=\argmin_{Y,q}~\mathcal{L}_{\mu}(Z_{k+1},Y,q;\Lambda_k),\label{Eq:ADMM:b}\\
\Lambda_{k+1}&=\Lambda_k - \frac{1}{\tau}\big[ Z_{k+1} - Y_{k+1}\big].\label{Eq:ADMM:c}
\end{align}
\end{subequations}
The ADMM algorithm terminates at iteration $k$ if for some tolerance parameter $\texttt{tol}>0$, it holds that 
\[
\frac{\|Z_{k+1}-Y_{k+1}\|}{1 + \|\tilde{A}\|_1}\le \texttt{tol}.
\]
The advantage of ADMM is that, based on the variable splitting trick, the subproblems \eqref{Eq:ADMM:a} and \eqref{Eq:ADMM:b} are easier to solve than the original SDP problem.

Specifically, the subproblem~\eqref{Eq:ADMM:a} reduces to
\begin{equation}
Z_{k+1}=\argmin_{Z\in\mathcal{C}}~\left\| 
Z - (Y_k + \tau\tilde{A} + \tau\Lambda_k)
\right\|_F^2,
\end{equation}
which amounts to solving an eigenvalue problem. See the detailed algorithm design in Remark~\ref{Remark:Symmetric:eig}.

Next, the subproblem~\eqref{Eq:ADMM:b} reduces to 
\begin{equation}
(Y_{k+1}, q_{k+1})=\argmin_{(Y,q)\in\mathcal{B}}~\left\| 
Y - (Z_{k+1} - \tau\Lambda_k)
\right\|_F^2,
\end{equation}
which amounts to solving a large-scale second-order cone program. See the detailed algorithm design in Remark~\ref{Remark:socp}.

\begin{remark}\label{Remark:Symmetric:eig}
Given a symmetric matrix $X\in\mathbb{R}^{(D+1)\times(D+1)}$, consider the optimization problem
\[
\min_{Z\in\mathcal{C}}~\|Z-X\|_F^2.
\]
Since this problem is unitary-invariant, its optimal solution is given by $Z^* = U\diag(a^*)U\trans$ for some vector $a^*\in\mathbb{R}^{D+1}$, where the matrix $X$ admits eigendecomposition $X=U\diag(b)U\trans$.
The vector $a^*$ can be obtained by solving the following problem:
\begin{equation}
a^*=\argmin~\left\{ 
\|a - b\|_2^2:\quad a\ge0, a^{(0)}=1, \sum_{i=0}^Da^{(i)}=2
\right\}.
\label{Eq:opt:a}
\end{equation}
Such a problem is a variant of the projection problem onto the simplex in Euclidean space.
We adopt the algorithm in \citep{shalev2006efficient} with complexity $O(D\log D)$ to solve this problem.
See details in Algorithm~\ref{alg:Dlog}.
\begin{algorithm}[H]
   \caption{An $O(D\log D)$-complexity algorithm to solving problem~\eqref{Eq:opt:a}}
   \label{alg:Dlog}
\begin{algorithmic}[1]
    \STATE{Sort $b^{(1:D)}$ to $\hat{b}$ such that $\hat{b}^{(1)}\le\cdots\le \hat{b}^{(D)}$.}
    \STATE{Find smallest index $\hat{j}$ such that $\hat{b}^{(j)}-\frac{1}{D-j+1}\left( 
    \sum_{i=j}^D\hat{b}^{(i)}-1
    \right)>0$.}
    \STATE{Compute $\theta = \frac{1}{D-\hat{j}+1}\left( 
    \sum_{i=\hat{j}}^D\hat{b}^{(i)}-1
    \right)$}
   \STATE{\textbf{Return} vector $a$ such that $a^{(0)}=1$ and $a^{(i)} = \max\{0, b^{(i)} - \theta\}, i\in[D]$.}
\end{algorithmic}
\end{algorithm}
\end{remark}

\begin{remark}\label{Remark:socp}
Given a matrix $X\in\mathbb{R}^{(D+1)\times(D+1)}$, consider the optimization problem 
\[
\min_{(Y,q)\in\mathcal{B}}~\|Y-X\|_F^2.
\]
It can be reformulated as a second-order cone program that could be solved efficiently based on some off-the-shelf solver:
\[
\begin{aligned}
\min_{Y,q\in\overline{\cQ},A_i, i\in[D]}&\quad \|\text{vec}(Y) - \text{vec}(X)\|_2^2\\
\mbox{s.t.}&\quad \|Y^{(i,:)}\|_1\le A_i, i\in[D],\\ 
&\quad (2A_i, Y^{(i,i)}-dq^{(i)}, Y^{(i,i)}+dq^{(i)})\in\mathcal{C}_3, 
Y^{(i,i)}\ge0, 
i\in[D],\\
&\quad (Y^{(i,1)},\ldots, Y^{(i,i)}-\frac{1}{2}q^{(i)},\ldots, Y^{(i,D)}, \frac{1}{2}q^{(i)})\in\mathcal{C}_{D+1}, i\in[D],
\end{aligned}
\]
where $\mathcal{C}_{D+1}$ denotes the second-order cone of dimension $D+1$:
\[
\mathcal{C}_{D+1} = \{(x,t):~x\in\mathbb{R}^D, t\in\mathbb{R}, \|x\|_2\le t\}.
\]
\end{remark}

\clearpage
\section{Proof of Example~\ref{Eq:example:1}}\label{Appendix:proof:example:1}
\begin{proof}{Proof of Example~\ref{Eq:example:1}.}
Note that the population version of the objective in \eqref{Eq:formula:MMD:opt:revision} becomes
\[ 
F(z) = \MMD^2(\mathcal{N}(0,1), \mathcal{N}(0,(1+\epsilon)^2); k_1)
-
\lambda\sigma_{\cH_1}^2(\mathcal{N}(0,1), \mathcal{N}(0,(1+\epsilon)^2); k_1),\quad\text{if }z^{(1)}\ne0,
\]
when $z=\widehat{z}$ and otherwise $F(z)=0$.
Therefore, taking the variance regularization
\[
\lambda\in \left[0, \frac{\MMD^2(\mathcal{N}(0,1), \mathcal{N}(0,(1+\epsilon)^2); k_1)}{\sigma_{\cH_1}^2(\mathcal{N}(0,1), \mathcal{N}(0,(1+\epsilon)^2)}\right),
\]
achieves the desired result.
It remains to compute $\MMD^2(\mathcal{N}(0,1), \mathcal{N}(0,(1+\epsilon)^2); k_1)$ and $\sigma_{\cH_1}^2(\mathcal{N}(0,1), \mathcal{N}(0,(1+\epsilon)^2)$ to finish the proof.
According to the definition, it holds that 
\begin{align*}
&\MMD^2(\mathcal{N}(0,1), \mathcal{N}(0,(1+\epsilon)^2); k_1)\\
=&
\mathbb{E}_{x,x'\sim \mathcal{N}(0,1)}\left[ 
k_1(x,x')
\right]
+\mathbb{E}_{y,y'\sim \mathcal{N}(0,(1+\epsilon)^2)}\left[ 
k_1(y,y')
\right]-2\mathbb{E}_{x\sim \mathcal{N}(0,1),
y\sim \mathcal{N}(0,(1+\epsilon)^2)}\left[ 
k_1(x,y)
\right]
\\
=&\mathbb{E}_{x,x'\sim \mathcal{N}(0,1)}\left[ 
k_1(x,x')
+
k_1((1+\epsilon)x, (1+\epsilon)x')
-2
k_1(x, (1+\epsilon)y)
\right]\\
=&\sqrt{\frac{\tau_1^2}{\tau_1^2+2}} + \sqrt{\frac{\tau_1^2}{\tau_1^2+2(1+\epsilon)^2}}-2\sqrt{\frac{\tau_1^2}{\tau_1^2+1 + (1+\epsilon)^2}},
\end{align*}
where the last step is by substituting the expression $k_1(x,y) = e^{-(x-y)^2/(2\tau_1^2)}$ and calculating several integral of exponential functions.
Also, we have that 
\begin{align*}
&\sigma_{\cH_1}^2(\mathcal{N}(0,1), \mathcal{N}(0,(1+\epsilon)^2); k_1)
=4\mathbb{E}[H_{1,2}H_{1,3}] - 4\MMD^4(\mathcal{N}(0,1), \mathcal{N}(0,(1+\epsilon)^2); k_1).
\end{align*}
According to the definition of $H_{i,j}$ in \eqref{Eq:H:i:j}, it holds that 
\begin{align*}
\mathbb{E}[H_{1,2}H_{1,3}]&=\mathbb{E}_{x_1,x_2,x_3,x_4\sim\mathcal{N}(0,1)}\Big[k_1(x_1,x_2)k_1(x_1,x_3) + 2k_1(x_1,x_2)k_1((1+\epsilon)x_3, (1+\epsilon)x_4)\\&
-4k_1(x_1,x_2)k(x_1, (1+\epsilon)x_3)
-4k_1(x_1, (1+\epsilon)x_2)k_1((1+\epsilon)x_3, (1+\epsilon)x_4)\\
&+4k_1(x_1,(1+\epsilon)x_2)k_1(x_1, (1+\epsilon)x_3) + k_1((1+\epsilon)x_1, (1+\epsilon)x_2)k_1((1+\epsilon)x_1, (1+\epsilon)x_3)
\Big]\\
&=\sqrt{\frac{\tau_1^4}{(\tau_1^2+1)(3+\tau_1^2)}} + \sqrt{\frac{4\tau_1^4}{(\tau_1^2+2)(\tau_1^2+2(1+\epsilon)^2)}}-\sqrt{\frac{16\tau_1^4}{
2\tau_1^2+1 + (1+\epsilon)^2 + (1+\tau_1^2)((1+\epsilon)^+\tau_1^2)
}}\\
&\qquad\qquad-\sqrt{\frac{16\tau_1^4}{
(\tau_1^2+1+(1+\epsilon)^2)(\tau_1^2+2(1+\epsilon)^2)
}}+\sqrt{\frac{16\tau_1^4}{(\tau_1^2+ (1+\epsilon)^2)(\tau_1^2+ (1+\epsilon)^2 + 2)}}\\
&\qquad\qquad+\sqrt{\frac{\tau_1^4}{(\tau_1^2+(1+\epsilon)^2)(\tau_1^2 + 3(1+\epsilon)^2)}}
\end{align*}
The proof is completed.
\if\paperversion1
\QED
\fi
\end{proof}

\clearpage
\section{Proofs of Technical Results in Section~\ref{Sec:STRS}}

\begin{proof}{Proof of Theorem~\ref{Thm:Eq:MIQP:quad}.}
A natural combinatorial reformulation of \eqref{Eq:general:MIQP} is
\begin{equation}
\max_{
\substack{
S\subseteq [D]:~|S|\le d\\
z\in\bR^D
}}~\left\{ 
z\trans Az+z\trans t:~\|z\|_2= 1, z^{(k)}=0, \forall k\notin S
\right\}.
\end{equation}
Given a size-$d$ set $S\subseteq[D]$, and problem parameters $A\in\bS_D, t\in\bR^D$, it holds that
\begin{align}
&\max_{z\in\bR^D}~\left\{ 
z\trans Az+z\trans t:~\|z\|_2= 1, z^{(k)}=0, \forall k\notin S
\right\}=\max_{z\in\bR^d}~\left\{ 
z\trans A^{(S,S)}z+z\trans t^{(S)}:~\|z\|_2= 1
\right\}.\label{Eq:MIQP:intermediate}
\end{align}
Next, we linearize the problem~\eqref{Eq:MIQP:intermediate} using the auxiliary variable defined as
\[
Z = \begin{pmatrix}
1\\z
\end{pmatrix}
\begin{pmatrix}
1\\z
\end{pmatrix}\trans = \begin{pmatrix}
1&z\trans\\z&zz\trans
\end{pmatrix}
\]
and the matrix
\[
\tilde{A}^{(S,S)}= \begin{pmatrix}
0&\frac{1}{2}(t^{(S)})\trans\\
\frac{1}{2}t^{(S)}&A^{(S,S)}
\end{pmatrix}.
\]
Assume the index of $Z$ and $\tilde{A}^{(S,S)}$ is over $\{0,1,\ldots,d\}^2$.
Then we equivalently reformulate the problem~\eqref{Eq:MIQP:intermediate} as
\begin{equation}
\begin{aligned}
\max_{Z\in\bS_{d+1}^+}&\quad \inp{\tilde{A}^{(S,S)}}{Z} \\
\mbox{s.t.}&\quad \text{rank}(Z)=1,\\
    & \quad Z^{(0,0)}=1, \text{Tr}(Z)=2.
\end{aligned}
\label{Eq:MIQP:intermediate:SDP}
\end{equation}
In particular, constraints $Z\succeq 0, \text{rank}(Z)=1, Z^{(0,0)}=1$ together imply that
\[
Z = \begin{pmatrix}
1&z\trans\\z&zz\trans
\end{pmatrix}
\]
for some vector $z\in\bR^d$, and the condition $\text{Tr}(Z)= 2$ implies $\|z\|_2=1$.
By \citep[Corollary~3]{li2022exactness}, 
we further obtain the following equivalent reformulation of problem~\eqref{Eq:MIQP:intermediate} when dropping the nonconvex rank constraint $\text{rank}(Z)=1$:
\begin{equation}
\begin{aligned}
\max_{Z\in\bS_{d+1}^+}&\quad \inp{\tilde{A}^{(S,S)}}{Z} \\
\mbox{s.t.}&\quad Z^{(0,0)}=1, \text{Tr}(Z)=2.
\end{aligned}
\label{Eq:MIQP:intermediate:SDP:no:rank}
\end{equation}
In summary, we obtain the following reformulation of \eqref{Eq:general:MIQP}:
\begin{equation}
\begin{aligned}
\max_{
\substack{
Z\in\bS^+_{d+1}, S\subseteq [D]:~|S|\le d\\
}}&\quad\inp{\tilde{A}^{(S,S)}}{Z}\\
\mbox{s.t.}&\quad Z^{(0,0)}=1, \text{Tr}(Z)=2.
\end{aligned}
\label{Eq:MIQP:quad:set}
\end{equation}
It remains to show the equivalence between formulations \eqref{Eq:MISDP:quad} and \eqref{Eq:MIQP:quad:set}.
We only need to show for any feasible $q\in\cQ$ with its support $S:=\{k:~q^{(k)}=1\}$, it holds that
\begin{equation}\label{Eq:relation:last}
\begin{aligned}
&\max_{Z\in\bS_{D+1}^+}~\bigg\{ 
\inp{\tilde{A}}{Z}:~Z_{i,i}\le q^{(i)}, i\in[D], Z^{(0,0)}=1, \text{Tr}(Z)=2
\bigg\}\\
=&
\max_{
\substack{
Z\in\bS^+_{d+1}
}}~\bigg\{ 
\inp{\tilde{A}^{(S,S)}}{Z}:~Z_{0,0}=1, \text{Tr}(Z)=2
\bigg\}.
\end{aligned}
\end{equation}
Since $Z\in\bS_{D+1}^+$ is a positive semi-definite matrix, the condition $Z^{(i,i)}=0$ for $i\in[D]\setminus S$ implies
\[
Z^{(i,j)}=0,\quad \forall (i,j)\notin S\times S.
\]
Leveraging this property, we check the relation~\eqref{Eq:relation:last} indeed holds true.
\if\paperversion1
\QED
\fi
\end{proof}

\begin{proof}{Proof of Corollary~\ref{Corollary:strong:MISDP}.}
It suffices to verify the following two valid inequalities hold for problem~\eqref{Eq:MISDP:quad}:
\begin{align}
\sum_{j\in[D]}(Z^{(i,j)})^2&\le Z^{(i,i)}q^{(i)}, \quad \forall i\in[D],\label{Eq:val:a}\\
\left( 
\sum_{j\in[D]}|Z^{(i,j)}|
\right)^2&\le dZ^{(i,i)}q^{(i)}, \quad \forall i\in[D].\label{Eq:val:b}
\end{align}
This verification step follows a similar argument in \citep[Lemma~2]{li2020exact}.
\if\paperversion1
\QED
\fi
\end{proof}

\begin{proof}{Proof of Theorem~\ref{Thm:ref:Eq:MISDP:quad}.}
We first re-write $f(q)$ as the optimal value to the following optimization problem:
\begin{align*}
\max_{Z\in\bS_{D+1}^+, U\ge0, Y,y, t}&\quad \inp{\tilde{A}}{Z}\\
&\quad Z^{(0,0)}=1,\quad                           & [\lambda_0]\\
&\quad \text{Tr}(Z)=2,\quad                        & [\lambda]\\
&\quad \sum_jU^{(i,j)}\le y^{(i)},\quad \forall i\in[D], & [\beta^{(i)}]\\
&\quad -U^{(i,j)}\le Z^{(i,j)}\le U^{(i,j)},\quad \forall i,j\in[D],       & [W_1^{(i,j)}, W_2^{(i,j)}]\\ 
&\quad \left\| 
(y_i;t_i)
\right\|_2\le \frac{1}{2}Z^{(i,i)}+\frac{d}{2}q^{(i)},\quad \forall i\in[D], & [\nu_1^{(i)}]\\\
&\quad t_i=\frac{1}{2}Z^{(i,i)}-\frac{d}{2}q^{(i)},\quad \forall i\in[D], & [\nu_2^{(i)}]\\
&\quad Y^{(i,:)}=Z^{(i,:)} - \frac{1}{2}q^{(i)}e_i,\quad \forall i\in[D], &[\Lambda^{(i,:)}]\\
&\quad \left\| 
Y^{(i,:)}
\right\|_2\le \frac{1}{2}q^{(i)}, \quad\forall i\in[D]. &[\mu^{(i)}]
\end{align*}
Here, we associate dual variables with primal constraints in brackets.
In detail, constraints corresponding to $[\beta^{(i)}], [W_1^{(i,j)}, W_2^{(i,j)}], [\nu_1^{(i)}], [\nu_2^{(i)}]$ are reformulation of the valid inequality \eqref{Eq:val:b}, and constraints corresponding to $[\Lambda^{(i,:)}]$ and $[\mu^{(i)}]$ are second-order conic reformulation of the valid inequality \eqref{Eq:val:a}.

Its Lagrangian dual reformulation becomes
\begin{align*}
&\min_{\substack{
\lambda,\lambda_0,\nu_2,\Lambda\\
\beta, W_1,W_2, \nu_1,\mu\ge0
}}
\max_{Z\in\bS_{D+1}^+, U\ge0, Y,y,t}\quad 
\inp{\tilde{A}}{Z} + \lambda_0\big(1 - Z^{(0,0)}\big) + \lambda\big(2-\text{Tr}(Z)\big)
+\sum_i\beta^{(i)}\big[y^{(i)} - \sum_jU^{(i,j)}\big]
\\&\qquad \qquad
 + 
  \sum_{i,j}W^{(i,j)}_1\big[ 
U^{(i,j)} + Z^{(i,j)}
\big]
 +
 \sum_{i,j}W^{(i,j)}_2\big[ 
U^{(i,j)} - Z^{(i,j)}
\big]
+\sum_i\nu_1^{(i)}\left[ 
\frac{1}{2}Z^{(i,i)}+\frac{d}{2}q^{(i)} - \left\| 
(y_i;t_i)
\right\|_2
\right]\\ 
&\qquad
+
\sum_i\nu_2^{(i)}\left( 
\frac{1}{2}Z^{(i,i)}-\frac{d}{2}q^{(i)}-t_i
\right)
+\sum_{i}\Lambda^{(i,:)}\left[ 
Z^{(i,:)} - \frac{1}{2}q^{(i)}e_i - Y^{(i,:)}
\right] + \sum_i\mu^{(i)}\left( 
\frac{1}{2}q^{(i)} - \left\| 
Y^{(i,:)}
\right\|_2
\right).
\end{align*}
Or equivalently, it can be written as
\begin{align*}
&\min_{\substack{
\lambda,\lambda_0,\nu_2,\Lambda\\
\beta, W_1,W_2, \nu_1,\mu\ge0
}}~\Bigg\{\lambda_0 + 2\lambda + \frac{d}{2}(\nu_1-\nu_2)\trans q
+\frac{1}{2}(\mu - \diag(\Lambda))\trans q\\
&\qquad + \max_{Z\in\bS_{D+1}^+}~\left\{ 
\inp{\tilde{A}}{Z}-\lambda_0Z^{(0,0)} - \lambda\text{Tr}(Z)+  \sum_{i,j}(W^{(i,j)}_1-W^{(i,j)}_2+\Lambda^{(i,j)})Z^{(i,j)}+\frac{1}{2}\sum_i(\nu_1^{(i)}+\nu^{(i)}_2)Z^{(i,i)}
\right\}\\
&\qquad + \max_{U\ge0}~\left\{ 
-\sum_i\beta^{(i)}\sum_jU^{(i,j)} + \sum_{i,j}(W^{(i,j)}_1+W^{(i,j)}_2)U^{(i,j)}
\right\} + \max_{Y}~\left\{ 
-\sum_i\Lambda^{(i,:)}Y^{(i,:)} - \sum_i\mu^{(i)}\|Y^{(i,:)}\|_2
\right\}\\ 
&\qquad + \max_{y,t}~\left\{ 
\sum_i\beta^{(i)}y^{(i)} - \sum_i\nu_1^{(i)}\|(y_i;t_i)\|_2 - \sum_i\nu_2^{(i)}t_i
\right\}
\Bigg\}.
\end{align*}
The inner maximization over $Z$ can be simplified into the constraint
\[
\begin{pmatrix}
-\lambda_0&\frac{1}{2}t\trans\\
\frac{1}{2}t&A-\lambda I_D + W_1-W_2 + \Lambda + \frac{1}{2}\diag(\nu_1 + \nu_2)
\end{pmatrix}\preceq0.
\]
The inner maximization over $U$ can be simplified as
\[
W_1 + W_2 - \diag(\beta)\le 0.
\]
The inner maximization over $Y$ can be simplified as
\[
\sum_j(\Lambda^{(i,j)})^2\le (\mu^{(i)})^2,\quad i\in[D].
\]
The inner maximization over $(y,t)$ can be simplified as
\[
(\beta^{(i)})^2 + (\nu_2^{(i)})^2\le (\nu_1^{(i)})^2,\quad i\in[D].
\]
Combining those relations, we arrive at the dual problem
\begin{align*}
\min_{\substack{
\lambda,\lambda_0,\nu_2,\Lambda\\
\beta, W_1,W_2, \nu_1,\mu\ge0
}}&\quad \lambda_0 + 2\lambda + 
q\trans\left[ 
\frac{d}{2}(\nu_1-\nu_2) + \frac{1}{2}(\mu - \diag(\Lambda))
\right]\\
&\quad \begin{pmatrix}
-\lambda_0&\quad\frac{1}{2}t\trans\\
\frac{1}{2}t&\quad A-\lambda I_D + W_1-W_2 + \Lambda + \frac{1}{2}\diag(\nu_1 + \nu_2)
\end{pmatrix}\preceq0,\\
&\quad W_1 + W_2 - \diag(\beta)\le 0,\\
&\quad \sum_j(\Lambda^{(i,j)})^2\le (\mu^{(i)})^2,\quad i\in[D],\\
&\quad (\beta^{(i)})^2 + (\nu_2^{(i)})^2\le (\nu_1^{(i)})^2,\quad i\in[D].
\end{align*}
\if\paperversion1
\QED
\fi
\end{proof}

\begin{proof}{Proof of Theorem~\ref{Thm:optval:relaxation:ratio}.}
The left-hand-side relation is easy to show.
The proof for the right-hand-side relation is separated into two parts: 
\begin{itemize}
    \item 
$\textsf{optval}\eqref{Eq:cvx:relax}\le \|t\|_2+d\cdot \{ 
\textsf{optval}\eqref{Eq:MISDP:quad} - \min_{k}|t[k]|
\}$; 
    \item
$\textsf{optval}\eqref{Eq:cvx:relax}
\le 
\|t\|_2+D/d\cdot \textsf{optval}\eqref{Eq:MISDP:quad}$.
\end{itemize}
\begin{enumerate}
    \item[(I)]%
For any feasible solution $(q,Z)$ to \eqref{Eq:cvx:relax}, we find
\begin{align*}
\sum_it^{(i)}Z^{(0,i)}&\le \sum_i|t^{(i)}||Z^{(0,i)}|
\le \sum_i|t^{(i)}|\sqrt{Z^{(0,0)}Z^{(i,i)}}\\
&=
\sum_i|t^{(i)}|\sqrt{Z^{(i,i)}}
\le \left( 
\sum_i|t^{(i)}|^2
\right)^{1/2}
\left( 
\sum_iZ^{(i,i)}
\right)^{1/2}
=\|t\|_2,
\end{align*}
where the first inequality is due to taking absolute values, the second inequality is because $Z\succeq0$ and $|Z^{(0,i)}|\le \sqrt{Z^{(0,0)}Z^{(i,i)}}$, and the last inequality is by the Cauchy-Schwarz inequality.

As a consequence, for any feasible solution $(q,Z)$ to \eqref{Eq:cvx:relax}, it holds that
\begin{equation}
    \inp{\tilde{A}}{Z}=\sum_{i,j}A^{(i,j)}Z^{(i,j)} + \sum_it^{(i)}Z^{(0,i)}\le 
\sum_{i,j}\left|A^{(i,j)}\right|\left|Z^{(i,j)}\right| + \|t\|_2.\label{Eq:cvx:relax:optval}
\end{equation}
On the other hand, it is easy to verify that for any $i\in[D]$, the following is a feasible solution to \eqref{Eq:MISDP:quad}:
\[
Z_i = \begin{pmatrix}
1\\e_i
\end{pmatrix}
\begin{pmatrix}
1\\e_i
\end{pmatrix}\trans\quad\text{or }\quad
Z_i = \begin{pmatrix}
1\\-e_i
\end{pmatrix}
\begin{pmatrix}
1\\-e_i
\end{pmatrix}\trans,
\]
where $e_i$ is a basis vector with the $i$-th element being $1$.
This yields
\[
\textsf{optval}\eqref{Eq:MISDP:quad}\ge \max\left\{A^{(i,i)} + t^{(i)}, 
A^{(i,i)} - t^{(i)}\right\} = A^{(i,i)} + |t^{(i)}|,\quad \forall i\in[D].
\]
Therefore, we obtain
\[
A^{(i,i)}\le \textsf{optval}\eqref{Eq:MISDP:quad} - |t^{(i)}|
\le \textsf{optval}\eqref{Eq:MISDP:quad} - \min_{i\in[D]}~|t^{(i)}|,
\]
and $|A^{(i,j)}|\le \sqrt{A^{(i,i)}A^{(j,j)}}\le \textsf{optval}\eqref{Eq:MISDP:quad} - \min_{i\in[D]}~|t^{(i)}|$ for any $i,j\in[D]$.
Combining this expression with \eqref{Eq:cvx:relax:optval} implies that 
\begin{equation}
\inp{\tilde{A}}{Z}\le \sum_{i,j}\left|Z^{(i,j)}\right|\cdot\left( 
\max_{i,j}\left|A^{(i,j)}\right|
\right) + \|t\|_2\le \sum_{i,j}\left|Z^{(i,j)}\right|\cdot\left( 
\textsf{optval}\eqref{Eq:MISDP:quad} - \min_{i\in[D]}~|t^{(i)}|
\right) + \|t\|_2.\label{Eq:exress:final}
\end{equation}
Also, because of the valid inequality $\left(\sum_j\left|Z^{(i,j)}\right|\right)^2\le dZ^{(i,i)}q^{(i)}$, it holds that 
\[
\sum_{i,j}\left|Z^{(i,j)}\right|\le \sqrt{d}\sum_i\sqrt{Z^{(i,i)}q^{(i)}}
\le \sqrt{d}\left(\sum_iZ^{(i,i)}\right)^{1/2}\left(\sum_iq^{(i)}\right)^{1/2}=d.
\]
Combining this relation with \eqref{Eq:exress:final} gives the desired result.

\item[(II)]
For any feasible solution $(Z,q)$ in \eqref{Eq:MISDP:quad}, we enforce $Z^{(0,i)}=Z^{(i,0)}=0$ for $i\in[D]$, then the updated solution is still feasible, with the associated objective value
\[
\inp{Z^{([D], [D])}}{A}.
\]
Therefore, we obtain the relation
\begin{equation}
\textsf{optval}\eqref{Eq:MISDP:quad}\ge 
\max_{Z\in\bS_D^+, q\in\cQ}~\bigg\{ 
\inp{Z}{A}:~Z^{(i,i)}\le q^{(i)}, i\in[D], 
\text{Tr}(Z)=1
\bigg\}
\ge d/D\cdot \lambda_{\max}(A),
\label{Eq:relation:d:D:lambda}
\end{equation}
where the last inequality is due to \citep[Proposition~2 and proof of Theorem~5]{li2020exact}.

For any feasible solution $(Z,q)$ in \eqref{Eq:cvx:relax}, according to Part~(I), it holds that $\sum_it^{(i)}Z^{(0,i)}\le \|t\|_2$, and therefore
\begin{align*}
\inp{\tilde{A}}{Z}&=\sum_{i,j}A^{(i,j)}Z^{(i,j)} + \sum_it^{(i)}Z^{(0,i)}\le 
\inp{A}{Z^{([D], [D])}} + \|t\|_2\\
&\le \max_{Z\succeq0, \text{Tr}(Z)=1}~\inp{A}{Z} + \|t\|_2 = \lambda_{\max}(A) + \|t\|_2.
\end{align*}
Combining this relation with \eqref{Eq:relation:d:D:lambda} gives the desired result.

\end{enumerate}
\if\paperversion1
\QED
\fi
\end{proof}

\begin{proof}{Proof of Theorem~\ref{Thm:ratio:truncation}\ref{Thm:ratio:truncation:I}.}
Let $z_* = \sum_{i}y^{(i)}e_i$ be the optimal solution of \eqref{Eq:general:MIQP}, where $e_i$ is the $i$-th basis vector.
Then it holds that
\begin{align*}
\textsf{optval}\eqref{Eq:general:MIQP}&=\sum_iy^{(i)}\big[ 
e_i\trans (Az_* + t)
\big]\\
&\le \sqrt{\sum_i(y^{(i)})^2}\sqrt{\sum_i(e_i\trans (Az_* + t))^2}\\
&\le \sqrt{d}\max_{i}~e_i\trans (Az_* + t)\\
&\le \sqrt{d}\max_{i}~\Big\{\max_{z\in\cZ}~e_i\trans (Az + t)\Big\}\\
&= \sqrt{d}\max_{i}~\Big\{e_i\trans (A\hat{z}_i + t)\Big\},
\end{align*}
where the last equality is because 
\[
\hat{z}_i=\argmax_{z\in\cZ}~e_i\trans(Az) = \argmax_{z\in\cZ}~e_i\trans(Az+t).
\]
Based on the observation above, one can assert that there exists $i\in[D]$ such that
\begin{equation}
\sqrt{d}e_i\trans (A\hat{z}_i + t)\ge \textsf{optval}\eqref{Eq:general:MIQP}.
\label{Eq:relation:Thm4:I}
\end{equation}
Next, we provide the lower bound for $V_{\mathrm{(I)}}$:
\begin{align*}
V_{\mathrm{(I)}}&=\max_i~\max\Big( 
e_i\trans Ae_i + e_i\trans t, 
\hat{z}_i\trans A\hat{z}_i +\hat{z}_i\trans t
\Big)\\ 
&\ge \max_i~\left\{\max\Big( 
e_i\trans Ae_i, 
\hat{z}_i\trans A\hat{z}_i
\Big)
 + 
 \min\Big( 
e_i\trans t, \hat{z}_i\trans t
 \Big)
\right\}\\ 
&\ge \max_i~\left\{
e_i\trans A\hat{z}_i
 + 
 \min\Big( 
e_i\trans t, \hat{z}_i\trans t
 \Big)
\right\}\\ 
&=
\max_i~\left\{
e_i\trans (A\hat{z}_i + t)
 + 
 \min\Big( 
0, (\hat{z}_i-e_i)\trans t
 \Big)
\right\}\\
&\ge 
\frac{1}{\sqrt{d}}\textsf{optval}\eqref{Eq:general:MIQP} + \max_i~ \min\Big( 
0, (\hat{z}_i-e_i)\trans t
 \Big)\\ 
 &\ge 
\frac{1}{\sqrt{d}}\textsf{optval}\eqref{Eq:general:MIQP} - 2\|t\|_{(d+1)},
\end{align*}
where the second inequality is because $A\succeq0$ and $0\le (e_i - \hat{z}_i)\trans A(e_i-\hat{z}_i)=(e_i\trans Ae_i + \hat{z}_i\trans A\hat{z}_i)-2e_i\trans A\hat{z}_i$, i.e., 
\[
e_i\trans A\hat{z}_i\le \frac{1}{2}(e_i\trans Ae_i + \hat{z}_i\trans A\hat{z}_i)\le \max\Big( 
e_i\trans Ae_i, 
\hat{z}_i\trans A\hat{z}_i
\Big),
\]
the third inequality is due to \eqref{Eq:relation:Thm4:I}, and the last inequality is because $\hat{z}_i-e_i$ is a $(d+1)$-sparse vector with $\|\hat{z}_i-e_i\|_2=2$, and
\[
\max_i~ \min\Big( 
0, (\hat{z}_i-e_i)\trans t
 \Big)
 \ge-\max_i\max_{a:~\|a\|_0\le d+1, \|a\|_2\le 2}~a\trans t = -2\|t\|_{(d+1)}.
\]
The proof is completed.
\if\paperversion1
\QED
\fi
\end{proof}

\begin{proof}{Proof of Theorem~\ref{Thm:ratio:truncation}\ref{Thm:ratio:truncation:II}.}
By \citep[Theorem~1.1]{adachi2017solving}, the primal-dual pair $(v,\lambda)$ of the trust region subproblem satisfies the following:
\[
\left\{
\begin{aligned}
(A - \lambda I)v&=-t\\
A&\preceq \lambda I\\ 
\|v\|_2&\le 1\\
\lambda(1 - \|v\|_2)&=0
\end{aligned}
\right.
\]
Let $\bar{z}$ be the $d$-sparse truncation of $v$.
Then it holds that
\begin{align*}
z\trans Av + t\trans z&=
z\trans Av +z\trans(-A + \lambda I)v\\
&=\lambda z\trans v=\lambda z\trans \bar{z}=\lambda\|\bar{z}\|_2\ge \lambda\sqrt{\frac{d}{D}}.
\end{align*}
On the other hand, 
\[
z\trans Av + t\trans z
\le 
\sqrt{z\trans Az}\sqrt{v\trans Av} + t\trans x \le \sqrt{z\trans Az}\cdot\Big( 
\lambda - t\trans v
\Big)^{1/2} + t\trans z,
\]
where the last inequality is because $(A-\lambda I)v=-t$ and therefore
\[
v\trans Av+t\trans v = \lambda\|v\|_2^2\le \lambda.
\]
By re-arrangement, it holds that
\[
\sqrt{\frac{d}{D}}\lambda\le \sqrt{z\trans Az}\cdot\Big( 
\lambda - t\trans v
\Big)^{1/2} + t\trans z.
\]
Or equivalently, the dual multiplier $\lambda$ satisfies
\[
\frac{d}{D}\lambda^2 - \left[ 
2\sqrt{\frac{d}{D}}z\trans t + z\trans Az
\right]\lambda + (z\trans t)^2 + (z\trans Az)(v\trans t)\le 0.
\]
Consequently, 
\[
\frac{d}{D}\lambda^2 - \left[ 
2\sqrt{\frac{d}{D}}\|t\|_{(d)} + z\trans Az
\right]\lambda - (z\trans Az)\|t\|\le 0.
\]
The determinant of the quadratic function on the left-hand-side above is non-negative:
\[
\Delta:=\left[ 
2\sqrt{\frac{d}{D}}\|t\|_{(d)} + z\trans Az
\right]^2 + \frac{4d}{D}z\trans Az\|t\|\ge 0.
\]
On the other hand,
\[
\sqrt{\Delta}\le z\trans Az + 2\sqrt{\frac{d}{D}}\|t\|_{(d)} + \frac{2d}{D}\|t\|_2.
\]
Hence, we find the upper bound of $\lambda$:
\begin{align*}
\lambda&\le \frac{2\sqrt{\frac{d}{D}}\|t\|_{(d)} + z\trans Az + \sqrt{\Delta}}{2\frac{d}{D}}\\
&\le 
\frac{D}{d}z\trans Az + \|t\|_2 + \sqrt{\frac{D}{d}}\|t\|_{(d)}\\
&\le \frac{D}{d}V_{\mathrm{(II)}} + \|t\|_2 + \left(\sqrt{\frac{D}{d}}+\frac{D}{d}\right)\|t\|_{(d)}.
\end{align*}
This, together with the fact that $\lambda \ge v\trans Av + t\trans v\ge \textsf{optval}\eqref{Eq:general:MIQP}$ completes the proof.
\if\paperversion1
\QED
\fi
\end{proof}

\begin{proof}{Proof of Theorem~\ref{Pro:relax:noncvx}.}
Define the following two sets:
\begin{align*}
T_d&:=\{z\in\mathbb{R}^D:~\|z\|_2\le 1, \|z\|_1\le\sqrt{d}\},\\
S_d&:=\{z\in\mathbb{R}^D:~\|z\|_2\le 1, \|z\|_0\le d\}.
\end{align*}
It has been shown in \citep[Lemma~1]{dey2022using} that there exists a factor $\rho\in(1, 1+\sqrt{d/(d+1)}]$ such that
\[
T_d\subseteq \rho\cdot\mathrm{conv}(S_d).
\]
It follows that
\begin{align*}
\textsf{optval}\eqref{Eq:general:MIQP:noncvx:relax}
\le&\max_{z\in \rho\cdot\mathrm{conv}(S_d)}~\{z\trans Az + z\trans t\}
=  \max_{z\in \mathrm{conv}(S_d)}~\{\rho^2 z\trans Az + \rho z\trans t\}\\
\leq & \max_{z\in \mathrm{conv}(S_d)}~\{\rho^2 z\trans Az + \rho^2 z\trans t\} = \rho^2\cdot \max_{z\in \mathrm{conv}(S_d)}~\{z\trans Az + z\trans t\} \\
=&\rho^2\textsf{optval}\eqref{Eq:general:MIQP},
\end{align*}
where the second inequality follows from the fact that in the optimal solution to the problem $\max_{z\in \mathrm{conv}(S_d)}~\{\rho^2 z\trans Az + \rho z\trans t\}$, we have that $z\trans t \geq 0$. The last equality is because the objective function $z\trans Az + z\trans t$ is a convex function. 
\if\paperversion1
\QED
\fi
\end{proof}

\begin{proof}{Proof of Proposition~\ref{Proposition:CVX:IP:reformulate}.}
The proof of this proposition is a simple extension from~\citep{dey2022using}.
\if\paperversion1
\QED
\fi
\end{proof}

\clearpage
\section{Extension of Technical Results in Section~\ref{Sec:test:power} for a Generic Kernel}
\label{Sec:extension:test:power}

We first make the following assumptions regarding the kernel choice $K_z(\cdot,\cdot)$, variance regularization value $\lambda$, and data distributions $\mu,\nu$.
\begin{assumption}\label{Assumptions:kernel}
The kernel $K_z(\cdot,\cdot)$ is uniformly bounded and satisfies the Lipshitz continuous condition, i.e., for any $z,z'\in\cZ, x,y\in\Omega$, it holds that $|K_z(x,y)|\le M$ and $|K_{z}(x,y) - K_{z'}(x,y)|\le L\|z-z'\|_2$.
\end{assumption}
\begin{assumption}\label{Assumption:regu:var}
Under the alternative hypothesis $\cH_1:~\mu\ne\nu$, there exists $\lambda\ge0$ such that for some $\zeta\in\cZ$, it holds that $\MMD^2(\mu,\nu;K_{\zeta})>0$ and
\begin{equation}
\Delta_{\zeta}\triangleq
\MMD^2(\mu,\nu;K_{\zeta})-\lambda\big[ 
\max_{z\in\cZ}\sigma^2_{\cH_1}(\mu,\nu;K_z)
-
\min_{z\in\cZ}\sigma^2_{\cH_1}(\mu,\nu;K_z)
\big]>0.\label{Eq:technical:variance}
\end{equation}
Here $\sigma^2_{\cH_1}(\mu,\nu;K_z)$ denotes the population version of the empirical variance estimator defined in \eqref{Eq:MMD:variance:empirical}.
\end{assumption}
Assumption~\ref{Assumptions:kernel} is a standard assumption used in the statistical analysis of kernel-based testing in literature.
Assumption~\ref{Assumption:regu:var} is imposed to ensure the expected value of the testing statistic is strictly positive.
It is worth noting that this assumption is not too restrictive.
In the following, we demonstrate that, under mild conditions, our proposed kernels in \eqref{Eq:linear:kernel}-\eqref{Eq:Gaussian:kernel} indeed satisfy Assumptions \ref{Assumptions:kernel} and \ref{Assumption:regu:var}.

\begin{proposition}[Sufficient Condition of {Assumptions \ref{Assumptions:kernel} and \ref{Assumption:regu:var}}]\label{Proposition:stuff:Assum}
\begin{enumerate}
    \item(Linear Kernel)
For the kernel in \eqref{Eq:linear:kernel}, Assumption~\ref{Assumptions:kernel} is guaranteed to hold with $M=\sqrt{d}$ and $L=\sqrt{2d}$.
As long as there exists $s^*\in[D]$ such that $\mathrm{Proj}_{s^*\#}\mu\ne \mathrm{Proj}_{s^*\#}\nu$, Assumption~\ref{Assumption:regu:var} is guaranteed to hold with\[
\lambda\in\left[0, \frac{\max_{z\in\cZ}\sum_{s\in[D]}z^{(s)}\MMD^2(\mathrm{Proj}_{s\#}\mu, \mathrm{Proj}_{s\#}\nu; k_s)}{16d}\right).
\]
    \item(Quadratic Kernel)
For the kernel in \eqref{Eq:quadratic:kernel}, Assumption~\ref{Assumptions:kernel} is guaranteed to hold with $M=2c^2+2d$ and $L=4d+2c\sqrt{2d}$.
As long as there exists $s^*\in[D]$ such that $\mathrm{Proj}_{s^*\#}\mu\ne \mathrm{Proj}_{s^*\#}\nu$ or 
$(\mathcal{A}(\mu, \nu))^{(s^*, s^*)}>0$ holds, Assumption~\ref{Assumption:regu:var} is guaranteed to hold with
\[
\lambda\in\left[0, \frac{\max_{z\in\cZ}z\trans \mathcal{A}(\mu,\nu)z + z\trans\mathcal{T}(\mu,\nu)}{16(2d+2c^2)^2}\right).
\]
     \item(Gaussian Kernel)
For the kernel in \eqref{Eq:Gaussian:kernel}, if additionally assuming that $\Omega\subseteq \{x\in\bR^D:~\|x\|_{\infty}\le R\}$, Assumption~\ref{Assumptions:kernel} is guaranteed to hold with $M=1$ and $L=\frac{2R}{\sigma\sqrt{e}}$.
As long as there exists $S\subseteq [D]$ with $|S|\le d$ such that $\mathrm{Proj}_{S\#}\mu\ne \mathrm{Proj}_{S\#}\nu$, Assumption~\ref{Assumption:regu:var} is guaranteed to hold with
\[
\lambda\in\left[0, \frac{\max_{z\in\cZ}\MMD^2(\mu,\nu; K_z)}{16}\right).
\]
\end{enumerate}
\end{proposition}

\begin{proposition}[Non-asymptotic Concentration Properties]\label{Thm:con:prop}
Under Assumption~\ref{Assumptions:kernel}, with probability at least $1-\delta$,
(i) the bias approximation error can be bounded as
\begin{align*}
\sup_{z\in\cZ}
&\Big|
S^2(\cv x^n,\cv y^n; K_z)
-\MMD^2(\mu,\nu;K_z)
\Big|
\le \epsilon_{n,\delta}^1\triangleq\frac{8}{\sqrt{n}}
\left[
M\sqrt{2\log\binom{D}{d}\frac{2}{\delta}+ 2d\log(4\sqrt{n})}+L
\right]\\
&\qquad\qquad\qquad\qquad =O\left( 
\frac{1}{\sqrt{n}}\left[ 
M\cdot\left(d(\log n + \log\frac{D}{d} + \log\frac{1}{\delta})\right)^{1/2}
+
L
\right]
\right),
\end{align*}
where $O(\cdot)$ hides constants that are independent to parameters $D,d,n,M,L$.\\
(ii) and the variance approximation error can be bounded as
\[
\begin{aligned}
&\sup_{z\in\cZ}
\Big|
\widehat{\sigma}^2_{\mathcal{H}_1}(\cv x^n, \cv y^n; K_z)
-\sigma^2_{\mathcal{H}_1}(\mu,\nu;K_z)
\Big|
\le \epsilon_{n,\delta}^2\triangleq\frac{64}{\sqrt{n}}\left[ 
7\sqrt{2\log\binom{D}{d}\frac{2}{\delta} + 2d\log(4\sqrt{n})} + \frac{18M^2}{\sqrt{n}} + 8LM
\right]\\
&\qquad\qquad\qquad\qquad\qquad\quad =O\left( 
\frac{1}{\sqrt{n}}\cdot \left[ 
LM + \frac{M^2}{\sqrt{n}} + \left(d(\log n + \log\frac{D}{d} + \log\frac{1}{\delta})\right)^{1/2}
\right]
\right),
\end{aligned}
\]
where $\sigma^2_{\mathcal{H}_1}(\mu,\nu;K_z)\triangleq \mathbb{E}_{\cv x^n\sim\mu,\cv y^n\sim\nu}[\widehat{\sigma}^2_{\mathcal{H}_1}(\cv x^n, \cv y^n; K_z)]$.
\end{proposition}
Proof of the proposition above follows similar covering number arguments in \citep[Theorem~6]{liu2020learning}.
The main difference is that when applying union bound on the set $\cZ$, the corresponding error bound is sharper because the covering number of sparse-constrained set $\cZ$ is much smaller.

\section{Proofs of Technical Results in Section~\ref{Sec:test:power}}
\begin{proof}{Proof of Proposition~\ref{Proposition:stuff:Assum}.}
\begin{enumerate}
    \item 
We first verify the boundness and Lipschitz continuity conditions.
Specifically, it holds that 
\begin{align*}
|K_z(x,y)|&\le \sum_{s\in[D]}|z^{(s)}|\le
\max_{z\in\cZ}~\sum_{s\in[D]}|z^{(s)}|\le 
\max_{z\in\mathbb{R}^d:~\|z\|_2=1}~\sum_{s\in[d]}|z^{(s)}|\le \sqrt{d},
\end{align*}
where the first inequality is because $|k_s(x,y)|\le 1$ for any $x,y\in\mathbb{R}$, and the third inequality is because any vector in $\cZ$ only has at most $d$ non-zero entries.
Next, we find
\begin{align*}
|K_z(x,y) - K_{z'}(x,y)|&=\left|\sum_{s\in[D]}
(z^{(s)} - (z')^{(s)})k_s(x^{(s)}, y^{(s)})\right|\\
&\le \sum_{s\in[D]}|z^{(s)} - (z')^{(s)}|=\|z-z'\|_1\\
&\le \sqrt{2d}\|z-z'\|_2,
\end{align*}
where the first inequality is because $|k_s(x, y)|\le 1$ for any $x,y\in\mathbb{R}, s\in[D]$, the second inequality is because the vector $z-z'$ only has at most $2d$ non-zero entries.
The remaining of Part~(I) can be proved by noting that 
\[
\MMD^2(\mu,\nu, K_{\bar{z}})\le \max_{z\in\cZ}\sum_{s\in[D]}z^{(s)}\MMD^2(\mathrm{Proj}_{s\#}\mu, \mathrm{Proj}_{s\#}\nu; k_s)
\]
and
\[
\max_{z\in\cZ}~\left| 
\sigma^2_{\cH_1}(\mu,\nu;K_z)
\right| \le 8d.
\]
\item
For quadratic kernel, we find
\begin{align*}
|K_z(x,y)|&\le 2\left(
\sum_{s\in[D]}z^{(s)}k_s(x^{(s)}, y^{(s)})
\right)^2 + 2c^2\le 2d + 2c^2,
\end{align*}
where the first inequality is based on the relation $(a+b)^2\le 2a^2+2b^2$, and the second inequality is because in Part~(I) we have shown that $|\sum_{s\in[D]}z^{(s)}k_s(x^{(s)}, y^{(s)})|\le \sqrt{d}$.

Besides, it holds that
\begin{align*}
|K_z(x,y) - K_{z'}(x,y)|&=\left|\sum_{s\in[D]}
(z^{(s)} - (z')^{(s)})k_s(x^{(s)}, y^{(s)})\right|
\left|\sum_{s\in[D]}
(z^{(s)} + (z')^{(s)})k_s(x^{(s)}, y^{(s)})+2c\right|.
\end{align*}
Recall the first term on the right-hand-side can be bounded by $\sqrt{2d}\|z-z'\|_2$, and the second term can be upper bounded by a constant:
\begin{align*}
&\left|\sum_{s\in[D]}
(z^{(s)} + (z')^{(s)})k_s(x^{(s)}, y^{(s)})+2c\right|\\
\le&
\sum_{s\in[D]}
|z^{(s)} + (z')^{(s)}||k_s(x^{(s)}, y^{(s)})| + 2c\\
\le&
\sum_{s\in[D]}|z^{(s)} + (z')^{(s)}|+ 2c\\
\le&\max_{v:~\|v\|_0\le 2d, \|v\|_2\le 2}\|v\|_1 + 2c\le2\sqrt{2d} + 2c.
\end{align*}
Combining those two relations gives the desired result.
The remaining of Part~(II) can be proved by noting that 
\[
\MMD^2(\mu,\nu, K_{\bar{z}})\le \max_{z\in\cZ}\max_{z\in\cZ}z\trans \mathcal{A}(\mu,\nu)z + z\trans\mathcal{T}(\mu,\nu)
\]
and
\[
\max_{z\in\cZ}~\left| 
\sigma^2_{\cH_1}(\mu,\nu;K_z)
\right| \le 8(2d+2c^2)^2.
\]
\item
The boundness of the Gaussian kernel is easy to check. 
The Lipscthiz continuity condition of the Gaussian kernel follows from \citep[Lemma~20]{liu2020learning}.
The remaining of Part~(III) can be proved by noting that 
\[
\max_{z\in\cZ}~\left| 
\sigma^2_{\cH_1}(\mu,\nu;K_z)
\right| \le 8.
\]
\end{enumerate}
\if\paperversion1
\QED
\fi
\end{proof}

Before showing the proof of Theorem~\ref{Thm:con:prop}, we list two useful technical lemmas.

\begin{lemma}[{\citep[Theorem~10]{Gretton12}}]\label{Lemma:G:A}
Assume the kernel $K_z(\cdot,\cdot)$ is uniformly bounded, i.e., for any $z\in\cZ, x,y\in\Omega$, it holds that $|K_z(x,y)|\le M$.
For fixed $z\in\cZ$, with probability at least $1-\delta$, 
\[
\Big|
S^2(\cv x^n,\cv y^n; K_z)
-\MMD^2(\mu,\nu;K_z)
\Big|\le \frac{16M}{\sqrt{2n}}\sqrt{\log\frac{2}{\delta}}.
\]
\end{lemma}
\begin{lemma}[{\citep[Lemma~17 and 18]{liu2020learning}}]\label{Lemma:G:B}
Assume the kernel $K_z(\cdot,\cdot)$ is uniformly bounded, i.e., for any $z\in\cZ, x,y\in\Omega$, it holds that $|K_z(x,y)|\le M$.
For fixed $z\in\cZ$, with probability at least $1-\delta$, 
\[
\Big|
\hat{\sigma}^2_{\mathcal{H}_1}(\cv x^n, \cv y^n; K_z)
-\sigma^2_{\mathcal{H}_1}(\mu,\nu;K_z)
\Big|\le 
448\sqrt{\frac{2}{n}\log\frac{2}{\delta}} + \frac{1152M^2}{n}.
\]
\end{lemma}

\begin{proof}{Proof of Theorem~\ref{Thm:con:prop}}
We first consider an $\epsilon$-cover of $\cZ$, denoted as $\{z_i\}_{i\in[T]}$. 
According to the definition of $\cZ$, it can be shown that $T\le \binom{D}{d}(4/\epsilon)^d$.
Applying the union bound regarding the concentration inequality in Lemma~\ref{Lemma:G:A}, we obtain with probability at least $1-\delta$,
\[
\max_{z\in\{z_i\}_{i\in[T]}}
\Big|
S^2(\cv x^n,\cv y^n; K_z)
-\MMD^2(\mu,\nu;K_z)
\Big|\le \frac{16M}{\sqrt{2n}}\sqrt{\log\frac{2T}{\delta}}.
\]
For any $z\in\cZ$, there exists $z'$ from $\{z_i\}_{i\in[T]}$ such that $\|z-z'\|_2\le \epsilon$.
Based on the Lipschitz assumption regarding the kernel function, we find with probability at least $1-\delta$, it holds that
\begin{align*}
&\sup_{z\in\cZ}~\Big|
S^2(\cv x^n,\cv y^n; K_z)
-\MMD^2(\mu,\nu;K_z)
\Big|\\
\le&\max_{z\in\{z_i\}_{i\in[T]}}
\Big|
S^2(\cv x^n,\cv y^n; K_z)
-\MMD^2(\mu,\nu;K_z)
\Big|+8L\epsilon\\
\le&\frac{16M}{\sqrt{2n}}\sqrt{\log\frac{2T}{\delta}}+8L\epsilon\le\frac{16M}{\sqrt{2n}}\sqrt{\log\binom{D}{d}\frac{2}{\delta} +d\log\frac{4}{\epsilon}}+8L\epsilon.
\end{align*}
Setting $\epsilon=1/\sqrt{n}$ gives the desired result.

Next, applying the union bound regarding the concentration inequality in Lemma~\ref{Lemma:G:B}, we obtain with probability at least $1-\delta$,
\[
\max_{z\in\{z_i\}_{i\in[T]}}
\Big|
\hat{\sigma}^2_{\mathcal{H}_1}(\cv x^n, \cv y^n; K_z)
-\sigma^2_{\mathcal{H}_1}(\mu,\nu;K_z)
\Big|\le 448\sqrt{\frac{2}{n}\log\frac{2T}{\delta}} + \frac{1152M^2}{n}.
\]
Similar as in the first part, we find with probability at least $1-\delta$, it holds that
\begin{align*}
&\sup_{z\in\cZ}~\Big|
\hat{\sigma}^2_{\mathcal{H}_1}(\cv x^n, \cv y^n; K_z)
-\sigma^2_{\mathcal{H}_1}(\mu,\nu;K_z)
\Big|\\
\le&\max_{z\in\{z_i\}_{i\in[T]}}
\Big|
\hat{\sigma}^2_{\mathcal{H}_1}(\cv x^n, \cv y^n; K_z)
-\sigma^2_{\mathcal{H}_1}(\mu,\nu;K_z)
\Big|+512LM\epsilon\\
\le&448\sqrt{\frac{2}{n}\log\frac{2T}{\delta}} + \frac{1152M^2}{n} + 512LM\epsilon\\
\le&448\sqrt{\frac{2}{n}\log\binom{D}{d}\frac{2}{\delta} + \frac{2}{n}d\log\frac{4}{\epsilon}} + \frac{1152M^2}{n} + 512LM\epsilon.
\end{align*}
Also, setting $\epsilon=1/\sqrt{n}$ gives the desired result.
\end{proof}

\begin{proof}{Proof of Theorem~\ref{Thm:asymptotic}.}
To simplify notation, let us define the population version of the objective in \eqref{Eq:formula:MMD:opt:revision} as follows:
\[
F^*(z) = \MMD^2(\mu,\nu;K_z) - \lambda\sigma^2_{\mathcal{H}_1}(\mu,\nu;K_z).
\]
We first derive the lower bound of $F^*(\hat{z}_{\mathrm{Tr}})$ in terms of $F^*(\bar{z})$ with $\bar{z}$ defined in Assumption~\ref{Assumption:regu:var} using concentration analysis.
It is clear that $|F^*(z) - F(z)|\le \epsilon_{n, \delta/2}$ with probability at least $1-\delta$.
As a consequence, with probability at least $1-\delta$, it holds that 
\begin{equation}
F^*(\hat{z}_{\mathrm{Tr}})\ge F(\hat{z}_{\mathrm{Tr}}) - \epsilon_{n_{\Tr},\delta/4}
\ge F(\bar{z}) - \epsilon_{n_{\Tr},\delta/4}
\ge F^*(\bar{z}) - 2\epsilon_{n_{\Tr},\delta/4},\label{Eq:F:pop}
\end{equation}
where we use this observation in the first and last inequalities, and the second inequality is because of the sub-optimality of $\bar{z}$.
Now we are ready to show part~(I) of this theorem.
By definition, we find
\begin{align*}
\mathbb{E}[T_{n_{\Te}}]&=\MMD^2(\mu,\nu;K_{\hat{z}_{\mathrm{Tr}}})\\
&=F^*(\hat{z}_{\mathrm{Tr}}) + \lambda\sigma^2_{\mathcal{H}_1}(\mu,\nu;K_{\hat{z}_{\mathrm{Tr}}})
\ge F^*(\hat{z}_{\mathrm{Tr}}) + \lambda\min_{z\in\cZ}\sigma^2_{\cH_1}(\mu,\nu;K_z).
\end{align*}
Combining the relation above and \eqref{Eq:F:pop} implies that, with probability at least $1-\delta$, it holds that 
\[
\mathbb{E}[T_{n_{\Te}}]\ge F^*(\bar{z}) - 2\epsilon_{n_{\Tr},\delta/4}+ \lambda\min_{z\in\cZ}\sigma^2_{\cH_1}(\mu,\nu;K_z)=\Delta_{\bar{z}} - 2\epsilon_{n_{\Tr},\delta/4}.
\]
The second part of this theorem follows from \citep[Theorem~12]{Gretton12}.
\if\paperversion1
\QED
\fi
\end{proof}

\begin{lemma}[Asymptotics of Inverse Error Function~{\citep{dominici2003inverse}}]\label{Lemma:lemma:inv}
Denote by $S(x)$ the inverse of the error function
\[
\Phi(x):=\frac{1}{\sqrt{2\pi}}\int_{-\infty}^xe^{-t^2/2}\diff t.
\]
As $x\to1$, it holds that 
\[
S(x)\to\sqrt{\mathcal{LW}\left( 
\frac{1}{2\pi(x-1)^2}
\right)},
\]
where $\mathcal{LW}(x)$ denotes the function Lambert $W(x)$ admitting the series expansion 
\[
\mathcal{LW}(x) = \sum_{n\ge1}~\frac{(-1)^{n-1}}{n!}x^n.
\]
Specifically, $\mathcal{LW}(x)\to \ln(x) - \ln\ln(x)$ as $x\to\infty$.
\end{lemma}

\begin{proof}{Proof of Theorem~\ref{Thm:non:power}.}
    It is worth noting that 
\begin{align*}
\bP(T_{n_{\Te}}>t_{\mathrm{thres}})&=\bP\left(T_{n_{\Te}}>\frac{\tau}{n_{\Te}}\right)
=
1 - \bP\left(
\frac{\sqrt{n_{\Te}}(T_{n_{\Te}} - \mathbb{E}T_{n_{\Te}})}{\sigma_{\cH_1}}
\le 
\frac{\tau}{\sigma_{\cH_1}\sqrt{n_{\Te}}}
-\frac{\sqrt{n_{\Te}}\mathbb{E}T_{n_{\Te}}}{\sigma_{\cH_1}}
\right)\\
&\ge 1 - \Phi\left( 
\frac{\tau}{\sigma_{\cH_1}\sqrt{n_{\Te}}}
-\frac{\sqrt{n_{\Te}}\mathbb{E}T_{n_{\Te}}}{\sigma_{\cH_1}}
\right) - \frac{C\rho}{\sigma_{\cH_1}^3\sqrt{n_{\Te}}},
\end{align*}
where  for the inequality above we apply the Berry–Esseen theorem to argue that the distribution of $\sqrt{n_{\Te}}(T_{n_{\Te}} - \mathbb{E}T_{n_{\Te}})/\sigma_{\cH_1}$ can be approximated as the normal distribution with residual error $O(1/\sqrt{n_{\Te}})$.
Therefore, as long as we ensure that 
\[
\frac{\tau}{\sigma_{\cH_1}\sqrt{n_{\Te}}}
-\frac{\sqrt{n_{\Te}}\mathbb{E}T_{n_{\Te}}}{\sigma_{\cH_1}}
\le \Phi^{-1}(\epsilon)
\Longleftrightarrow 
\mathbb{E}T_{n_{\Te}}\ge \frac{\tau}{n_{\Te}} + \frac{\Phi^{-1}(1 - \epsilon)}{\sqrt{n_{\Te}}},
\]
it holds that the testing power is lower bounded:
\[
\bP(T_{n_{\Te}}>t_{\mathrm{thres}})\ge 1-\epsilon - \frac{C\rho}{\sigma_{\cH_1}^3\sqrt{n_{\Te}}}.
\]
Taking $\epsilon = 1/\sqrt{n_{\Te}}$ and applying the asymptotic formula on the inverse cdf $\Phi^{-1}(\cdot)$ in Lemma~\ref{Lemma:lemma:inv} gives the desired result.
The type-I risk upper bound follows a similar argument.
\if\paperversion1
\QED
\fi
\end{proof}

\end{document}